\documentclass{article} % For LaTeX2e
\usepackage[dvipsnames]{xcolor}
\usepackage{iclr2025_conference,times}
\usepackage{amssymb}

\usepackage{phi-shortcuts}
\usepackage{tabularx}

\title{Depth Pro: Sharp Monocular Metric Depth in Less Than a Second}

\author{Aleksei Bochkovskii\And Ama\"{e}l Delaunoy\And Hugo Germain\And Marcel Santos\AND Yichao Zhou\And Stephan R. Richter\\[8pt]Apple\And Vladlen Koltun}

\iclrfinalcopy 
\begin{document}

\maketitle

\begin{abstract}
We present a foundation model for zero-shot metric monocular depth estimation. Our model, Depth Pro, synthesizes high-resolution depth maps with unparalleled sharpness and high-frequency details. The predictions are metric, with absolute scale, without relying on the availability of metadata such as camera intrinsics. And the model is fast, producing a 2.25-megapixel depth map in 0.3 seconds on a standard GPU. These characteristics are enabled by a number of technical contributions, including an efficient multi-scale vision transformer for dense prediction, a training protocol that combines real and synthetic datasets to achieve high metric accuracy alongside fine boundary tracing, dedicated evaluation metrics for boundary accuracy in estimated depth maps, and state-of-the-art focal length estimation from a single image. Extensive experiments analyze specific design choices and demonstrate that Depth Pro outperforms prior work along multiple dimensions. 
We release code \& weights at \url{https://github.com/apple/ml-depth-pro}
\end{abstract}

\section{Introduction}
\label{sec:intro}
\begin{figure}[t!]
    \centering
    \renewcommand{\arraystretch}{0.5}
    \begin{tabular}{@{}c@{\hspace{0.3mm}}c*{2}{@{\hspace{0.2mm}}c}@{}}
        \raisebox{1.3cm}[0pt][0pt]{\rotatebox[origin=c]{90}{\footnotesize Input}}&
        \stackinset{l}{3pt}{t}{0.5pt}{\adjincludegraphics[height=2cm,trim={{.35\width} {.35\height} {.6\width} {.5\height}},clip,cfbox=White 0.25mm -0.25mm]{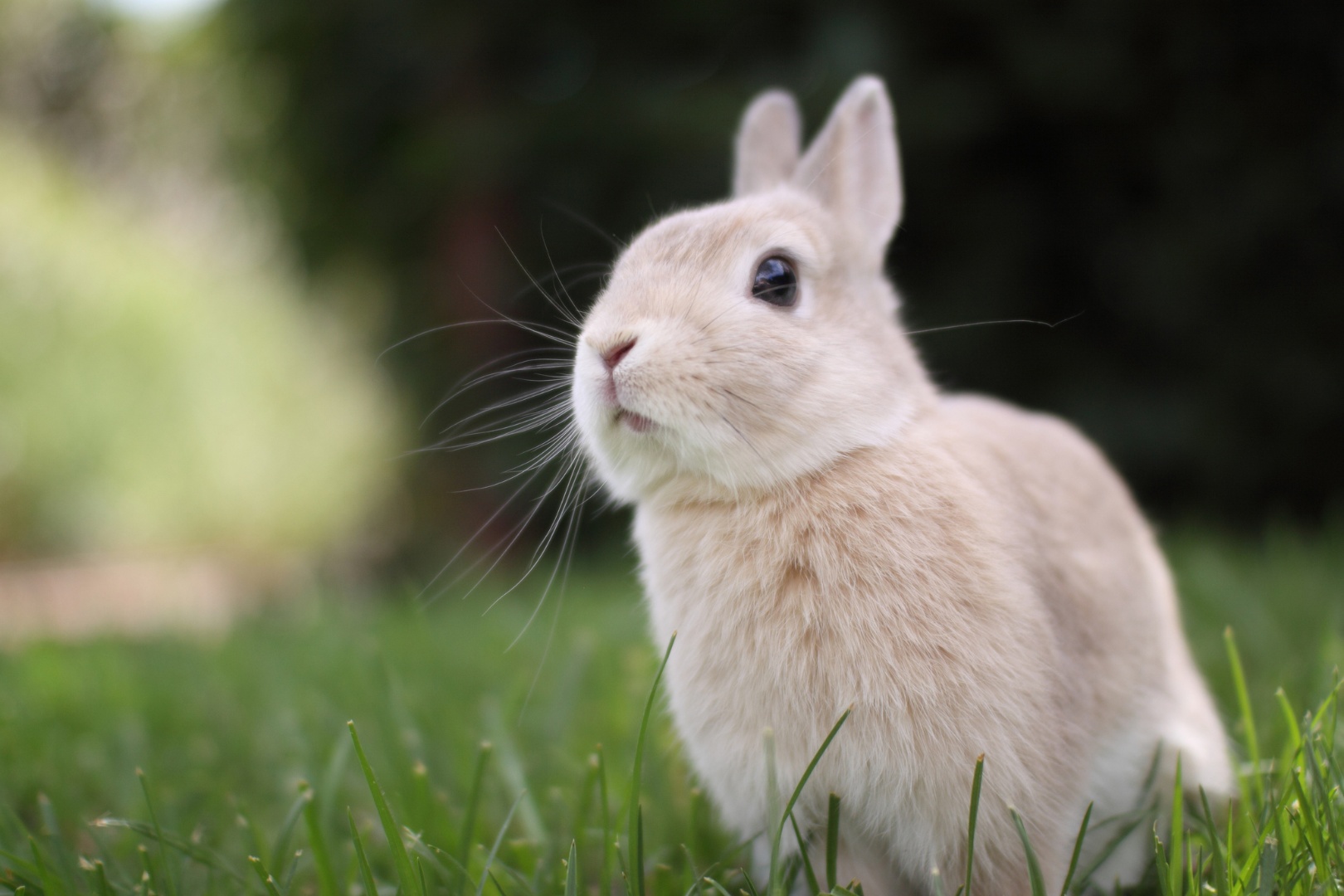}}{
            \begin{tikzpicture}
                \node[anchor=south west,inner sep=0] (image) at (0,0) {\begin{overpic}[height=0.2\textwidth]
                {fig/am2k/m_0aa9dd6a.jpg}
                % \put (2,2) {\textcolor{white}{Input image}}
                \end{overpic}
                };
                \begin{scope}[x={(image.south east)},y={(image.north west)}]
                \draw[white,thick] (0.35,0.35) rectangle (0.4,0.5);
                \end{scope}
            \end{tikzpicture}
        }
        &
        \stackinset{r}{3pt}{b}{0.5pt}{\adjincludegraphics[height=1.5cm,trim={{.6\width} {.8\height} {.3\width} {.05\height}},clip,cfbox=White 0.25mm -0.25mm]{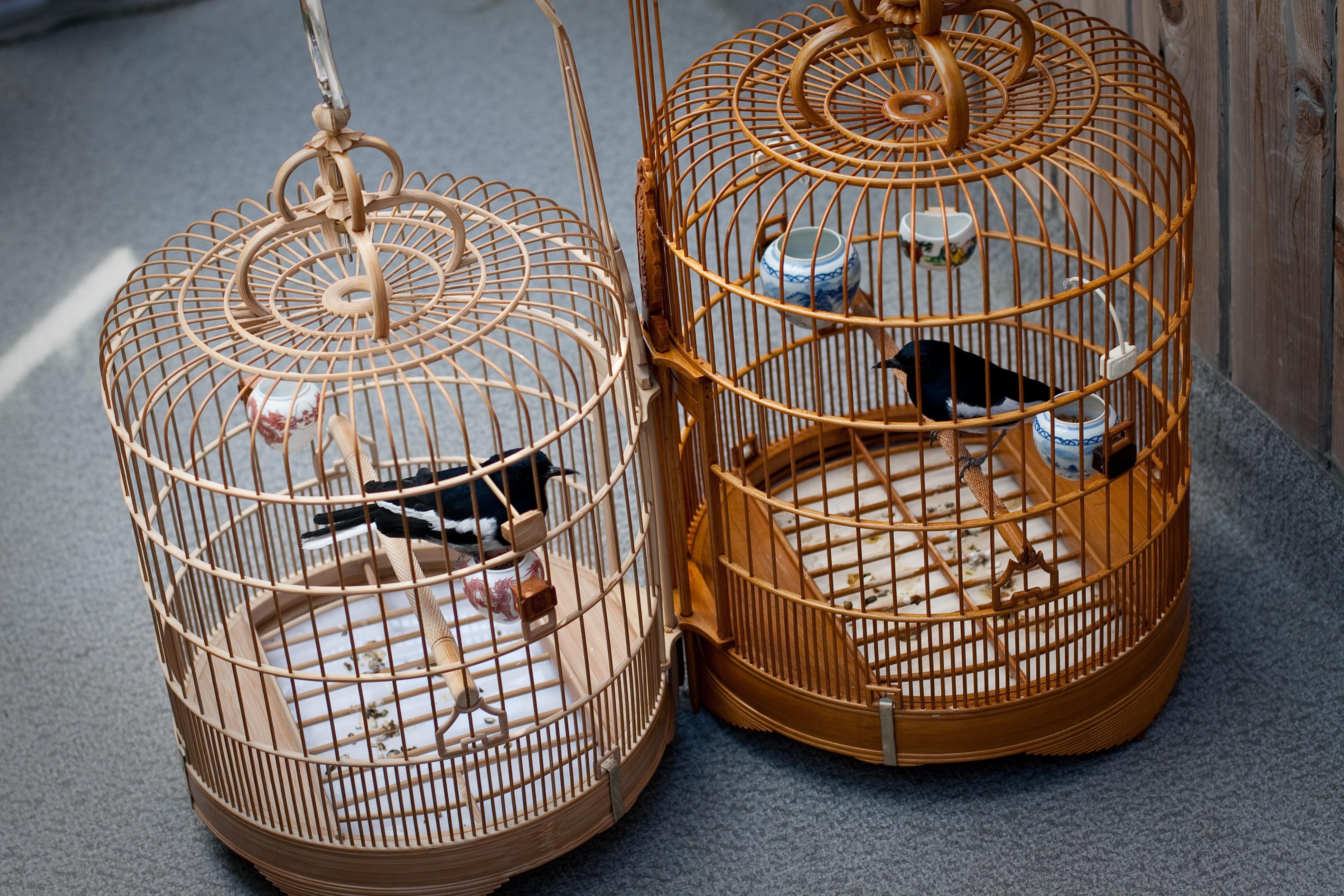}}{
            \begin{tikzpicture}
                \node[anchor=south west,inner sep=0] (image) at (0,0) {\includegraphics[height=0.2\textwidth]{fig/dis5k/4892828841_7f1bc05682_o.jpg}};
                \begin{scope}[x={(image.south east)},y={(image.north west)}]
                \draw[white,thick] (0.6,0.8) rectangle (0.7,0.95);
                \end{scope}
            \end{tikzpicture}
        }&
        \stackinset{l}{3pt}{t}{0.5pt}{\adjincludegraphics[height=2cm,trim={{.8\width} {.2\height} {.02\width} {.5\height}},clip,cfbox=White 0.25mm -0.25mm]{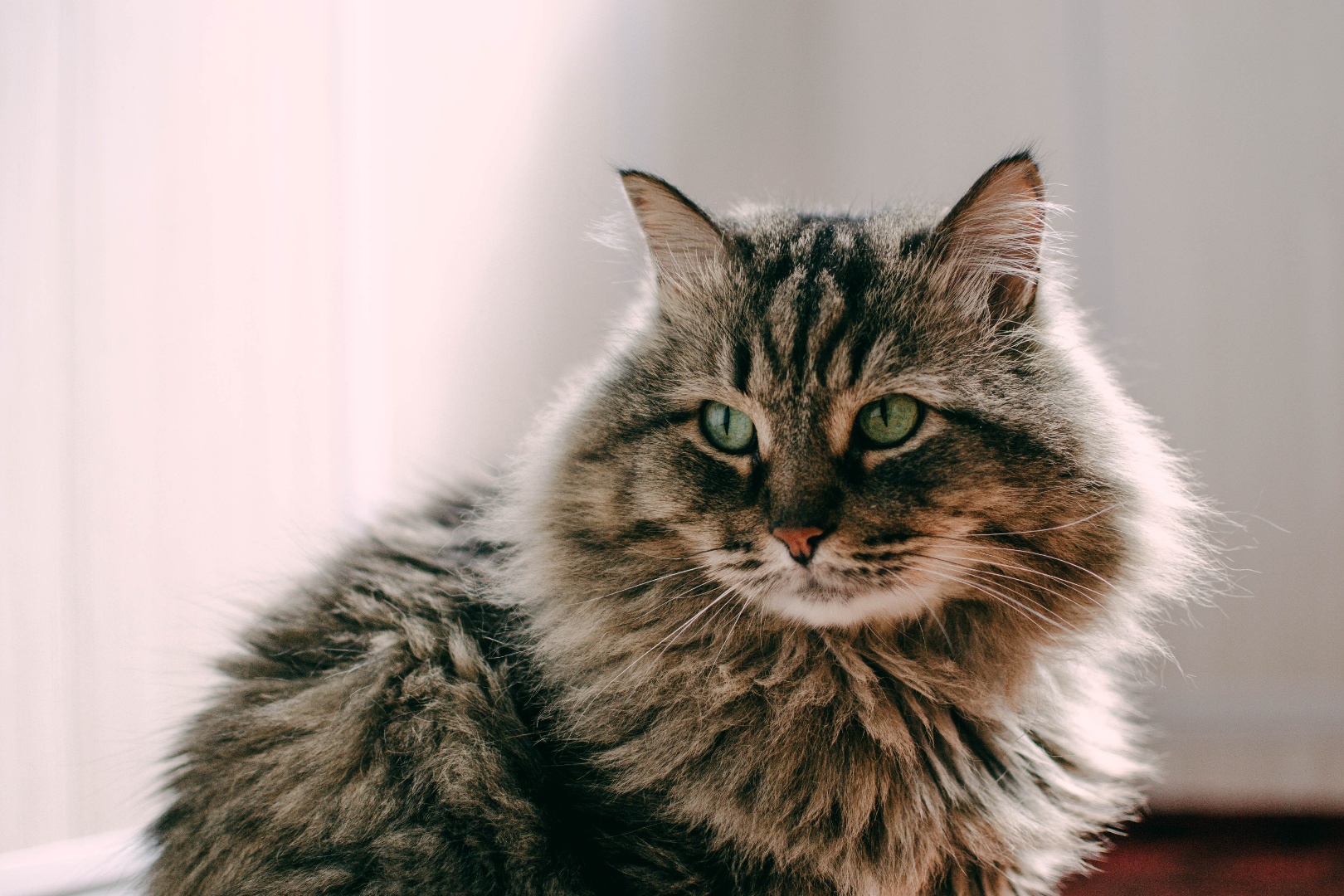}}{
            \begin{tikzpicture}
                \node[anchor=south west,inner sep=0] (image) at (0,0) {\includegraphics[height=0.2\textwidth]{fig/am2k/m_75b2ab26.jpg}};
                \begin{scope}[x={(image.south east)},y={(image.north west)}]
                    \draw[white,thick] (0.8,0.2) rectangle (0.98,0.5);
                \end{scope}
            \end{tikzpicture}
        }\\
        \raisebox{1.3cm}[0pt][0pt]{\rotatebox[origin=c]{90}{\footnotesize Depth Pro}}&
        \stackinset{l}{3pt}{t}{0.5pt}{\adjincludegraphics[height=2cm,trim={{.35\width} {.35\height} {.6\width} {.5\height}},clip,cfbox=White 0.25mm -0.25mm]{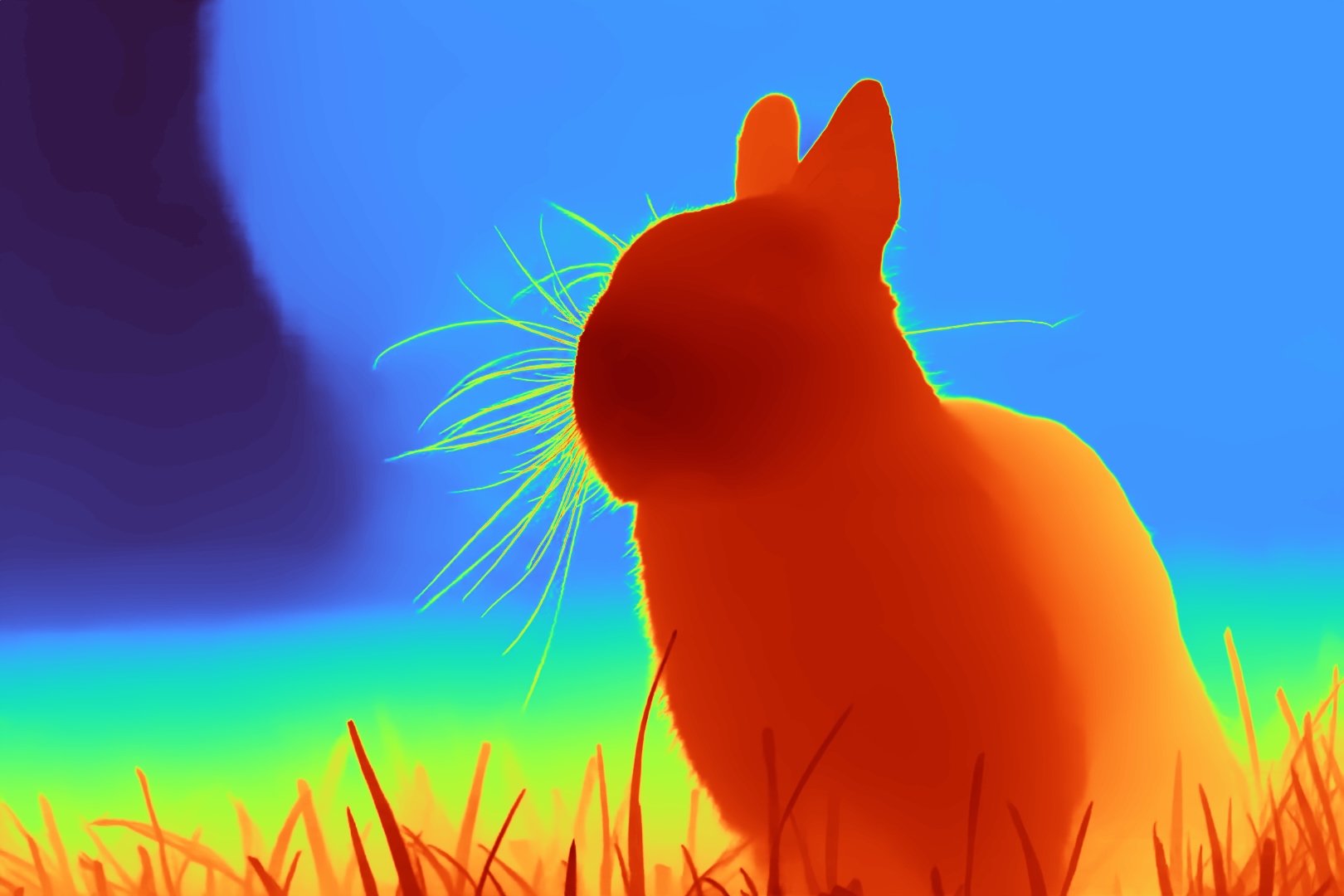}}{
            \begin{tikzpicture}
                \node[anchor=south west,inner sep=0] (image) at (0,0) {\begin{overpic}[height=0.2\textwidth]{fig/am2k/m_0aa9dd6a_depthpro_pv5i9yzjef.jpg}
                % \put (2,2) {\textcolor{white}{Depth Pro}}
                \end{overpic}
                };
                \begin{scope}[x={(image.south east)},y={(image.north west)}]
                \draw[white,thick] (0.35,0.35) rectangle (0.4,0.5);
                \end{scope}
            \end{tikzpicture}
        }&
        \stackinset{r}{3pt}{b}{0.5pt}{\adjincludegraphics[height=1.5cm,trim={{.6\width} {.8\height} {.3\width} {.05\height}},clip,cfbox=White 0.25mm -0.25mm]{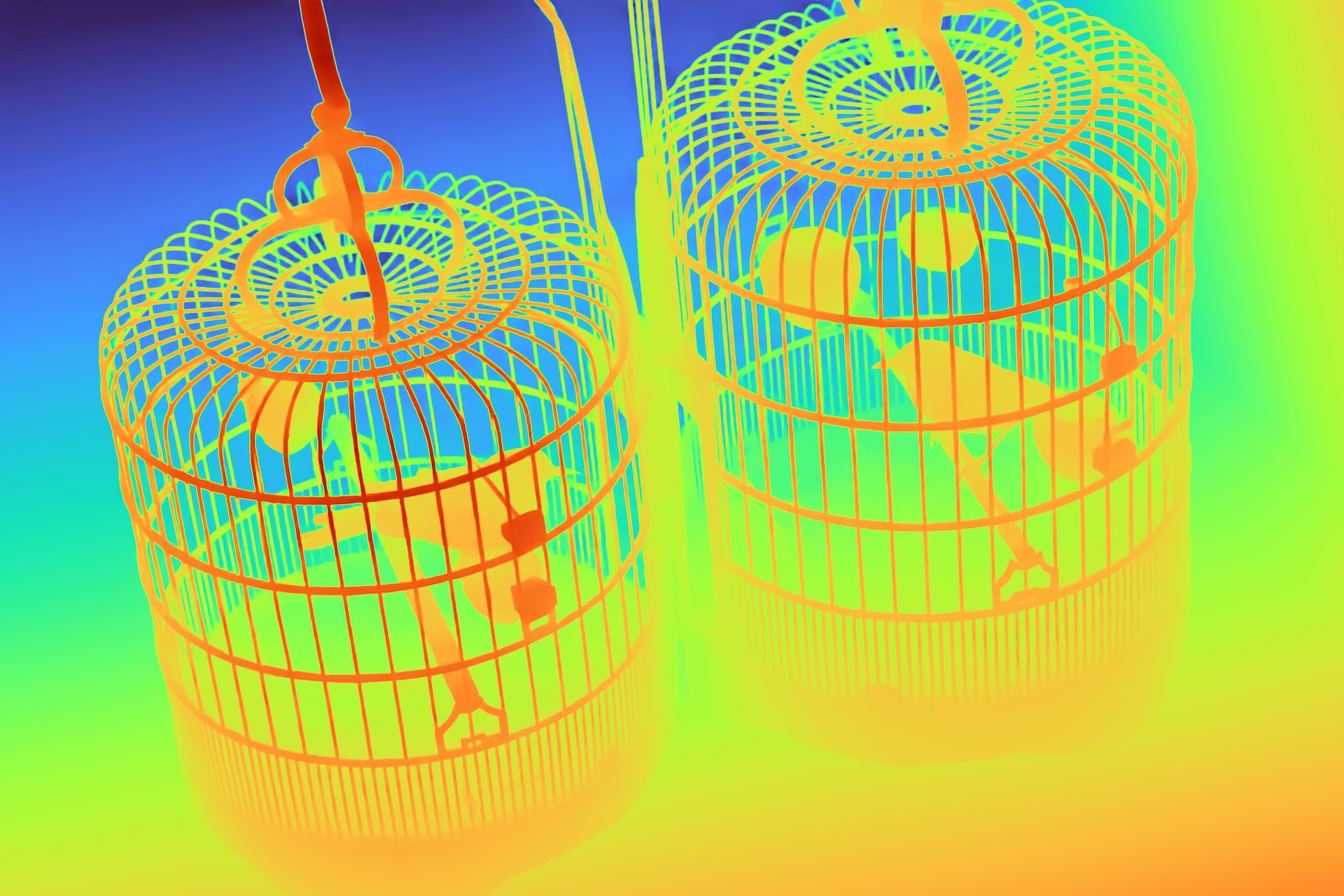}}{
            \begin{tikzpicture}
                \node[anchor=south west,inner sep=0] (image) at (0,0) {\includegraphics[height=0.2\textwidth]{fig/dis5k/4892828841_7f1bc05682_o_depthpro_pv5i9yzjef.jpg}};
                \begin{scope}[x={(image.south east)},y={(image.north west)}]
                \draw[white,thick] (0.6,0.8) rectangle (0.7,0.95);
                \end{scope}
            \end{tikzpicture}
        }&
        \stackinset{l}{3pt}{t}{0.5pt}{\adjincludegraphics[height=2cm,trim={{.8\width} {.2\height} {.02\width} {.5\height}},clip,cfbox=White 0.25mm -0.25mm]{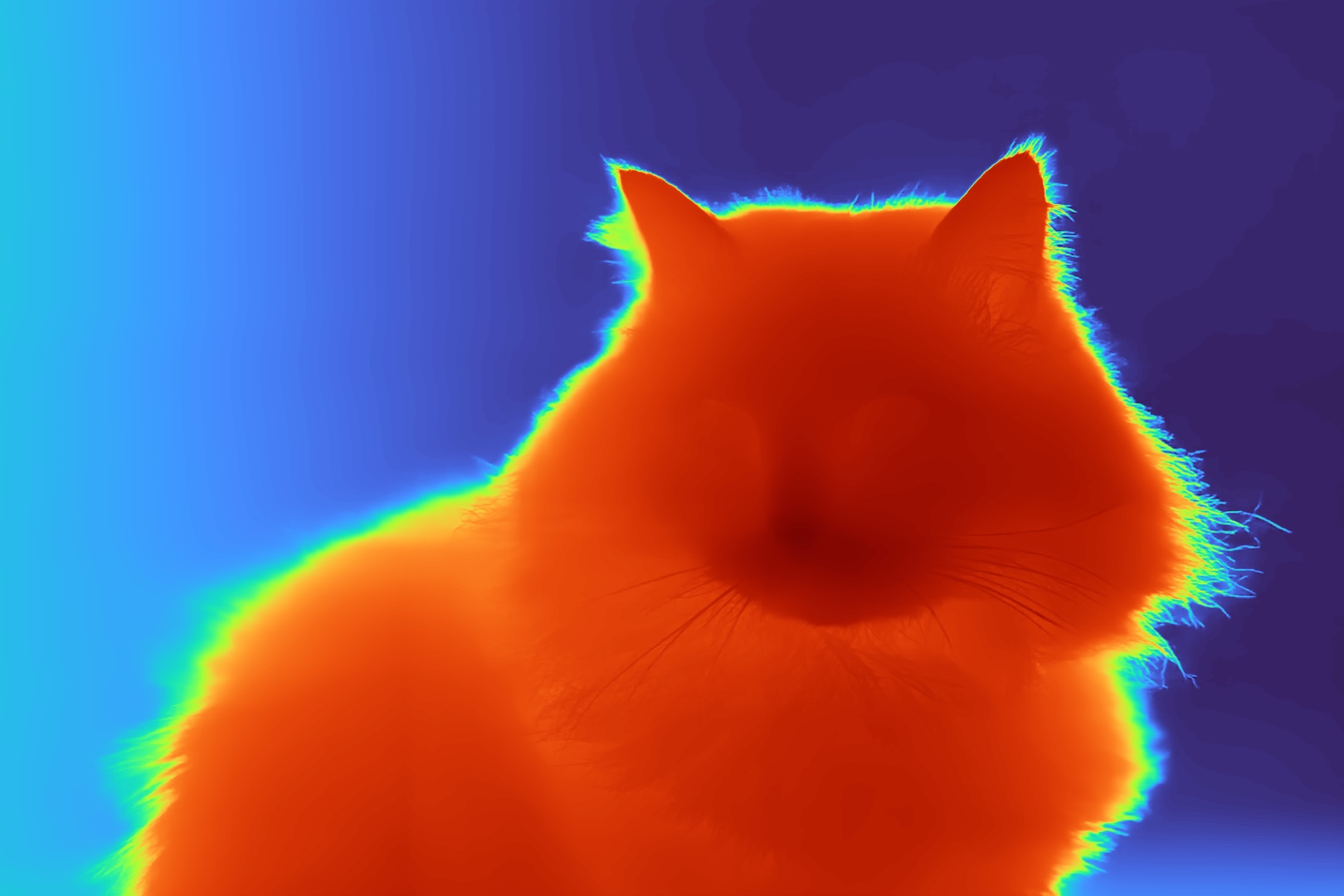}}{
            \begin{tikzpicture}
                \node[anchor=south west,inner sep=0] (image) at (0,0) {\includegraphics[height=0.2\textwidth]{fig/am2k/m_75b2ab26_depthpro_pv5i9yzjef.jpg}};
                \begin{scope}[x={(image.south east)},y={(image.north west)}]
                \draw[white,thick] (0.8,0.2) rectangle (0.98,0.5);
                \end{scope}
            \end{tikzpicture}
        }\\
        \raisebox{1.3cm}[0pt][0pt]{\rotatebox[origin=c]{90}{\footnotesize Marigold}}&
        \stackinset{l}{3pt}{t}{0.5pt}{\adjincludegraphics[height=2cm,trim={{.35\width} {.35\height} {.6\width} {.5\height}},clip,cfbox=White 0.25mm -0.25mm]{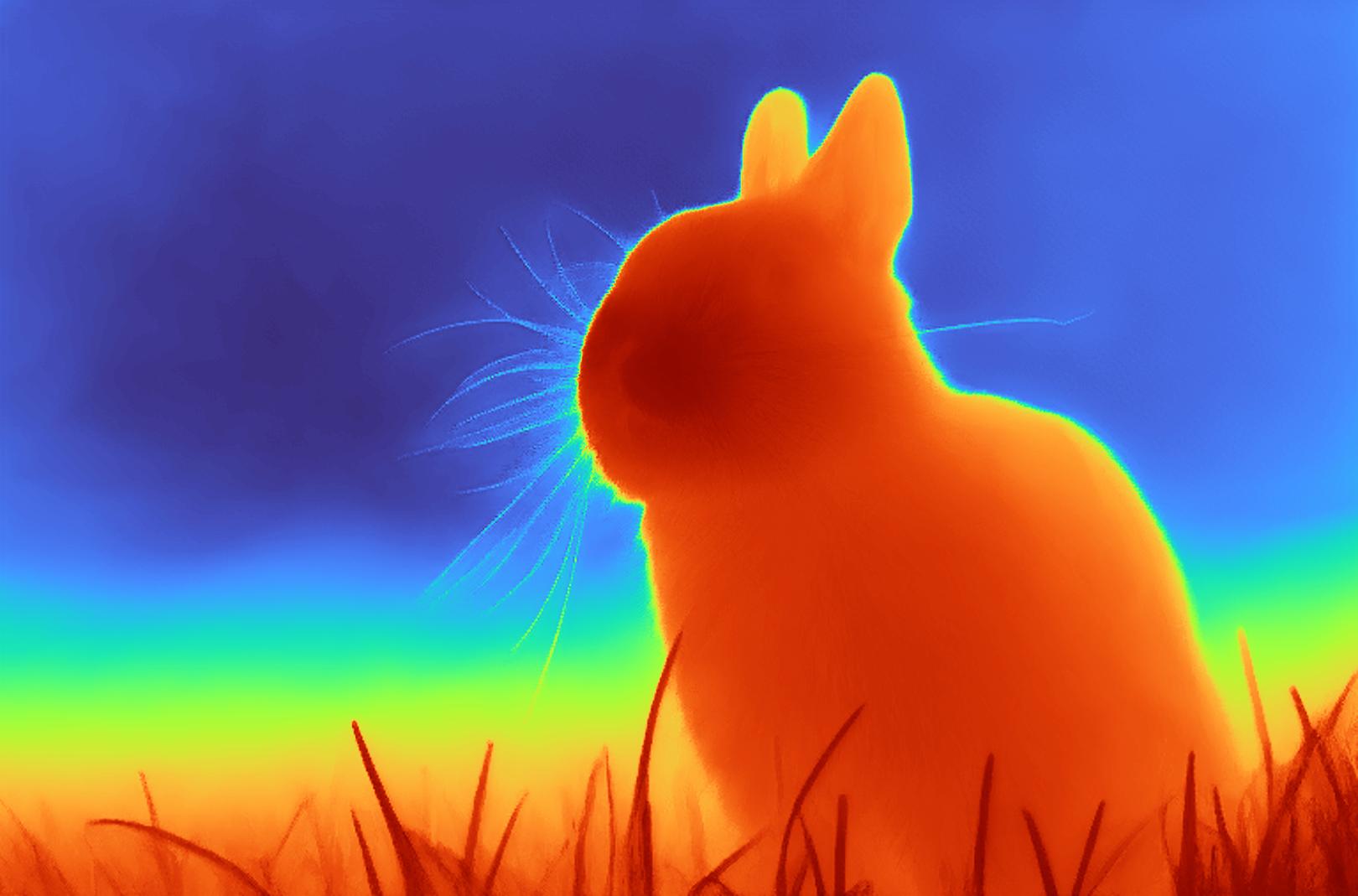}}{
            \begin{tikzpicture}
                \node[anchor=south west,inner sep=0] (image) at (0,0) {\begin{overpic}[height=0.2\textwidth]{fig/am2k/m_0aa9dd6a_marigold.jpg}
                % \put (2,2) {\textcolor{white}{Marigold}}
                \end{overpic}};
                \begin{scope}[x={(image.south east)},y={(image.north west)}]
                \draw[white,thick] (0.35,0.35) rectangle (0.4,0.5);
                \end{scope}
            \end{tikzpicture}
        }&
        \stackinset{r}{3pt}{b}{0.5pt}{\adjincludegraphics[height=1.5cm,trim={{.6\width} {.8\height} {.3\width} {.05\height}},clip,cfbox=White 0.25mm -0.25mm]{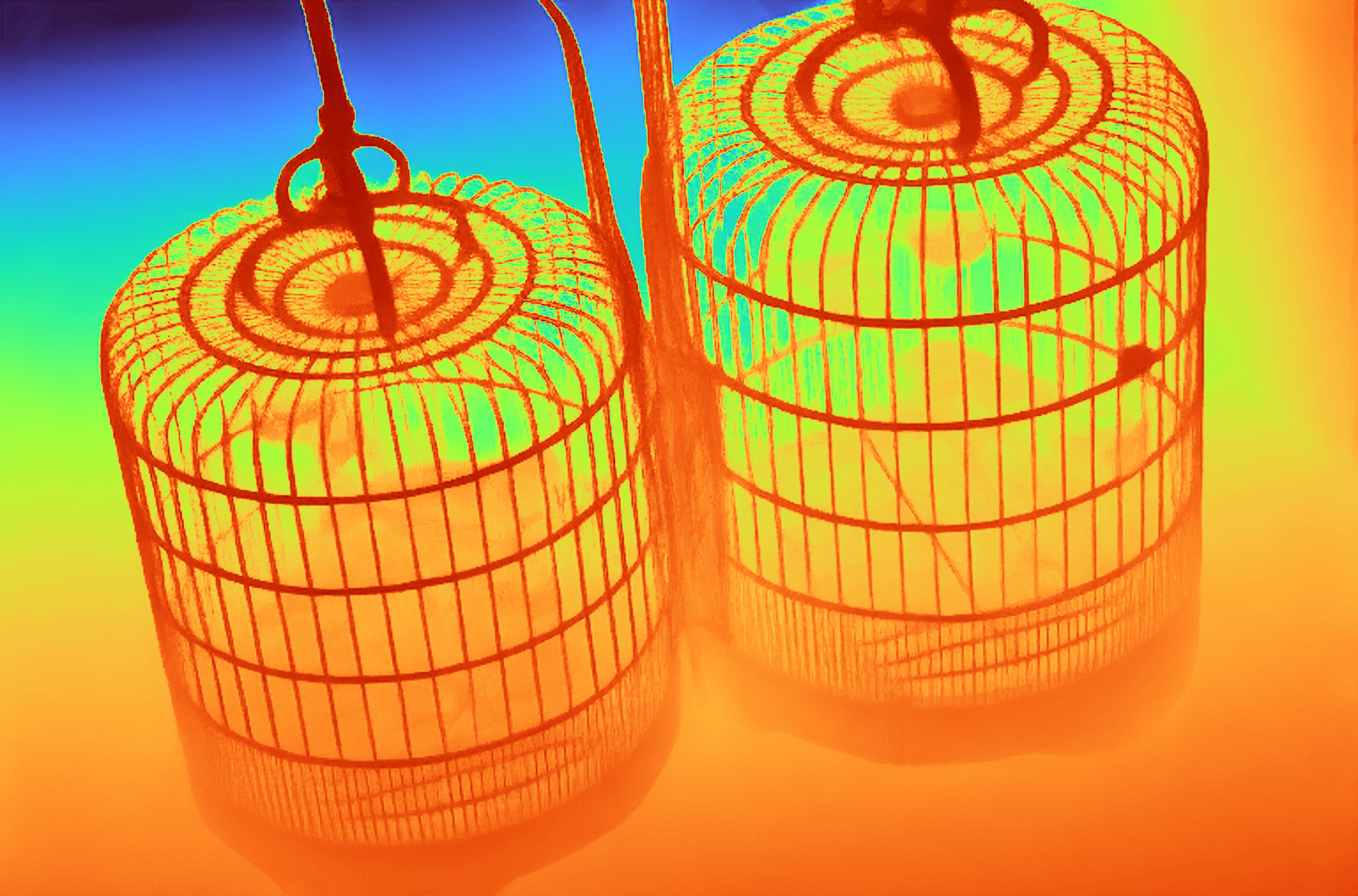}}{
            \begin{tikzpicture}
                \node[anchor=south west,inner sep=0] (image) at (0,0) {\includegraphics[height=0.2\textwidth]{fig/dis5k/4892828841_7f1bc05682_o_marigold.jpg}};
                \begin{scope}[x={(image.south east)},y={(image.north west)}]
                \draw[white,thick] (0.6,0.8) rectangle (0.7,0.95);
                \end{scope}
            \end{tikzpicture}
        }&
        \stackinset{l}{3pt}{t}{0.5pt}{\adjincludegraphics[height=2cm,trim={{.8\width} {.2\height} {.02\width} {.5\height}},clip,cfbox=White 0.25mm -0.25mm]{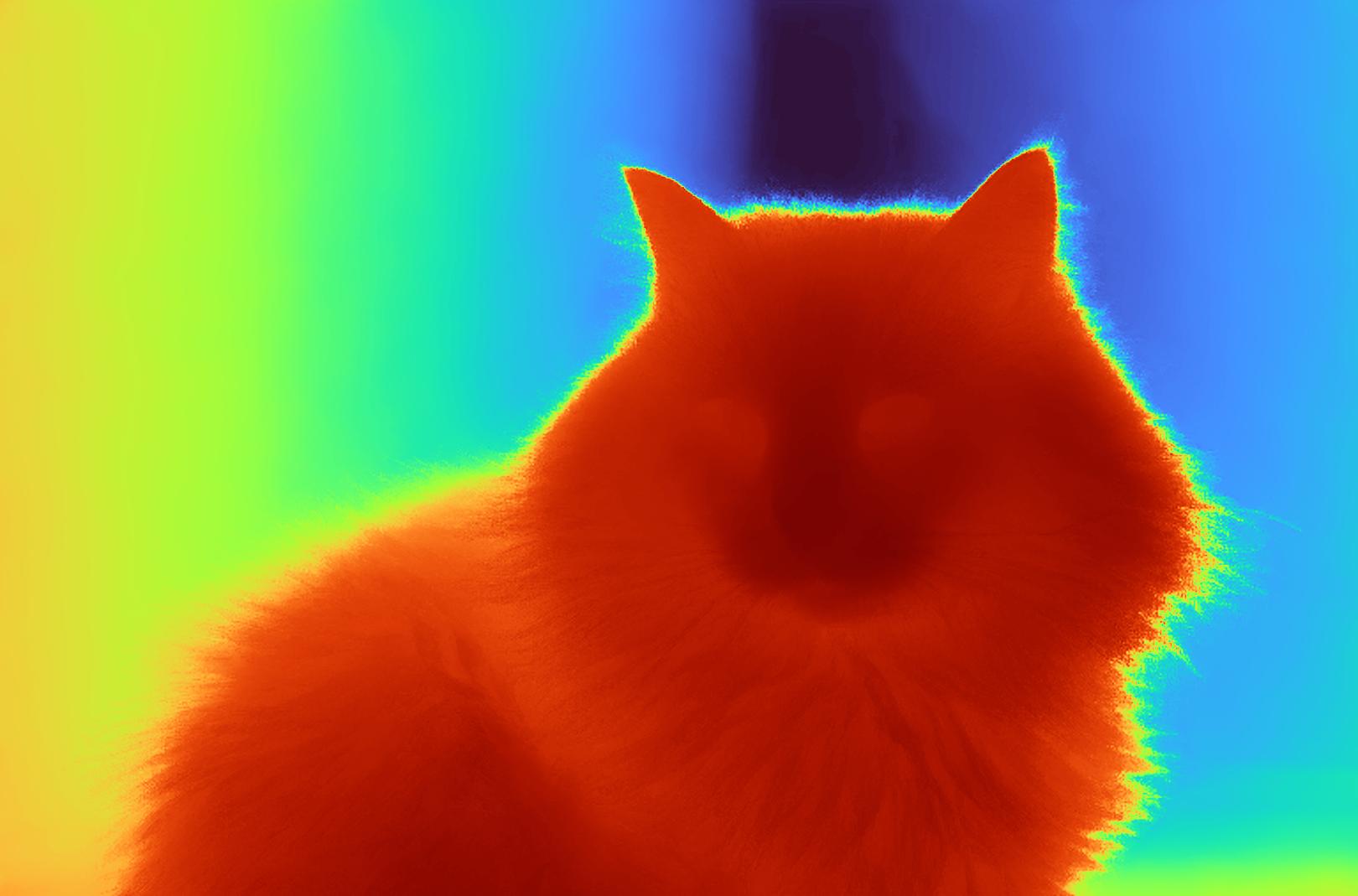}}{
            \begin{tikzpicture}
                \node[anchor=south west,inner sep=0] (image) at (0,0) {\includegraphics[height=0.2\textwidth]{fig/am2k/m_75b2ab26_marigold.jpg}};
                \begin{scope}[x={(image.south east)},y={(image.north west)}]
                \draw[white,thick] (0.8,0.2) rectangle (0.98,0.5);
                \end{scope}
            \end{tikzpicture}
        }\\
        \raisebox{1.3cm}[0pt][0pt]{\rotatebox[origin=c]{90}{\footnotesize Depth Anything v2}}&
        \stackinset{l}{3pt}{t}{0.5pt}{\adjincludegraphics[height=2cm,trim={{.35\width} {.35\height} {.6\width} {.5\height}},clip,cfbox=White 0.25mm -0.25mm]{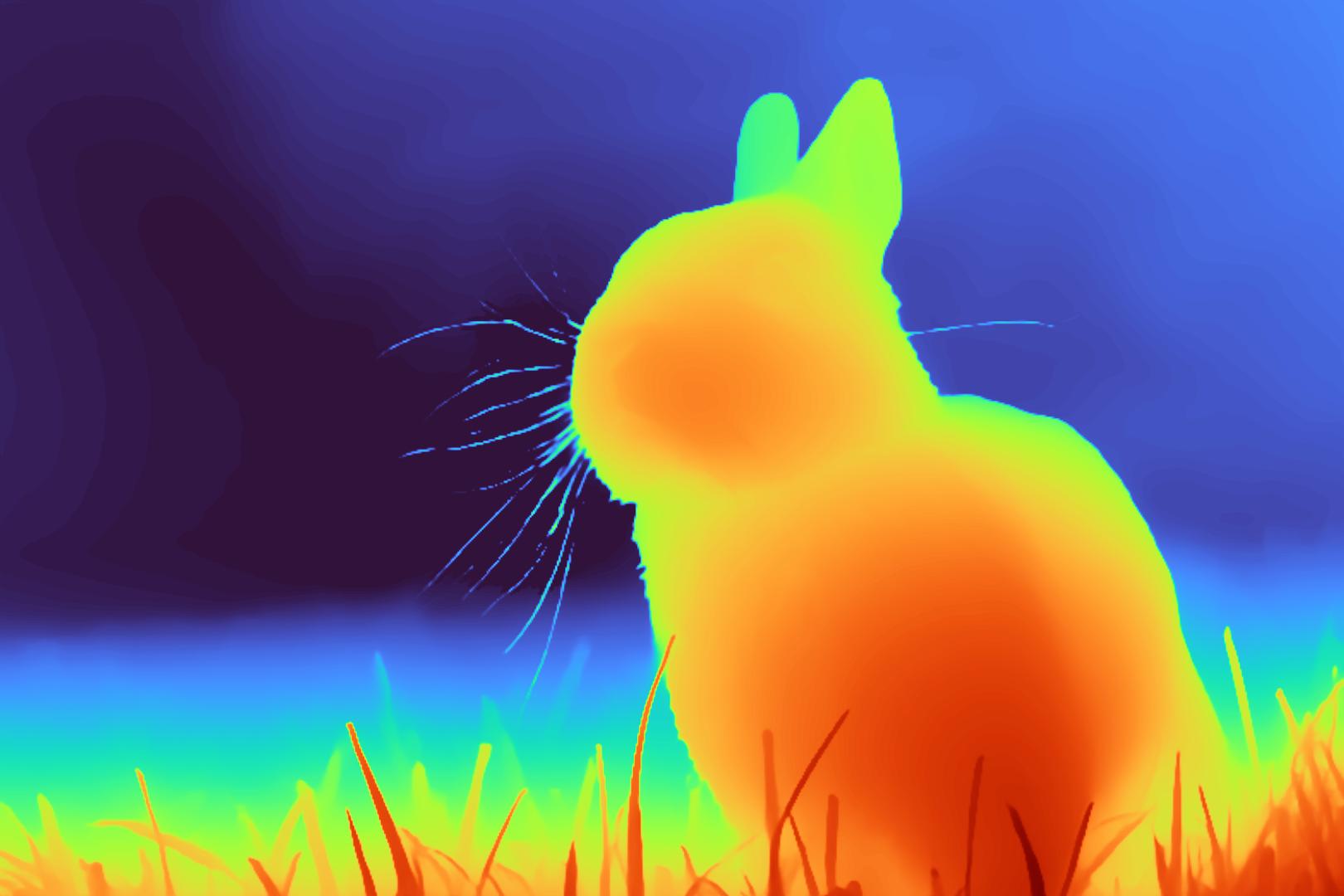}}{
            \begin{tikzpicture}
                \node[anchor=south west,inner sep=0] (image) at (0,0) {\begin{overpic}[height=0.2\textwidth]{fig/am2k/m_0aa9dd6a_depth_anything_v2_relative.jpg}
                % \put (2,2) {\textcolor{white}{Depth Anything v2}}
                \end{overpic}};
                \begin{scope}[x={(image.south east)},y={(image.north west)}]
                \draw[white,thick] (0.35,0.35) rectangle (0.4,0.5);
                \end{scope}
            \end{tikzpicture}
        }&
        \stackinset{r}{3pt}{b}{0.5pt}{\adjincludegraphics[height=1.5cm,trim={{.6\width} {.8\height} {.3\width} {.05\height}},clip,cfbox=White 0.25mm -0.25mm]{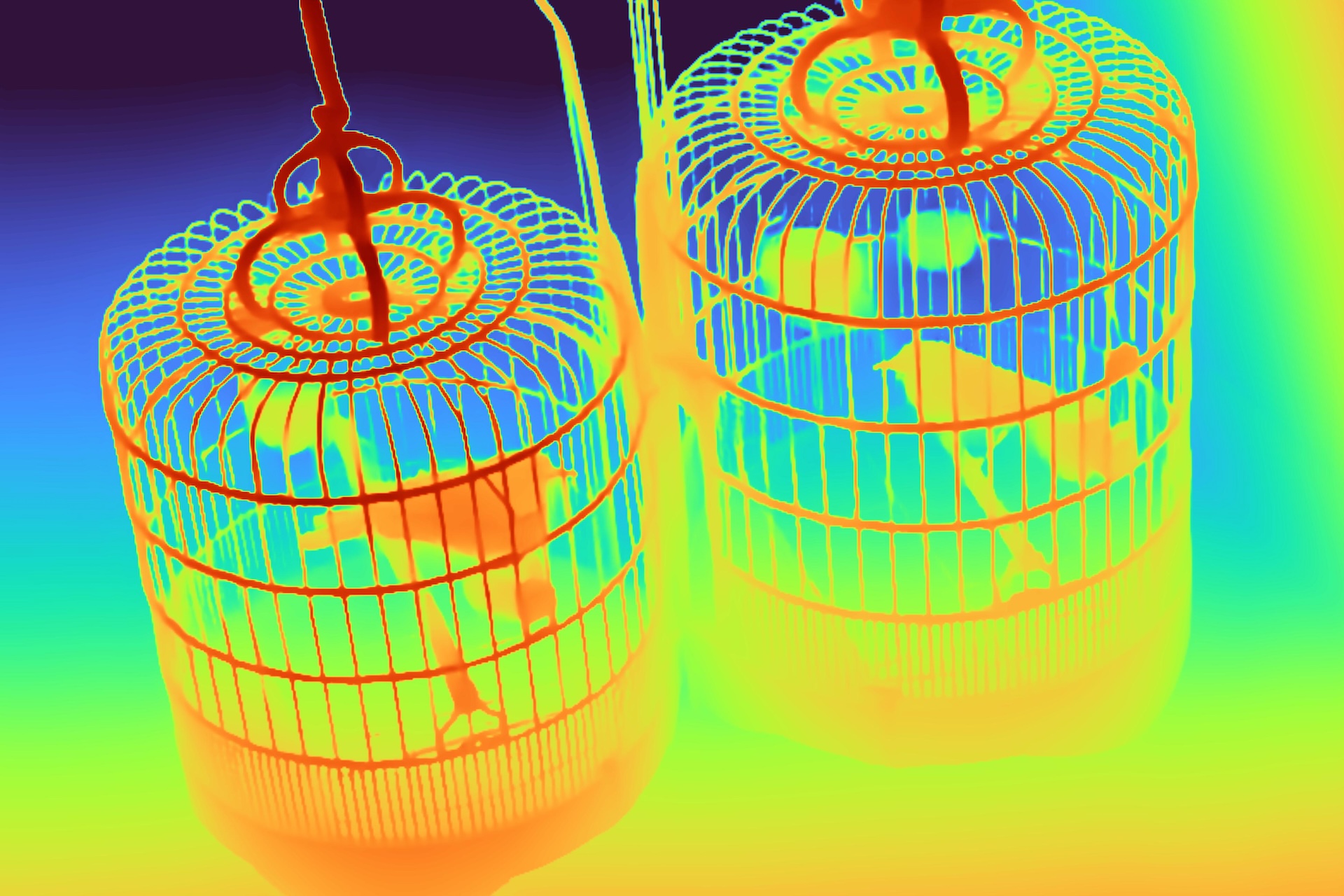}}{
            \begin{tikzpicture}
                \node[anchor=south west,inner sep=0] (image) at (0,0) {\includegraphics[height=0.2\textwidth]{fig/dis5k/4892828841_7f1bc05682_o_depth_anything_v2_relative.jpg}};
                \begin{scope}[x={(image.south east)},y={(image.north west)}]
                \draw[white,thick] (0.6,0.8) rectangle (0.7,0.95);
                \end{scope}
            \end{tikzpicture}
        }&
        \stackinset{l}{3pt}{t}{0.5pt}{\adjincludegraphics[height=2cm,trim={{.8\width} {.2\height} {.02\width} {.5\height}},clip,cfbox=White 0.25mm -0.25mm]{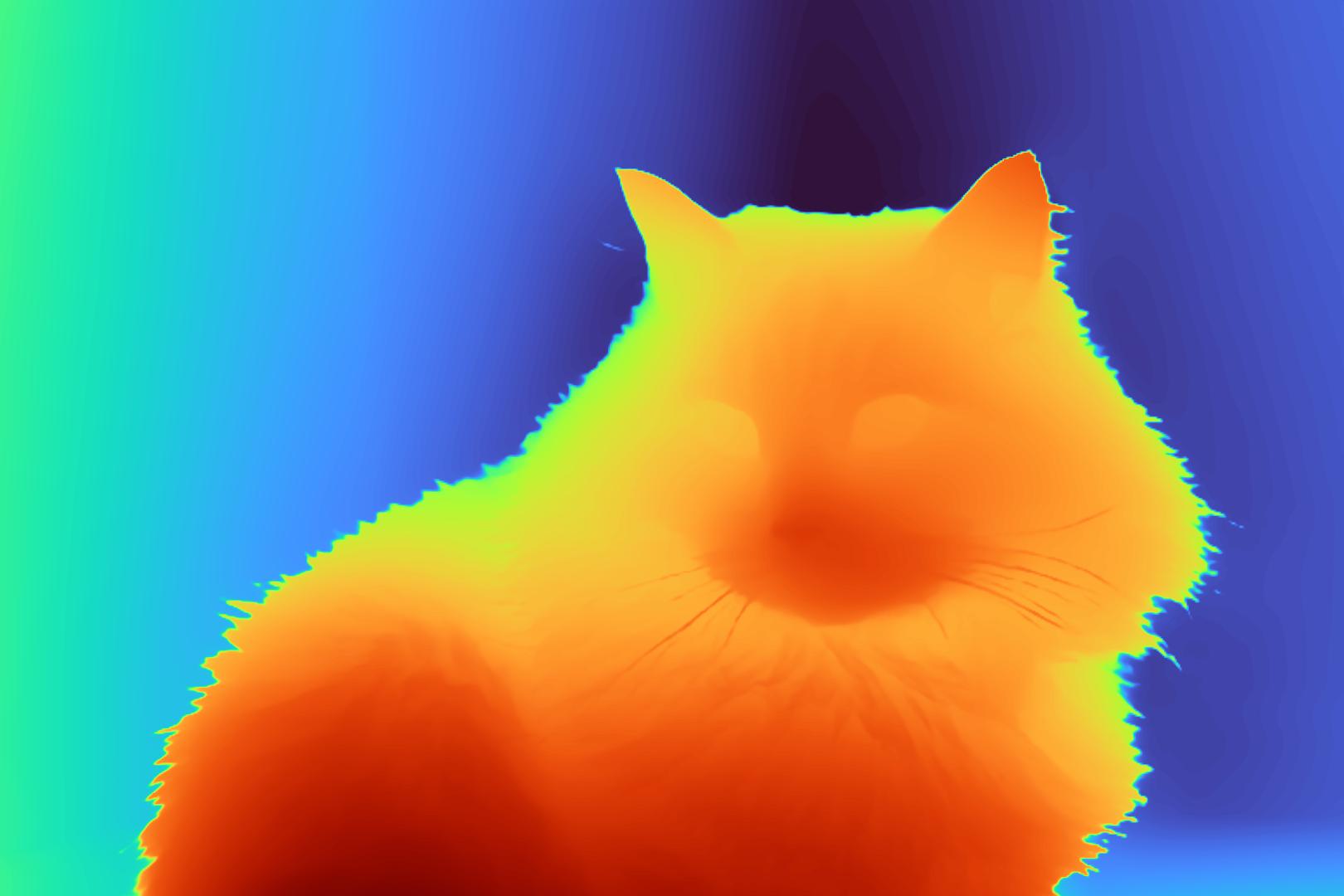}}{
            \begin{tikzpicture}
                \node[anchor=south west,inner sep=0] (image) at (0,0) {\includegraphics[height=0.2\textwidth]{fig/am2k/m_75b2ab26_depth_anything_v2_relative.jpg}};
                \begin{scope}[x={(image.south east)},y={(image.north west)}]
                \draw[white,thick] (0.8,0.2) rectangle (0.98,0.5);
                \end{scope}
            \end{tikzpicture}
        }\\
        \raisebox{1.3cm}[0pt][0pt]{\rotatebox[origin=c]{90}{\footnotesize Metric3D v2}}&
        \stackinset{l}{3pt}{t}{0.5pt}{\adjincludegraphics[height=2cm,trim={{.35\width} {.35\height} {.6\width} {.5\height}},clip,cfbox=White 0.25mm -0.25mm]{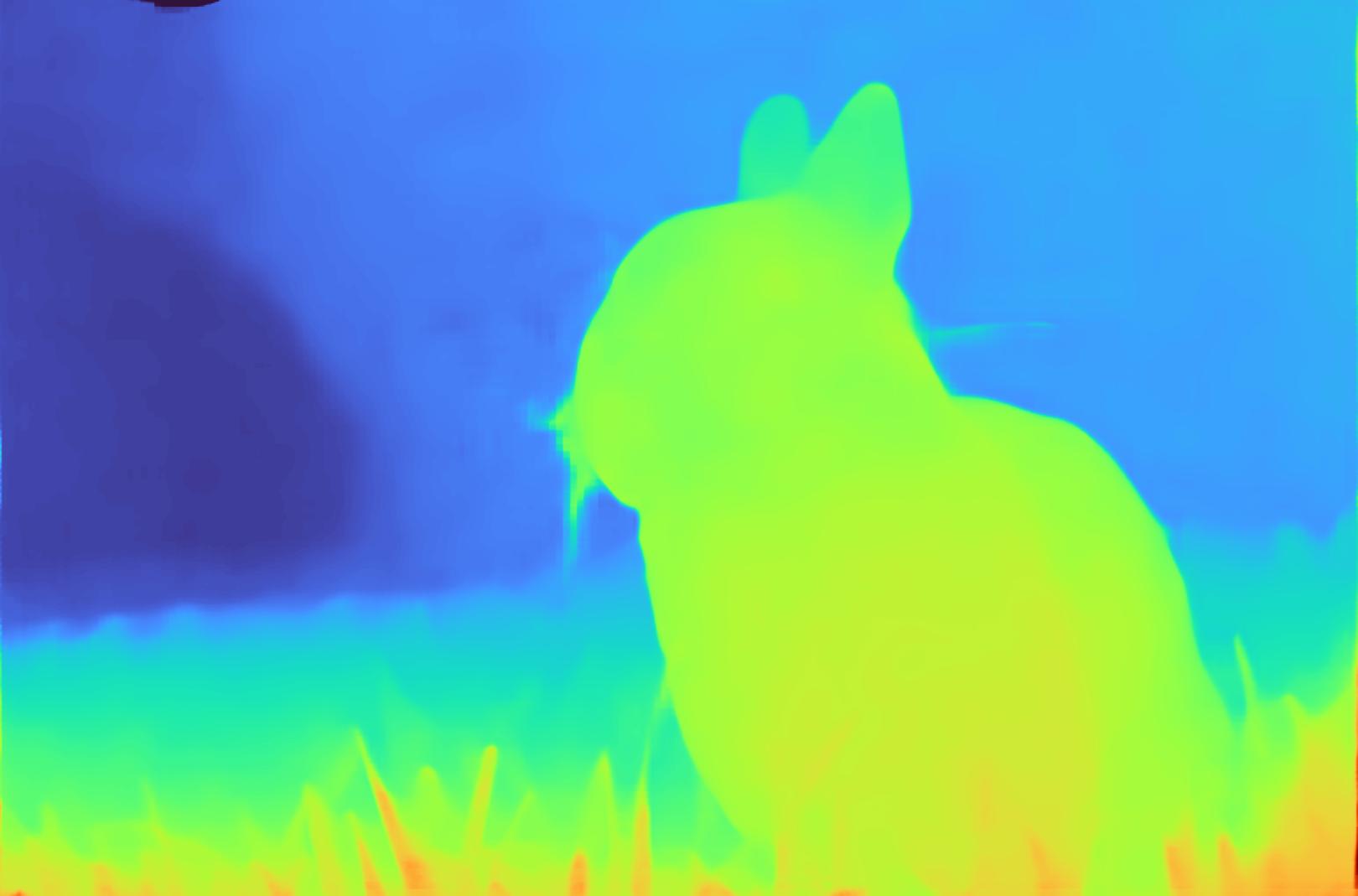}}{
            \begin{tikzpicture}
                \node[anchor=south west,inner sep=0] (image) at (0,0) {\begin{overpic}[height=0.2\textwidth]{fig/am2k/m_0aa9dd6a_metric3d_v2g.jpg}
                % \put (2,2) {\textcolor{white}{Metric3D v2}}
                \end{overpic}
                };
                \begin{scope}[x={(image.south east)},y={(image.north west)}]
                \draw[white,thick] (0.35,0.35) rectangle (0.4,0.5);
                \end{scope}
            \end{tikzpicture}
        }&
        \stackinset{r}{3pt}{b}{0.5pt}{\adjincludegraphics[height=1.5cm,trim={{.6\width} {.8\height} {.3\width} {.05\height}},clip,cfbox=White 0.25mm -0.25mm]{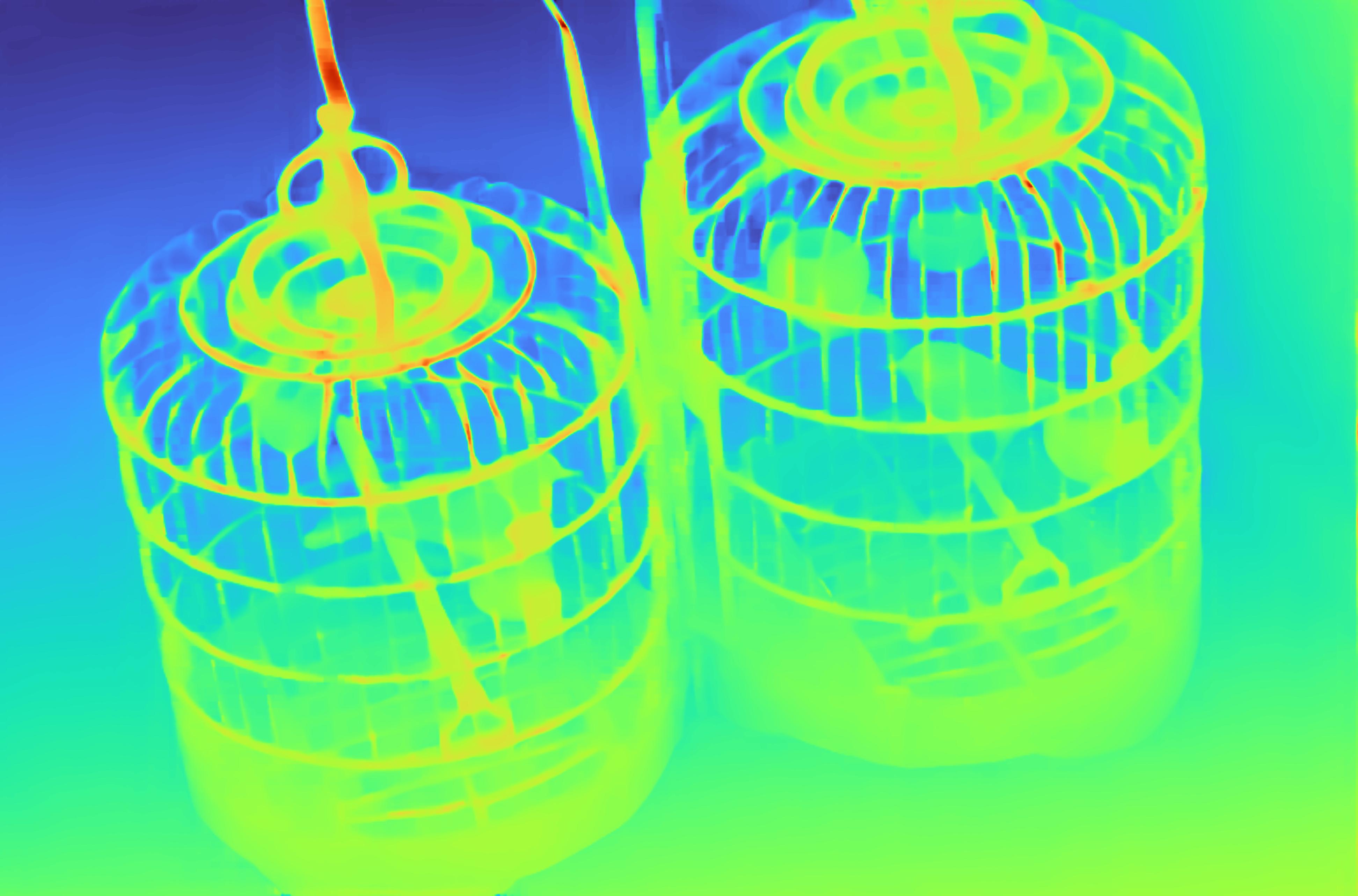}}{
            \begin{tikzpicture}
                \node[anchor=south west,inner sep=0] (image) at (0,0) {\includegraphics[height=0.2\textwidth]{fig/dis5k/4892828841_7f1bc05682_o_metric3d_v2g.jpg}};
                \begin{scope}[x={(image.south east)},y={(image.north west)}]
                \draw[white,thick] (0.6,0.8) rectangle (0.7,0.95);
                \end{scope}
            \end{tikzpicture}
        }&
        \stackinset{l}{3pt}{t}{0.5pt}{\adjincludegraphics[height=2cm,trim={{.8\width} {.2\height} {.02\width} {.5\height}},clip,cfbox=White 0.25mm -0.25mm]{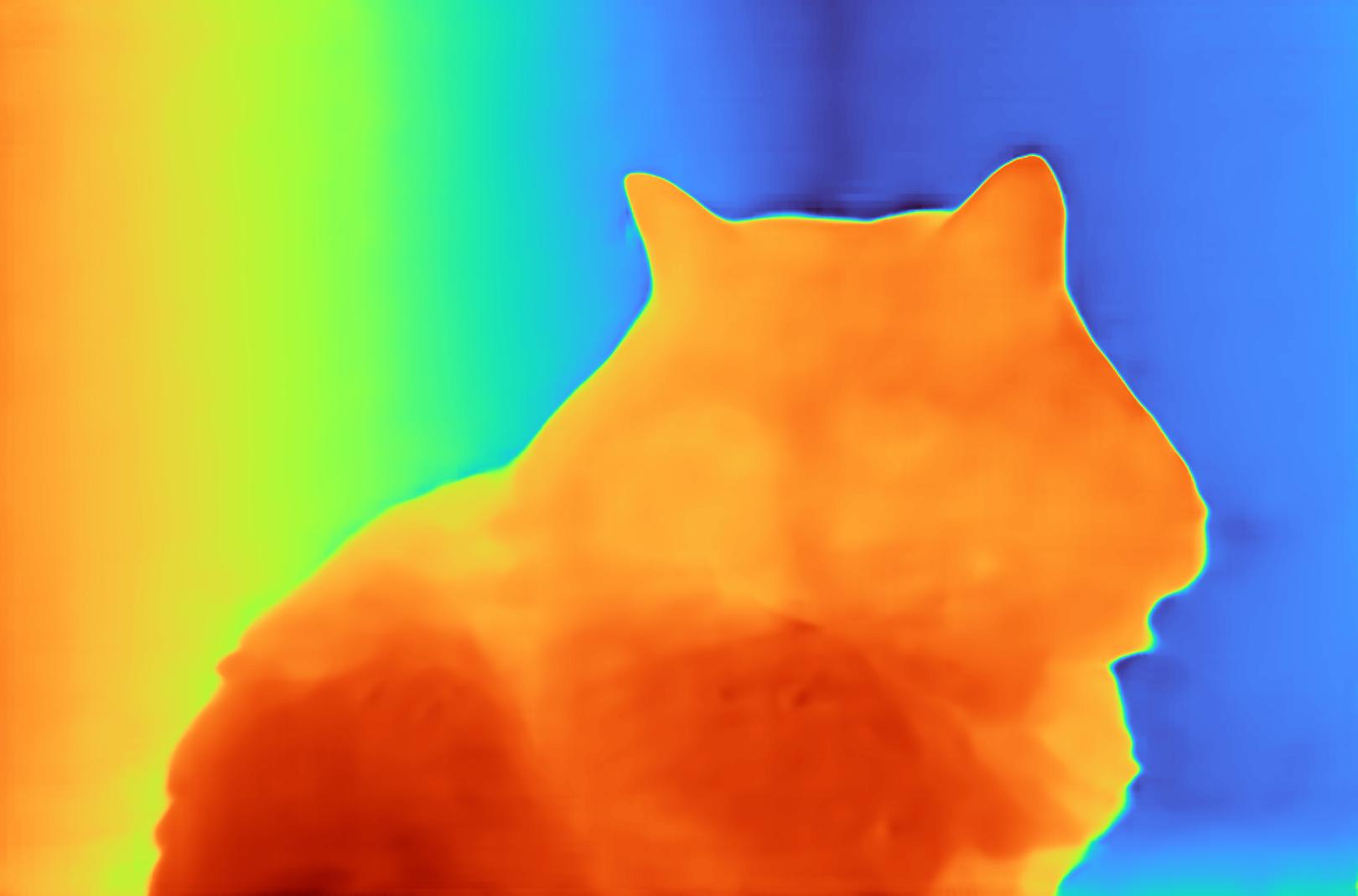}}{
            \begin{tikzpicture}
                \node[anchor=south west,inner sep=0] (image) at (0,0) {\includegraphics[height=0.2\textwidth]{fig/am2k/m_75b2ab26_metric3d_v2g.jpg}};
                \begin{scope}[x={(image.south east)},y={(image.north west)}]
                \draw[white,thick] (0.8,0.2) rectangle (0.98,0.5);
                \end{scope}
            \end{tikzpicture}
        }\\
    \end{tabular}
    \vspace{-2mm}
   \caption{Results on images from the AM-2k~\citep{Li2022:IJCV} (1st \& 3rd column) and DIS-5k~\citep{Qin2022:ECCV} (2nd column) datasets. Input image on top, estimated depth maps from Depth Pro, Marigold~\citep{Ke2024:CVPR}, Depth Anything v2~\citep{Yang2024:arxiv}, and Metric3D v2~\citep{Hu2024:arxiv} below. Depth Pro produces zero-shot metric depth maps with absolute scale at 2.25-megapixel native resolution in 0.3 seconds on a V100 GPU.}
  \label{fig:teaser}
  \vspace{-5mm}
\end{figure}

Zero-shot monocular depth estimation underpins a growing variety of applications, such as advanced image editing, view synthesis, and conditional image generation. Inspired by MiDaS~\citep{Ranftl2022:TPAMI} and many follow-up works~\citep{Ranftl2021:ICCV,Ke2024:CVPR,Yang2024:CVPR,Piccinelli2024:CVPR,Hu2024:arxiv}, applications increasingly leverage the ability to derive a dense pixelwise depth map for any image.

Our work is motivated in particular by novel view synthesis from a single image, an exciting application that has been transformed by advances in monocular depth estimation~\citep{Hedman2017:SIGGRAPH, Shih2020:CVPR, Jampani2021:ICCV, Khan2023:ICCV}.
Applications such as view synthesis imply a number of desiderata for monocular depth estimation. First, the depth estimator should work zero-shot on any image, not restricted to a specific domain~\citep{Ranftl2022:TPAMI,Yang2024:CVPR}. Furthermore, the method should ideally produce \emph{metric} depth maps in this zero-shot regime, to accurately reproduce object shapes, scene layouts, and absolute scales~\citep{Guizilini2023:ICCV,Hu2024:arxiv}. For the broadest applicability `in the wild', the method should produce metric depth maps with absolute scale even if no camera intrinsics (such as focal length) are provided with the image~\citep{Piccinelli2024:CVPR}. This enables view synthesis scenarios such as ``Synthesize a view of this scene from 63 mm away'' for essentially arbitrary single images~\citep{Dodgson2004}.

Second, for the most compelling results, the monocular depth estimator should operate at high resolution and produce fine-grained depth maps that closely adhere to image details such as hair, fur, and other fine structures~\citep{Miangoleh2021:CVPR,Ke2024:CVPR,Li2024:CVPR}.
One benefit of producing sharp depth maps that accurately trace intricate details is the elimination of ``flying pixels''% near boundaries
, which can degrade image quality in applications such as view synthesis~\citep{Jampani2021:ICCV}.

Third, for many interactive application scenarios, the depth estimator should operate at low latency, processing a high-resolution image in less than a second to support interactive view synthesis ``queries'' on demand. Low latency is a common characteristic of methods that reduce zero-shot monocular depth estimation to a single forward pass through a neural network \citep{Ranftl2021:ICCV,Yang2024:CVPR,Piccinelli2024:CVPR}, \, but it is not always shared by methods that employ more computationally demanding machinery at test time~\citep{Ke2024:CVPR,Li2024:CVPR}.

\begin{wrapfigure}{r}[0pt]{0.5\linewidth}
\centering\scriptsize
\vspace{-3mm}
% \vspace{10cm}
%
\includegraphics[width=0.70\linewidth]{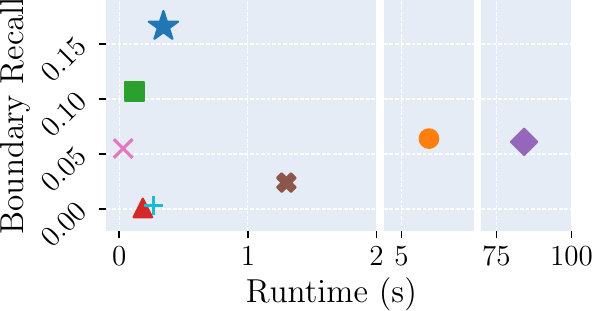}
\raisebox{0.48\height}{\includegraphics[width=0.25\linewidth]{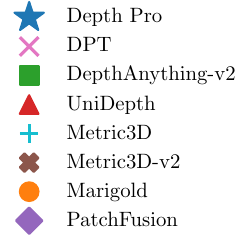}}
\vspace{-1mm}
\caption{Boundary recall versus runtime. Depth Pro outperforms prior work by a multiplicative factor in boundary accuracy while being orders of magnitude faster than works focusing on fine-grained predictions (\eg Marigold, PatchFusion).}
\label{fig:am_vs_runtime}
\vspace{-3ex}
\end{wrapfigure}

In this work, we present a foundation model for zero-shot metric monocular depth estimation that meets all of these desiderata. Our model, Depth Pro, produces metric depth maps with absolute scale on arbitrary images `in the wild' without requiring metadata such as camera intrinsics. It operates at high resolution, producing 2.25-megapixel depth maps (with a native output resolution of $1536 \times 1536$ before optional further upsampling) in 0.3 seconds on a V100 GPU. \Fig~\ref{fig:teaser} shows some representative results. Depth Pro dramatically outperforms all prior work in sharp delineation of object boundaries, including fine structures such as hair, fur, and vegetation. As shown in \Fig~\ref{fig:am_vs_runtime}, Depth Pro offers unparalleled boundary tracing, outperforming all prior work by a multiplicative factor in boundary recall. (See \Sec~\ref{sec:experiments} for additional detail.) Compared to the prior state of the art in boundary accuracy \citep{Ke2024:CVPR,Li2024:CVPR}, Depth Pro is one to two orders of magnitude faster, yields much more accurate boundaries, and provides metric depth maps with absolute scale.

Depth Pro is enabled by a number of technical contributions.
First, we derive a new set of metrics that enable leveraging highly accurate matting datasets for quantifying the accuracy of boundary tracing in evaluating monocular depth maps. We analyze the effect of typical output resolutions and find that a high resolution is necessary but not sufficient to improve boundary accuracy.
%for improving boundary accuracy, it is necessary but not sufficient to predict depth maps at high resolution.
Second, we design an efficient multi-scale ViT-based architecture for capturing the global image context while also adhering to fine structures at high resolution.
Third, we devise a set of loss functions and a training curriculum that promote sharp depth estimates while training on real-world datasets that provide coarse and inaccurate supervision around boundaries, along with synthetic datasets that offer accurate pixelwise ground truth but limited realism.
Fourth, we contribute zero-shot focal length estimation from a single image that dramatically outperforms the prior state of the art.

\section{Related work}
\label{sec:related}
Early work on monocular depth estimation focused on training on individual datasets recorded with a single camera~\citep{Saxena2009:TPAMI, Eigen2014:NIPS, Eigen2015:ICCV}. Although this setup directly enabled metric depth predictions, it was limited to single datasets and narrow domains.

\mypara{Zero-shot depth estimation.}
MegaDepth~\citep{Li2018:CVPR} demonstrated that training on a diverse dataset allows generalizing monocular depth prediction beyond a specific domain. MiDaS~\citep{Ranftl2022:TPAMI} advanced this idea by training on a large mix of diverse datasets with a scale-and-shift-invariant loss. Follow-up works applied this recipe to transformer-based architectures~\citep{Ranftl2021:ICCV,Birkl2023:arxiv} and further expanded the set of feasible datasets through self-supervision~\citep{Spencer2023:ICCV,Yang2024:CVPR}.
A line of work uses self-supervision to learn from unlabeled image and video data~\citep{Petrovai2022:CVPR, Yang2024:CVPR}. A number of recent approaches~\citep{Ke2024:CVPR, Gui2024:arxiv} harness diffusion models to synthesize relative depth maps.
Although some of these methods demonstrated excellent generalization, their predictions are ambiguous in scale and shift, which precludes downstream applications that require accurate shapes, sizes, or distances.

\mypara{Zero-shot metric depth.}
A line of work sought to improve metric depth prediction through a global distribution of depth values~\citep{Fu2018:CVPR, Bhat2021:CVPR, Bhat2022:ECCV, Li2024:TIP} and further conditioning on scene type~\citep{Bhat2023:arxiv}.
A different approach directly takes into account camera intrinsics. Cam-Convs~\citep{Facil2019:CVPR} conditioned convolutions on the camera intrinsics. LeReS~\citep{Yin2021:CVPR} trains a separate network for undistorting point clouds to recover scale and shift, Metric3D~\citep{Yin2023:ICCV} scales images or depth maps to a canonical space and remaps estimated depth given the focal length, and ZeroDepth~\citep{Guizilini2023:ICCV} learns camera-specific embedddings in a variational framework. DMD~\citep{Saxena2023:arxiv} conditions a diffusion model on the field of view. Metric3D~v2~\citep{Hu2024:arxiv} leverages surface normals as an auxilliary output to improve metric depth. All of these methods require the camera intrinsics to be known and accurate. More recent works attempt to reason about unknown camera intrinsics either through a separate network~\citep{Spencer2024:arxiv} or by predicting a camera embedding for conditioning its depth predictions in a spherical space~\citep{Piccinelli2024:CVPR}.  
Akin to these recent approaches, our method does not require the focal length to be provided as input.
We propose to directly estimate the field of view from intermediate features of the depth prediction network, and show that this substantially outperforms the prior state of the art in the task of cross-domain focal length estimation.

\mypara{Sharp occluding contours.}
SharpNet~\citep{RamamonjisoaICCVW19} incorporates normals and occluding contour constraints, but requires additional contour and normal supervision during training.
BoostingDepth~\citep{Miangoleh2021:CVPR} obtains detailed predictions from a low-resolution network by applying it independently to image patches. Since the patches  lack global context, BoostingDepth fuses them through a sophisticated multi-step pipeline. 
PatchFusion~\citep{Li2024:CVPR} refines this concept through image-adaptive patch sampling and tailored modules that enable end-to-end training.
A recent line of work~\citep{Gui2024:arxiv,Ke2024:CVPR} leverages diffusion priors to enhance the sharpness of occlusion boundaries.
These approaches predominantly focus on predicting relative (rather than metric) depth.
We propose a simpler architecture without task-specific modules or diffusion priors and demonstrate that even sharper and more accurate results can be obtained while producing metric depth maps and reducing runtime by more than two orders of magnitude.

Guided depth super-resolution uses the input image to upsample low-resolution depth predictions~\citep{Metzger2023:CVPR, Zhong2023:ACM}. SMDNet~\citep{Tosi2021:CVPR} predicts bimodal mixture densities to sharpen occluding contours. And Ramamonjisoa~\etal~\citep{Ramamonjisoa2020:CVPR} introduce a module for learning to sharpen depth boundaries of a pretrained network.
These works are orthogonal to ours and could be applied to further upsample our high-resolution predictions.

To evaluate boundary tracing in predicted depth maps, \citet{Koch2018:ECCVW} introduce the iBims dataset with manual annotations of occluding contours and corresponding metrics.
The need for manual annotation and highly accurate depth ground truth constrain the benchmark to a small set of indoor scenes.
We contribute metrics based on segmentation and matting datasets that provide a complementary view by enabling evaluation on complex, dynamic environments or scenes with exceedingly fine detail for which ground-truth depth is impossible to obtain.

\mypara{Multi-scale vision transformers.}
Vision transformers (ViTs) have emerged as the dominant general-purpose architecture for perception tasks but operate at low resolution~\citep{Dosovitskiy2021:ICLR}. 
The computational complexity of their attention layers prohibits 
na\"{i}vely scaling them to higher resolutions and several works proposed alternatives~\citep{Zhu2021:ICLR,Liu2021:ICCV, Li2022:CVPR,Chu2021:Neurips,Liu2022:CVPR,Liu2023:CVPR,Cai2023:ICCV,Jaegle2022:ICLR}.
Another line of work modified the ViT architecture to produce a hierarchy of features~\citep{Fan2021:ICCV,Xie2021:Neurips,Yuan2021:Neurips,Ranftl2021:ICCV,Chen2021:ICCV,Lee2022:CVPR}.

Rather than modifying the ViT architecture, which requires computationally expensive retraining, we propose an architecture that applies a plain ViT backbone at multiple scales and fuses predictions into a single high-resolution output. This design benefits from ongoing improvements in ViT pretraining, as new variants can be easily swapped in~\citep{Oquab2024:TMLR,Peng2022:arxiv,Sun2023:arxiv}.

Pretrained vision transformers have been adapted for semantic segmenation and object detection. ViT-Adapter~\citep{Chen2023:ICLR} and ViT-CoMer~\citep{Xia2024:CVPR} supplement a pretrained ViT with a convolutional network for dense prediction, whereas 
ViT-Det~\citep{Li2022:ECCV} builds a feature pyramid on top of a pretrained ViT.
Distinct from these, we fuse features from the ViT applied at multiple scales to learn global context together with local detail.

\section{Method}
\label{sec:method}

\subsection{Network}
\label{sec:network}
\begin{figure}[t]
    \centering
    \includegraphics[trim=0.7cm 13.3cm 1.1cm 0cm, clip, width=1\linewidth]{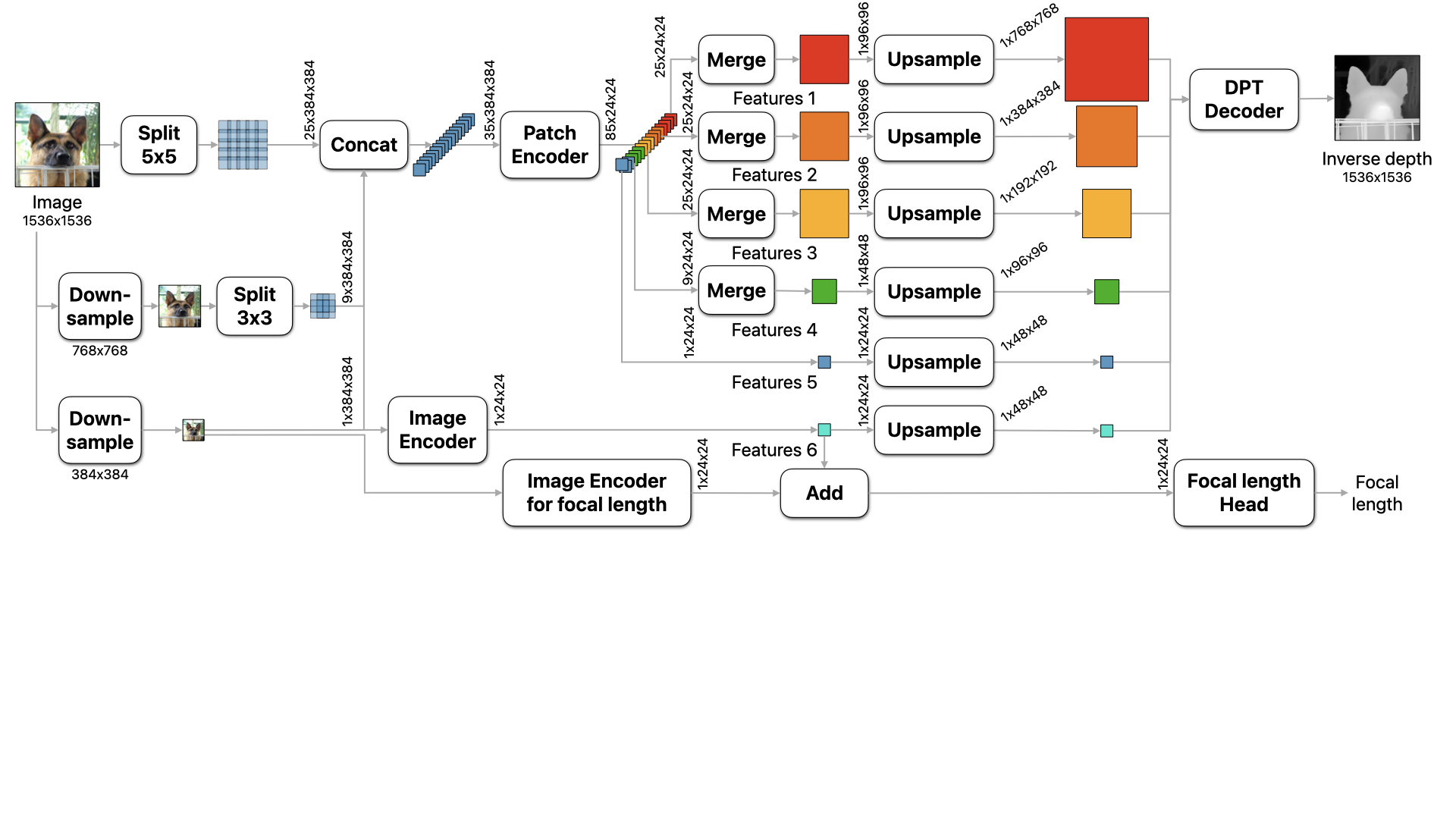}
    \vspace{-5mm}
    \caption{Overview of the network architecture. An image is downsampled at several scales. At each scale, it is split into patches, which are processed by a ViT-based patch encoder, with weights shared across scales. Patches are merged into feature maps, upsampled, and fused via a DPT decoder. Predictions are anchored by a separate image encoder that provides global context.}
    \label{fig:architecture}
    \vspace{-4mm}
\end{figure}

The key idea behind our network is to apply plain ViT encoders~\citep{Dosovitskiy2021:ICLR} on patches extracted at multiple scales and fuse their predictions into a single high-resolution dense depth prediction in an end-to-end trainable model (see \Fig~\ref{fig:architecture}).
As the patch encoder shares weights across all scales, it may intuitively learn a scale-invariant representation.
The image encoder anchors the patch predictions in a global context. It is applied to the whole input image, downsampled to the base input resolution of the chosen encoder backbone (in our case 384$\times$384).

The whole network operates at a fixed resolution of 1536$\times$1536, which was chosen as a multiple of the ViT's 384$\times$384. This guarantees a sufficiently large receptive field and constant runtimes for any image while preventing out-of-memory errors (which we repeatedly observed for variable-resolution approaches on large images). Confirming this design choice, our results in \Sec~\ref{sec:experiments} and \Tab~\ref{tab:runtime} demonstrate that Depth Pro is consistently orders of magnitude faster than variable-resolution approaches while being more accurate and producing sharper boundaries.
A key benefit of assembling our architecture from plain ViT encoders over custom encoders is the abundance of pretrained ViT-based backbones that can be harnessed~\citep{Oquab2024:TMLR,Peng2022:arxiv,Sun2023:arxiv}. 
We evaluate several pretrained backbones and compare our architecture to other high-resolution architectures in the appendices (\Tab~\ref{tab:backbone_alternatives} and \Sec~\ref{sec:high_resolution_alternatives}).

After initial downsampling to 1536$\times$1536, the input image is split into patches of $384\times384$ at each scale. For the two finest scales, we let patches overlap to avoid seams, which yields 25 and 9 patches, respectively. In total, we extract 35 patches, concatenate them along the batch dimension to allow efficient batch processing, and feed them to the patch encoder.
This yields a feature tensor at resolution $24\times24$ per input patch (Features 3 -- 6 in \Fig~\ref{fig:architecture}).
At the finest scale we further extract intermediate features (Features 1 \& 2 in \Fig~\ref{fig:architecture}) to capture finer-grained details, yielding additional $25 + 25 = 50$ feature patches.
We merge the feature patches into maps (detailed in Sec.~\ref{sec:merge}), which are fed into the decoder module, which resembles the DPT decoder~\citep{Ranftl2021:ICCV}.

In addition to sharing representations across scales, the patch-based application of the encoder network allows trivial parallelization as patches can be processed independently.
Another source of computational efficiency comes from the lower computational complexity of patch-based processing in comparison to scaling up the ViT to higher resolutions.
The reason is multi-head self-attention~\citep{Vaswani2017:NeurIPS}, whose computational complexity scales quadratically with the number of input pixels, and thus quartically in image dimension.

\subsection{Sharp monocular depth estimation}

\paragraph{Training objectives.}
For each input image $I$, our network $f$ predicts a canonical inverse depth image $C = f(I)$. To obtain a dense metric depth map $D_m$, we scale by the horizontal field of view, represented by the focal length $f_{\mathit{px}}$ and the width $w$~\citep{Yin2023:ICCV}: $D_m = \tfrac{f_{\mathit{px}}}{w C}.\label{eq:prediction}$

We train with several objectives, all based on canonical inverse depth, because this prioritizes areas close to the camera over farther areas or the whole scene, and thus supports visual quality in applications such as novel view synthesis~(see \Sec~\ref{sec:depth_representation}).
Let $\hat{C}$ be the ground-truth canonical inverse depth.
For all metric datasets we compute the mean absolute error~($\LMAE$, \Eq~\ref{eq:mae}) per pixel $i$, and discard pixels with an error in the top $20\%$ per image for real-world (as opposed to synthetic) datasets:
\begin{align}
\mathcal{L}_{\mathit{MAE}}(\hat{C},C) = \frac{1}{N}\sum^N_i|\hat{C}_i - C_i|.\label{eq:mae}
\end{align}
For all non-metric datasets (\ie those without reliable camera intrinsics or inconsistent scale), we normalize predictions and ground truth via the mean absolute deviation from the median~\citep{Ranftl2022:TPAMI} before applying a loss. We further compute errors on the first and second derivatives of (canoncial) inverse depth maps at multiple scales.
Let $\nabla_{*}$ indicate a spatial derivative operator~$*$, such as Scharr (S)~\citep{Scharr1997} or Laplace (L), and $p$ the error norm. We define the multi-scale derivative loss over $M$ scales as
\begin{align}
    \mathcal{L}_{*, p, M}(C, \hat{C}) = \frac{1}{M}\sum^M_j\frac{1}{N_j}\sum_i^{N_j} |\nabla_{*} C^j_i - \nabla_{*} \hat{C}^j_i|^p,
\end{align}
where the scales $j$ are computed by blurring and downsampling the inverse depth maps by a factor of 2 per scale.
As shorthands we define the Mean Absolute Gradient Error $\mathcal{L}_{\mathit{MAGE}} = \mathcal{L}_{S,1,6}$, the Mean Absolute Laplace Error $\mathcal{L}_{\mathit{MALE}} = \mathcal{L}_{L,1,6}$, and the Mean Squared Gradient Error $\mathcal{L}_{\mathit{MSGE}} = \mathcal{L}_{S,2,6}$.

\paragraph{Training curriculum.}
We propose a training curriculum motivated by the following observations.
First, training on a large mix of real-world and synthetic datasets improves generalization as measured by zero-shot accuracy~\citep{Ranftl2022:TPAMI,Ranftl2021:ICCV,Yang2024:CVPR,Hu2024:arxiv}.
Second, synthetic datasets provide pixel-accurate ground truth, whereas real-world datasets often contain missing areas, mismatched depth, or false measurements on object boundaries.
Third, predictions get sharper over the course of training.

Based on these observations, we design a two-stage training curriculum.
In the first stage, we aim to learn robust features that allow the network to generalize across domains.
To that end, we train on a mix of all labeled training sets.
Specifically, we minimize $\LMAE$ on metric datasets and its normalized version on non-metric datasets. $\LMAE$ is chosen for its robustness in handling potentially corrupted real-world ground truth. To steer the network towards sharp boundaries, we aim to also supervise on gradients of the predictions. Done na\"{i}vely, however, this can hinder optimization and slow down convergence. We found that a scale-and-shift-invariant loss on gradients, applied only to synthetic datasets, worked best. Controlled experiments are reported in \Sec~\ref{sec:supp_training_objectives} of the appendices.

The second stage of training is designed to sharpen boundaries and reveal fine details in the predicted depth maps. To minimize the effect of inaccurate ground truth, at this stage we only train on synthetic datasets that provide high-quality pixel-accurate ground truth. (Note that this inverts the common practice of first training on synthetic data and then fine-tuning on real data~\citep{Gaidon2016:CVPR,Gomez2023:arxiv,Sun2021:CVPR}.)
Specifically, we again minimize the $\LMAE$ and supplement it with a selection of losses on the first- and second-order derivatives: $\mathcal{L}_{\mathit{MAGE}}$, $\mathcal{L}_{\mathit{MALE}}$, and $\mathcal{L}_{\mathit{MSGE}}$.
We provide a detailed specification of the loss functions that are applied at each stage in the appendices.

\paragraph{Evaluation metrics for sharp boundaries.}
\label{sec:method_edge_metrics}
Applications such as novel view synthesis require depth maps to adhere to object boundaries. This is particularly challenging for thin structures. Misaligned or blurry boundaries can make objects appear distorted or split into parts. Common benchmarks for monocular depth prediction rarely take boundary sharpness into account. This may be attributed in part to the lack of diverse and realistic datasets with precise pixel-accurate ground-truth depth. To address this shortcoming, we propose a new set of metrics specifically for the evaluation of depth boundaries. Our key observation is that we can leverage existing high-quality annotations for matting, saliency, or segmentation as ground truth for depth boundaries. We treat annotations for these tasks as binary maps, which define a foreground/background relationship between an object and its environment. (This relationship may not hold in every case, especially for segmentation masks. However, we can easily discard such problematic cases through manual inspection. It is much easier to filter out a segmentation mask than to annotate it.) To ensure that the relationship holds, we only consider pixels around edges in the binary map.

We first define the metrics for depth maps and later derive the formulation for binary segmentation masks.
Motivated by the ranking loss~\citep{Chen2016:NIPS}, we use the pairwise depth ratio of neighboring pixels to define a foreground/background relationship.
Let $i,j$ be the locations of two neighboring pixels. We then define an occluding contour $c_d$ derived from a depth map $d$ as $c_d(i,j) = \left[ \tfrac{d(j)}{d(i)} > (1 + \tfrac{t}{100}) \right]$, where $[\cdot]$ is the Iverson bracket. Intuitively, this indicates the presence of an occluding contour between pixels $i$ and $j$ if their corresponding depth differs by more than $t\%$.
For all pairs of neighboring pixels, we can then compute the precision ($P$) and recall ($R$) as
\begin{equation}
    \text{P}(t) = \frac{\sum_{i,j\in N(i)} c_d(i,j) \wedge c_{\hat{d}}(i,j)}{ \sum_{i,j\in N(i)} c_{d}(i,j)} \text{ and }
    \text{R}(t) = \frac{\sum_{i,j\in N(i)} c_d(i,j) \wedge c_{\hat{d}}(i,j)}{ \sum_{i,j\in N(i)} c_{\hat{d}}(i,j)}.\label{eq:recall}
\end{equation}
Note that both $P$ and $R$ are scale-invariant. In our experiments, we report the F1 score.
To account for multiple relative depth ratios,
we further perform a weighted averaging of the F1 values with thresholds that range linearly from $t_{min}=5$ to $t_{max}=25$,
with stronger weights towards high threshold values.
Compared to other edge-based metrics (such as the edge accuracy and completion from iBims~\citep{Koch2018:ECCVW}), our metric does not require any manual edge annotation,
but simply pixelwise ground truth, which is easily obtained for synthetic datasets.

Similarly, we can also identify occluding contours from binary label maps that can be derived from real-world segmentation, saliency, and matting datasets.
Given a binary mask $b$ over the image, we define the presence of an occluding contour $c_b$ between pixels $i,j$ as $c_b(i,j) = b(i) \land \lnot b(j)$.
With this definition at hand, we compute the recall $\text{R}(t)$ by replacing the occluding contours from depth maps in \Eq~\ref{eq:recall} with those from binary maps. Since the binary maps commonly label whole objects, we cannot obtain ground-truth occluding contours that do not align with object silhouettes. Thus the boundary annotation is incomplete~-- some but not all occluding contours are identified by this procedure. Therefore we can only compute the recall but not the precision for binary maps.

To penalize blurry edges, we suppress non-maximum values of
$c_{\hat{d}}$ within the valid bounds of $c_{\hat{d}}(i,j)$ connected components. We report additional experiments and qualitative results in \Sec~\ref{sec:supp_additional_experiments}.

\subsection{Focal length estimation}
\label{sec:fov}
To handle images that may have inaccurate or missing \texttt{EXIF} metadata, we supplement our network with a focal length estimation head. A small convolutional head ingests frozen features from the depth estimation network and task-specific features from a separate ViT image encoder to predict the horizontal angular field-of-view.  We use $\mathcal{L}_2$ as the training loss.  We train the focal length head and the ViT encoder after the depth estimation training. Separating the focal length training has several benefits over joint training with the depth network. It avoids the necessity of balancing the depth and focal length training objectives. It also allows training the focal length head on a different set of datasets, excluding some narrow-domain single-camera datasets that are used in training the depth estimation network, and adding large-scale image datasets that provide focal length supervision but no depth supervision. Further details are provided in \Sec~\ref{sec:focal_head}.

\section{Experiments}
\label{sec:experiments}

This section summarizes the key results. Additional details and experiments are reported in the appendices, including details on datasets, hyperparameters, experimental protocols, and the comparison of runtimes, which is summarized in \Fig~{\ref{fig:am_vs_runtime}}. The appendices also report controlled experiments, including controlled studies on network architectures, training objectives, and training curricula.

Here we summarize a number of key comparisons of Depth Pro to state-of-the-art metric monocular depth estimation systems.
One challenge in conducting such a comparison is that many leading recent systems are trained on bespoke combinations of datasets. Some systems use proprietary datasets that are not publicly available, and some use datasets that are only available under restrictive licenses. Some recent systems also train on unlabeled datasets or incorporate pretrained models (e.g., diffusion models) that were trained on additional massive datasets.
This rules out the possibility of a comparison that controls for training data (e.g., only comparing to systems that use the same datasets we do). At this stage of this research area, the only feasible comparison to other leading cross-domain monocular depth estimation models is on a full system-to-system basis. Fully trained models (each trained on a large, partially overlapping and partially distinct collection of datasets) are compared to each other zero-shot on datasets that none of the compared systems trained on.

\mypara{Zero-shot metric depth.}
We evaluate our method's ability to predict zero-shot \emph{metric} depth and compare against the state of the art in \Tab~\ref{tab:sota_0shot_metric}.
% Baselines
Our baselines include Depth Anything~\citep{Yang2024:CVPR}, Metric3D~\citep{Yin2023:ICCV}, PatchFusion~\citep{Li2024:CVPR}, UniDepth~\citep{Piccinelli2024:CVPR}, ZeroDepth~\citep{Guizilini2023:ICCV} and ZoeDepth~\citep{Bhat2023:arxiv}. We also report results for the very recent Depth Anything v2~\citep{Yang2024:arxiv} and Metric3D v2~\citep{Hu2024:arxiv}.

% Metric
As an overall summary measure of metric depth accuracy, \Tab~\ref{tab:sota_0shot_metric} uses the $\delta_1$ metric~\citep{Ladicky2014:CVPR}, which is commonly used for this purpose~\citep{Yin2023:ICCV,Yang2024:CVPR,Piccinelli2024:CVPR}. It is defined as the percentage of inlier pixels, for which the predicted and ground-truth depths are within 25\% of each other. We picked this metric for its robustness, with the strictest threshold found in the literature ($25\%$).
\begin{table}[thbp]
    \newcolumntype{Y}{S[table-format=2.1,table-auto-round]}
    \centering
    \caption{\textbf{Zero-shot metric depth accuracy.}
    We report the $\delta_1$ score per dataset (higher is better) and aggregate performance across datasets via the average rank (lower is better).
    Methods in \textcolor{gray}{gray} are not strictly zero-shot. Results on additional metrics and datasets are presented in the appendices.
  }
  % \vspace{-1mm}
    \scriptsize
    \begin{tabularx}{\linewidth}{@{}X@{}Y*{5}{@{\hspace{4.2mm}}Y}@{\hspace{4mm}}|@{\hspace{4mm}}Y}
\toprule
Method
& {Booster} & {ETH3D} & {Middlebury} & {NuScenes} & {Sintel} & {Sun-RGBD} & {\textbf{Avg. Rank}$\downarrow$} \\
\midrule
\textcolor{Gray}{DepthAnything~\citep{Yang2024:CVPR}}
& \cellsecond{\grayed{52.3}} & \grayed{9.3}
& \grayed{39.3} & \grayed{35.4}
& \grayed{6.9} & \grayed{85.0} & \grayed{5.7} \\
\textcolor{Gray}{DepthAnything v2 ~\citep{Yang2024:arxiv}}
& \cellfirst{\grayed{59.5}} & \grayed{36.3}
& \grayed{37.2} & \grayed{17.7}
& \grayed{5.9} & \grayed{72.4} & \grayed{5.8} \\
\textcolor{Gray}{Metric3D ~\citep{Yin2023:ICCV}}
&  \grayed{4.7} &  \grayed{34.2}
& \grayed{13.6} & \cellthird{\grayed{64.4}}
& \cellthird{\grayed{17.3}} & \grayed{16.9} &  \grayed{5.8} \\
\textcolor{Gray}{Metric3D v2 ~\citep{Hu2024:arxiv}}
& \grayed{39.4} & \cellfirst{\grayed{87.7}}
& \grayed{29.9} & \cellsecond{\grayed{82.6}}
& \cellsecond{\grayed{38.3}} & \grayed{75.6} &  \grayed{\cellsecond{3.7}} \\
PatchFusion ~\citep{Li2024:CVPR}
& 22.6 & \cellsecond{51.8}
& \cellthird{49.9} & 20.4
& 14.0 & 53.6 & 5.2 \\
UniDepth ~\citep{Piccinelli2024:CVPR}
& 27.6 & 25.3
& 31.9 & \cellfirst{83.6}
& 16.5 & \cellfirst{95.8} & \cellthird{4.2} \\
ZeroDepth ~\citep{Guizilini2023:ICCV}
& OOM & OOM
& 46.5 & 64.3
& 12.9 & OOM & 4.6 \\
ZoeDepth ~\citep{Bhat2023:arxiv}
& 21.6 & 34.2
&  \cellsecond{53.8} & 28.1
&  7.8 & \cellthird{85.7} & 5.3  \\
\midrule
Depth Pro (Ours)
& \cellthird{46.6} & \cellthird{41.5}
& \cellfirst{60.5} & 49.1
& \cellfirst{40.0} & \cellsecond{89.0} & \cellfirst{2.5} \\
\bottomrule
\end{tabularx}
  \label{tab:sota_0shot_metric}
\end{table}

Corresponding tables for additional metrics can be found in \Sec~\ref{sec:supp_zero_shot} of the appendices, including $\mathit{AbsRel}$~\citep{Ladicky2014:CVPR}, $\mathit{Log}_{10}$, $\delta_2$ and $\delta_3$ scores, as well as point-cloud metrics~\citep{Spencer2022:TMLR}.
\Tab~\ref{tab:sota_0shot_metric} also reports the average rank of each method across datasets, a common way to summarize cross-dataset performance~\citep{Ranftl2022:TPAMI}.

% Datasets
We report results on Booster~\citep{Ramirez2024:PAMI}, Middlebury~\citep{Scharstein2014:GCPR}, Sun-RGBD~\citep{Song2015:CVPR}, ETH3D~\citep{Schps2017:CVPR}, nuScenes~\citep{Caesar2020:CVPR}, and Sintel~\citep{Butler2012:ECCV}, because, to our knowledge, they were never used in training any of the evaluated systems.
Despite our best efforts, we were not able to run ZeroDepth on Booster, Middlebury, or Sun-RGBD as it consistently ran out of memory due to the high image resolutions. More details on our evaluation setup can be found in \Sec~\ref{sec:details} of the appendix.

% Results interpretation
The results in \Tab~\ref{tab:sota_0shot_metric} confirm the findings of \citet{Piccinelli2024:CVPR}, who observed considerable domain bias in some of the leading metric depth estimation models.
Notably, Depth Anything v1~\&~v2 focus on \emph{relative} depth estimation; for metric depth, they provide different models for different domains, fine-tuned either for indoor or for outdoor scenes. Metric3D v1~\&~v2 provide domain-invariant models, but their performance depends strongly on careful selection of the crop size at test time, which is performed \emph{per domain} in their experiments and thus violates the zero-shot premise. We tried setting the crop size automatically based on the aspect ratio of the image, but this substantially degraded the performance of Metric3D; for this reason, we use the recommended non-zero-shot protocol, with the recommended per-domain crop sizes. Since domain-specific models and crop sizes violate the strict zero-shot premise we (and other baselines) operate under, we mark the Depth Anything and Metric3D results in gray in \Tab~\ref{tab:sota_0shot_metric}.

We find that Depth Pro demonstrates the strongest generalization by consistently scoring among the top approaches per dataset and obtaining the best average rank across all datasets.

\begin{table}[tb]
  \centering
  \caption{
  \textbf{Zero-shot boundary accuracy.}
  We report the F1 score for dataset with ground-truth depth, and boundary recall ($R$) for matting and segmentation datasets. Qualitative results are shown on a sample from the AM-2k dataset~\citep{Li2022:IJCV}. Higher is better for all metrics.
  }
  \scriptsize
  \begin{tabular}{@{}llcccccc}
    & Method & Sintel F1$\uparrow$ & Spring F1$\uparrow$ & iBims F1$\uparrow$ & AM R$\uparrow$ & P3M R$\uparrow$ & DIS R$\uparrow$ \\
    \midrule
    \multirow{6}{*}{\rotatebox[origin=r]{90}{Absolute}} & DPT~\citep{Ranftl2021:ICCV} & 0.181 & 0.029 & 0.113 & 0.055 & 0.075 & 0.018 \\
    & Metric3D~\citep{Yin2023:ICCV} & 0.037 & 0.000 & 0.055 & 0.003 & 0.003 & 0.001 \\
    & Metric3D v2~\citep{Hu2024:arxiv} & \cellsecond{0.321} & 0.024 & 0.096 & 0.024 & 0.013 & 0.006 \\
    & ZoeDepth~\citep{Bhat2023:arxiv} & 0.027 & 0.001 & 0.035 & 0.008 & 0.004 & 0.002 \\
    & PatchFusion~\citep{Li2024:CVPR} & 0.312 & \cellthird{0.032} & \cellthird{0.134} & 0.061 & \cellthird{0.109} & \cellsecond{0.068} \\
    & UniDepth ~\citep{Piccinelli2024:CVPR} & \cellthird{0.316} & 0.000 & 0.039 & 0.001 & 0.003 & 0.000 \\
    \midrule
    \multirow{2}{*}{\rotatebox[origin=r]{90}{Rel.}} & DepthAnything~\citep{Yang2024:CVPR} & 0.261 & \cellthird{0.045} & 0.127 & 0.058 & 0.094 & 0.023 \\
    & DepthAnything v2~\citep{Yang2024:arxiv} & 0.228 & \cellsecond{0.056} & 0.111 & \cellsecond{0.107} & \cellsecond{0.131} & \cellthird{0.056} \\
    & Marigold~\citep{Ke2024:CVPR} & 0.068 & 0.032 & \cellsecond{0.149} & \cellthird{0.064} & 0.101 & 0.049 \\
    \midrule
    & Depth Pro (Ours) & \cellfirst{0.409} & \cellfirst{0.079} & \cellfirst{0.176} & \cellfirst{0.173} & \cellfirst{0.168} & \cellfirst{0.077} \\
  \bottomrule
  \end{tabular}\\
  \vspace{3mm}
  \small
  \begin{tabular}{@{}c@{\hspace{1mm}}c@{\hspace{1mm}}c@{\hspace{1mm}}c@{\hspace{1mm}}c@{\hspace{1mm}}c@{}}
   \includegraphics[width=0.158\textwidth]{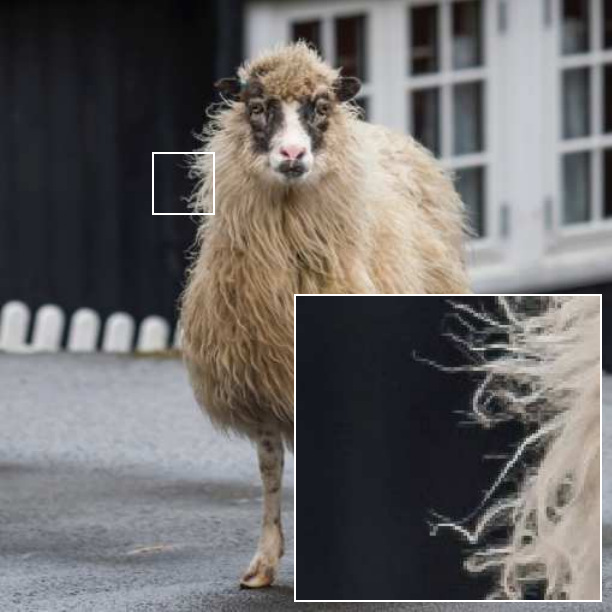} &
   \includegraphics[width=0.158\textwidth]{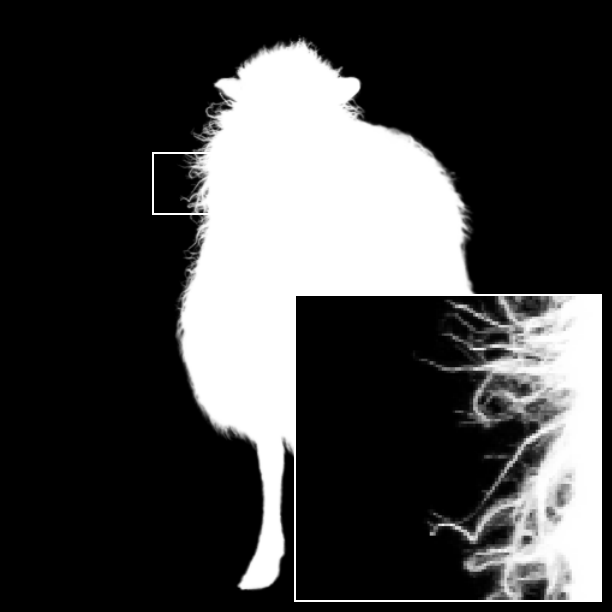} &
   \includegraphics[width=0.158\textwidth]{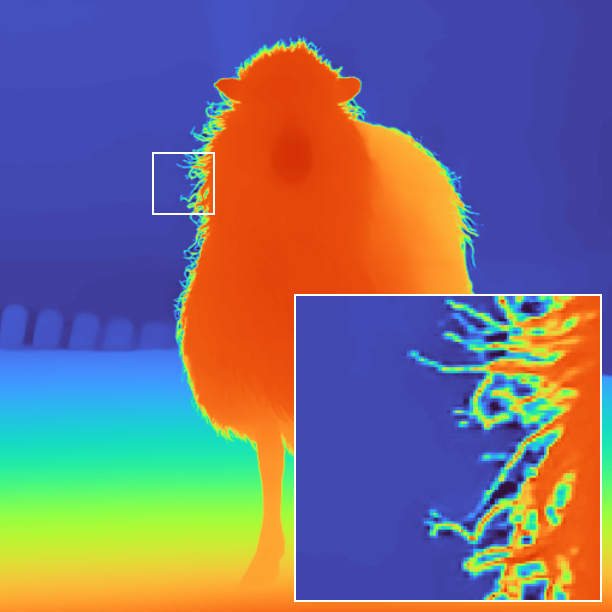} &
   \includegraphics[width=0.158\textwidth]{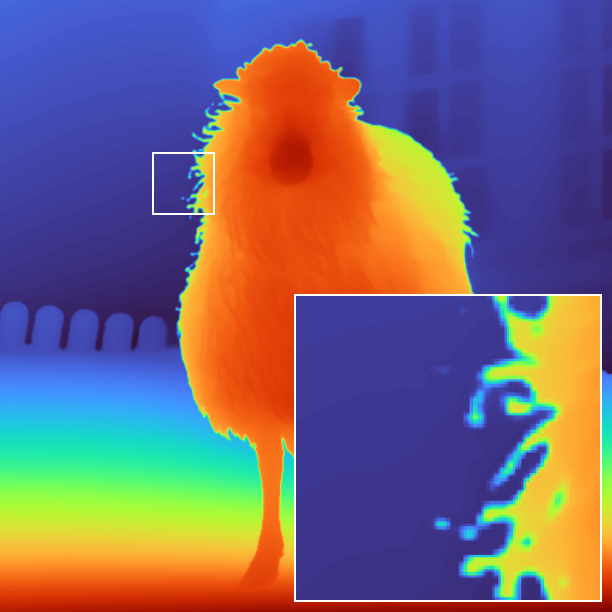} &
   \includegraphics[width=0.158\textwidth]{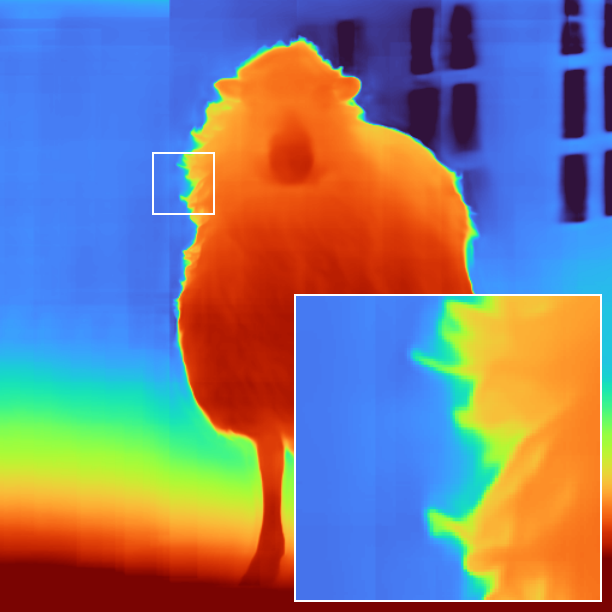} &
   \includegraphics[width=0.158\textwidth]{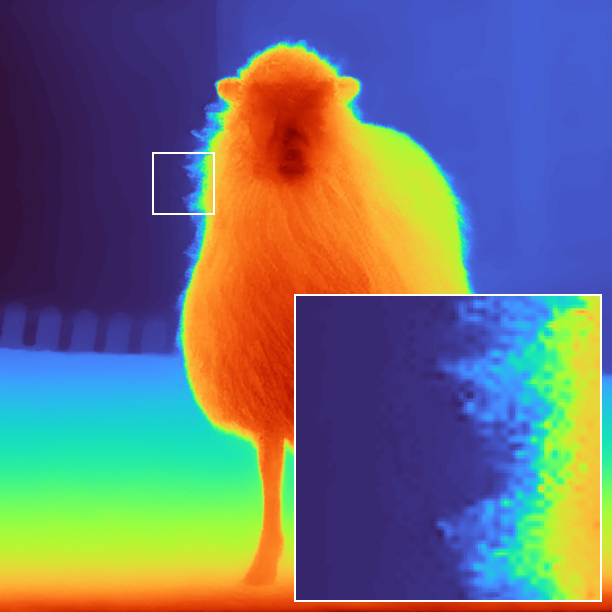}\\   {\scriptsize Image} & {\scriptsize Alpha Matte} & {\scriptsize Depth Pro (Ours)} & {\scriptsize DepthAnything v2} & {\scriptsize PatchFusion} & {\scriptsize Marigold}
 \end{tabular}
 \label{tab:sota_boundaries}
\end{table}

\mypara{Zero-shot boundaries.}
\Tab~\ref{tab:sota_boundaries} summarizes the evaluation of boundary accuracy for Depth Pro and several baselines. This evaluation is conducted in a zero-shot setting: models are only evaluated on datasets that were not seen during training.
% Baselines
Since our boundary metrics are scale-invariant, our baselines here also include methods that only predict relative (rather than absolute metric) depth.
Our absolute baselines include Metric3D~\citep{Yin2023:ICCV},
Metric3D v2 (`giant' model)~\citep{Hu2024:arxiv},
PatchFusion~\citep{Li2024:CVPR}, UniDepth~\citep{Piccinelli2024:CVPR}, and ZoeDepth~\citep{Bhat2023:arxiv}.
We also report results for the relative variants of Depth Anything v1~\&~v2~\citep{Yang2024:CVPR, Yang2024:arxiv} because they yield sharper boundaries than their metric counterparts.
Lastly, we include Marigold~\citep{Ke2024:CVPR}, a recent diffusion-based relative depth model that became popular due to its high-fidelity predictions.
% Metric
We use the boundary metrics introduced in \Sec~\ref{sec:method_edge_metrics}, and report the average boundary F1 score for datasets with ground-truth depth, and boundary recall ($R$) for datasets with matting or segmentation annotations. For image matting datasets, a pixel is marked as occluding when the value of the alpha matte is above $0.1$.

% Datasets
The datasets include Sintel~\citep{Butler2012:ECCV} and Spring~\citep{Mehl2023:CVPR}, which are synthetic. We also include the iBims dataset~\citep{Koch2018:ECCVW} which is often used specifically to evaluate depth boundaries, despite having low resolution. We refer to the appendices for a full slate of iBims-specific metrics.
To evaluate high-frequency structures encountered in natural images (such as hair or fur), we use AM-2k~\citep{Li2022:IJCV} and P3M-10k~\citep{Li2021:ACMMM}, which are high-resolution image matting datasets that were used to evaluate image matting models~\citep{LiJ2023:arxiv}.
Additionally, we further report results on the DIS-5k~\citep{Qin2022:ECCV} image segmentation dataset. This is an object segmentation dataset that provides highly accurate binary masks across diverse images. We manually remove samples in which the segmented object is occluded by foreground objects.
\Fig~{\ref{fig:am_vs_runtime}} visually summarizes the boundary recall metric on the AM-2k dataset, as a function of runtime.

\begin{figure}[t!]
    \centering
    \begin{tabular}{@{}c@{\hspace{0.3mm}}c*{2}{@{\hspace{0.2mm}}c}@{}}
        \raisebox{1.25cm}[0pt][0pt]{\rotatebox[origin=c]{90}{\footnotesize Input}} &
        \stackinset{r}{3pt}{t}{0.5pt}{\adjincludegraphics[height=1.7cm,trim={{.12\width} {.5\height} {.73\width} {.2\height}},clip,cfbox=White 0.25mm -0.25mm]{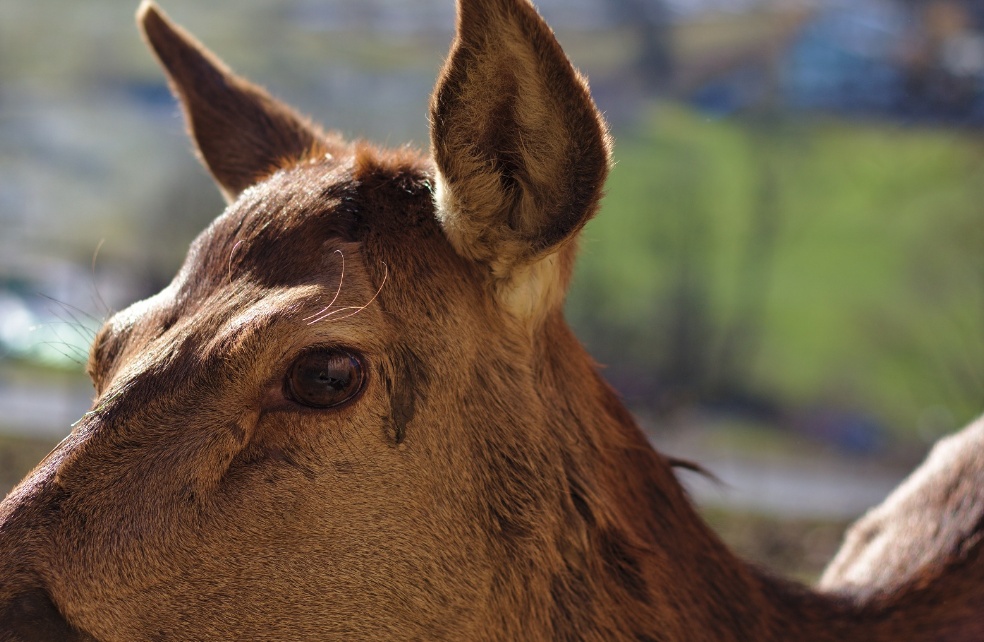}}{
            \begin{tikzpicture}
                \node[anchor=south west,inner sep=0] (image) at (0,0) {\begin{overpic}[height=0.19\textwidth]
                {fig/tmpi/m_16b22afc/m_16b22afc.jpg}
                %\put (2,2) {\textcolor{white}{Input}}
                \end{overpic}
                };
                \begin{scope}[x={(image.south east)},y={(image.north west)}]
                \draw[white,thick] (0.12,0.5) rectangle (0.27,0.8);
                \end{scope}
            \end{tikzpicture}
        }
        &
        % %
        \stackinset{r}{3pt}{b}{0.5pt}{\adjincludegraphics[height=1.1cm,trim={{.62\width} {.65\height} {.23\width} {.2\height}},clip,cfbox=White 0.25mm -0.25mm]{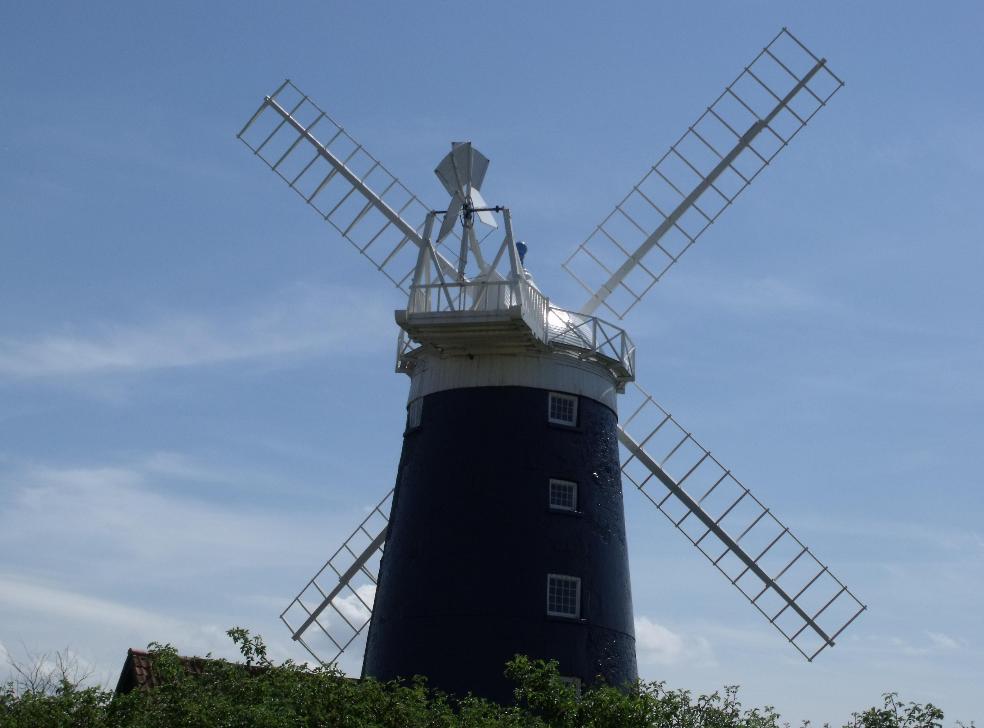}}{
            \begin{tikzpicture}
                \node[anchor=south west,inner sep=0] (image) at (0,0) {\includegraphics[height=0.19\textwidth]{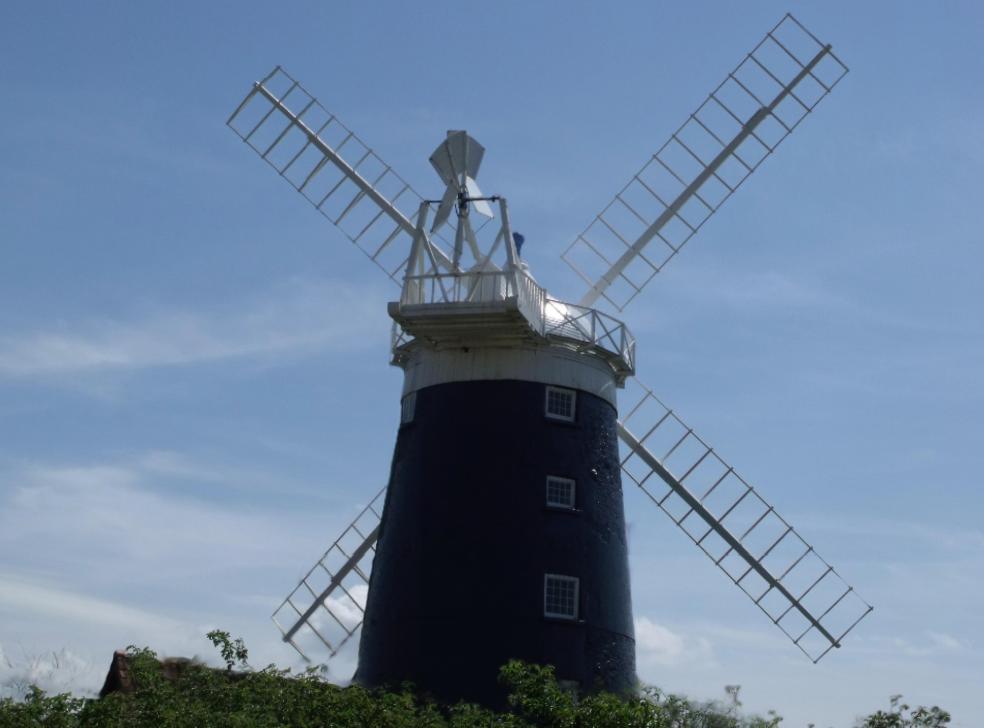}};
                \begin{scope}[x={(image.south east)},y={(image.north west)}]
                \draw[white,thick] (0.62, 0.65) rectangle (0.77, 0.8);
                \end{scope}
            \end{tikzpicture}
        }&
        \stackinset{r}{3pt}{t}{0.5pt}{\adjincludegraphics[height=1cm,trim={{.4\width} {.47\height} {.5\width} {.43\height}},clip,cfbox=White 0.25mm -0.25mm]{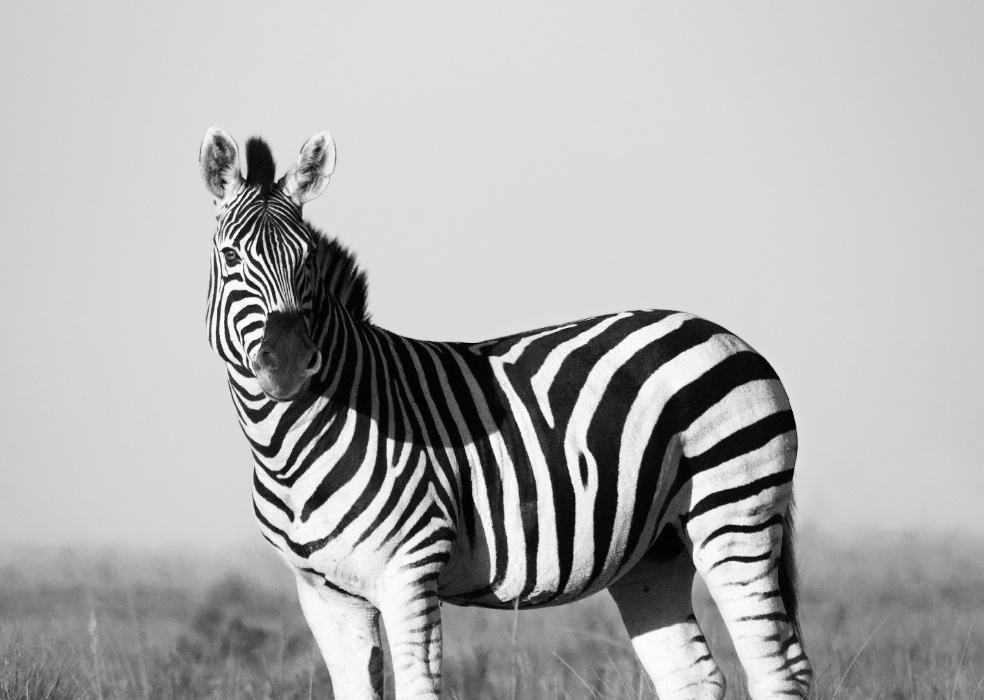}}{
            \begin{tikzpicture}
                \node[anchor=south west,inner sep=0] (image) at (0,0) {\includegraphics[height=0.19\textwidth]{fig/tmpi/m_0c7da85d/m_0c7da85d.jpg}};
                \begin{scope}[x={(image.south east)},y={(image.north west)}]
                \draw[white,thick] (0.4, 0.47) rectangle (0.5, 0.57);
                \end{scope}
            \end{tikzpicture}
        }\\
        \raisebox{1.25cm}[0pt][0pt]{\rotatebox[origin=c]{90}{\footnotesize Depth Pro}} &
        \stackinset{r}{3pt}{t}{0.5pt}{\adjincludegraphics[height=1.7cm,trim={{.13\width} {.49\height} {.72\width} {.21\height}},clip,cfbox=White 0.25mm -0.25mm]{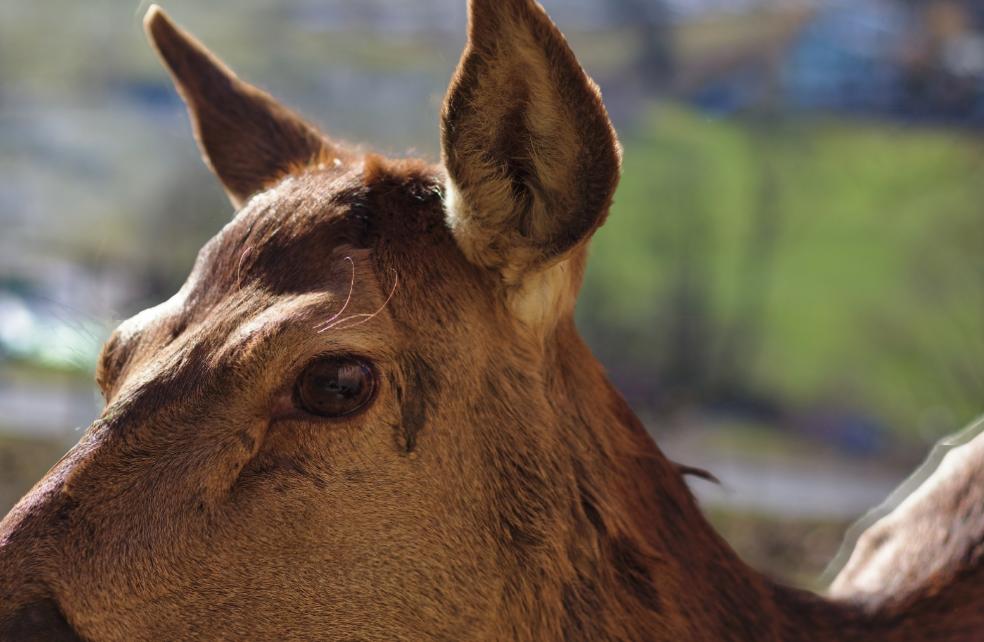}}{
            \begin{tikzpicture}
                \node[anchor=south west,inner sep=0] (image) at (0,0) {\begin{overpic}[height=0.19\textwidth]{fig/tmpi/m_16b22afc/m_16b22afc_depthpro_frame0078.jpg}
                % \put (2,2) {\textcolor{white}{Depth Pro}}
                \end{overpic}
                };
                \begin{scope}[x={(image.south east)},y={(image.north west)}]
                \draw[white,thick] (0.13,0.49) rectangle (0.28,0.79);
                \end{scope}
            \end{tikzpicture}
        }&
        \stackinset{r}{3pt}{b}{0.5pt}{\adjincludegraphics[height=1.1cm,trim={{.62\width} {.66\height} {.23\width} {.19\height}},clip,cfbox=White 0.25mm -0.25mm]{fig/tmpi/024/024_depthpro_frame0038.jpg}}{
            \begin{tikzpicture}
                \node[anchor=south west,inner sep=0] (image) at (0,0) {\includegraphics[height=0.19\textwidth]{fig/tmpi/024/024_depthpro_frame0038.jpg}};
                \begin{scope}[x={(image.south east)},y={(image.north west)}]
                \draw[white,thick] (0.62, 0.66) rectangle (0.77, 0.81);
                \end{scope}
            \end{tikzpicture}
        }&
        \stackinset{r}{3pt}{t}{0.5pt}{\adjincludegraphics[height=1cm,trim={{.4\width} {.45\height} {.5\width} {.45\height}},clip,cfbox=White 0.25mm -0.25mm]{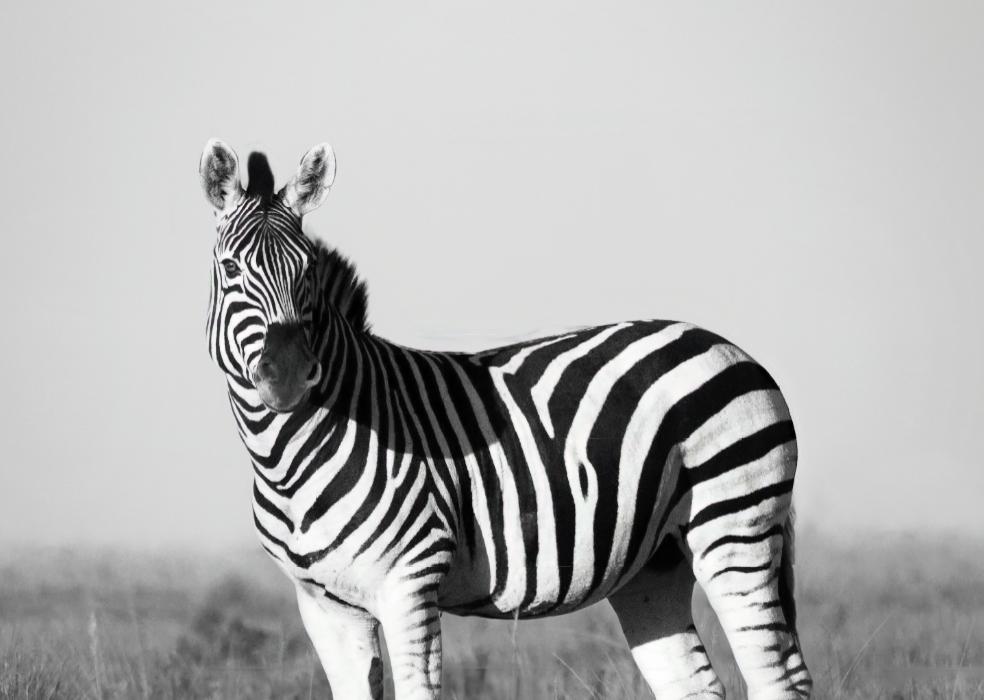}}{
            \begin{tikzpicture}
                \node[anchor=south west,inner sep=0] (image) at (0,0) {\includegraphics[height=0.19\textwidth]{fig/tmpi/m_0c7da85d/m_0c7da85d_depthpro_frame0000.jpg}};
                \begin{scope}[x={(image.south east)},y={(image.north west)}]
                \draw[white,thick] (0.4, 0.45) rectangle (0.5, 0.55);
                \end{scope}
            \end{tikzpicture}
        }\\
        %%%%
        \raisebox{1.25cm}[0pt][0pt]{\rotatebox[origin=c]{90}{\footnotesize Depth Anything v2}} &
        \stackinset{r}{3pt}{t}{0.5pt}{\adjincludegraphics[height=1.7cm,trim={{.13\width} {.49\height} {.72\width} {.21\height}},clip,cfbox=White 0.25mm -0.25mm]{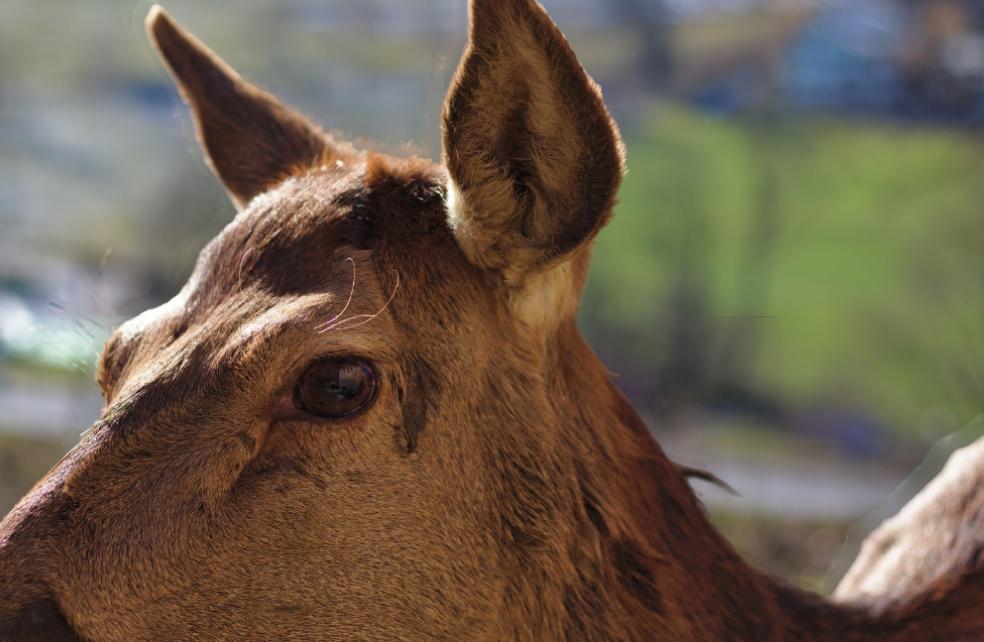}}{
            \begin{tikzpicture}
                \node[anchor=south west,inner sep=0] (image) at (0,0) {\begin{overpic}[height=0.19\textwidth]{fig/tmpi/m_16b22afc/m_16b22afc_depth_anything_v2_relative_frame0078.jpg}
                % \put (2,2) {\textcolor{white}{Depth Anything v2}}
                \end{overpic}};
                \begin{scope}[x={(image.south east)},y={(image.north west)}]
                \draw[white,thick] (0.13,0.49) rectangle (0.28,0.79);
                \end{scope}
            \end{tikzpicture}
        }&
        \stackinset{r}{3pt}{b}{0.5pt}{\adjincludegraphics[height=1.1cm,trim={{.62\width} {.66\height} {.23\width} {.19\height}},clip,cfbox=White 0.25mm -0.25mm]{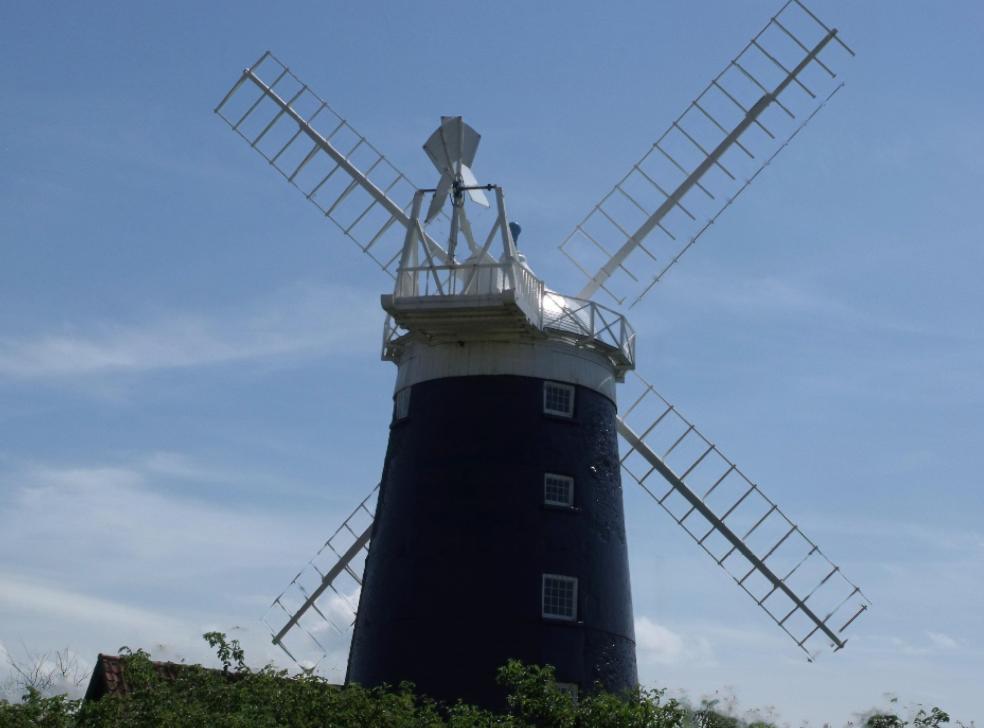}}{
            \begin{tikzpicture}
                \node[anchor=south west,inner sep=0] (image) at (0,0) {\includegraphics[height=0.19\textwidth]{fig/tmpi/024/024_depth_anything_v2_relative_frame0038.jpg}};
                \begin{scope}[x={(image.south east)},y={(image.north west)}]
                \draw[white,thick] (0.62, 0.66) rectangle (0.77, 0.81);
                \end{scope}
            \end{tikzpicture}
        }&
        \stackinset{r}{3pt}{t}{0.5pt}{\adjincludegraphics[height=1cm,trim={{.4\width} {.45\height} {.5\width} {.45\height}},clip,cfbox=White 0.25mm -0.25mm]{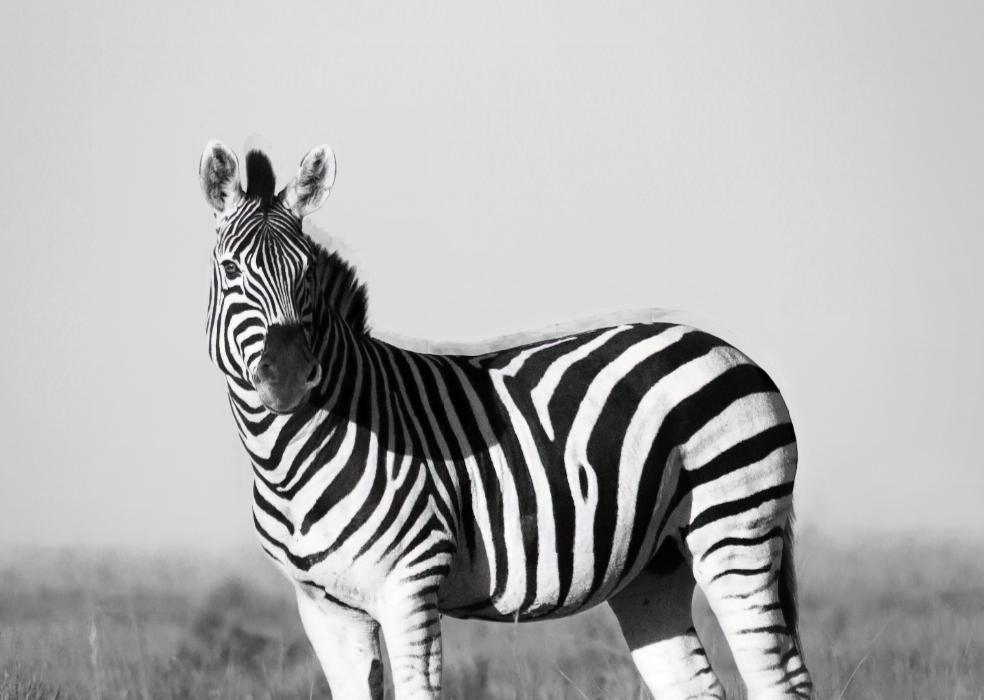}}{
            \begin{tikzpicture}
                \node[anchor=south west,inner sep=0] (image) at (0,0) {\includegraphics[height=0.19\textwidth]{fig/tmpi/m_0c7da85d/m_0c7da85d_depth_anything_v2_relative_frame0000.jpg}};
                \begin{scope}[x={(image.south east)},y={(image.north west)}]
                \draw[white,thick] (0.4, 0.45) rectangle (0.5, 0.55);
                \end{scope}
            \end{tikzpicture}
        }\\
        %%%%
        \raisebox{1.25cm}[0pt][0pt]{\rotatebox[origin=c]{90}{\footnotesize Marigold}} &
        \stackinset{r}{3pt}{t}{0.5pt}{\adjincludegraphics[height=1.7cm,trim={{.13\width} {.49\height} {.72\width} {.21\height}},clip,cfbox=White 0.25mm -0.25mm]{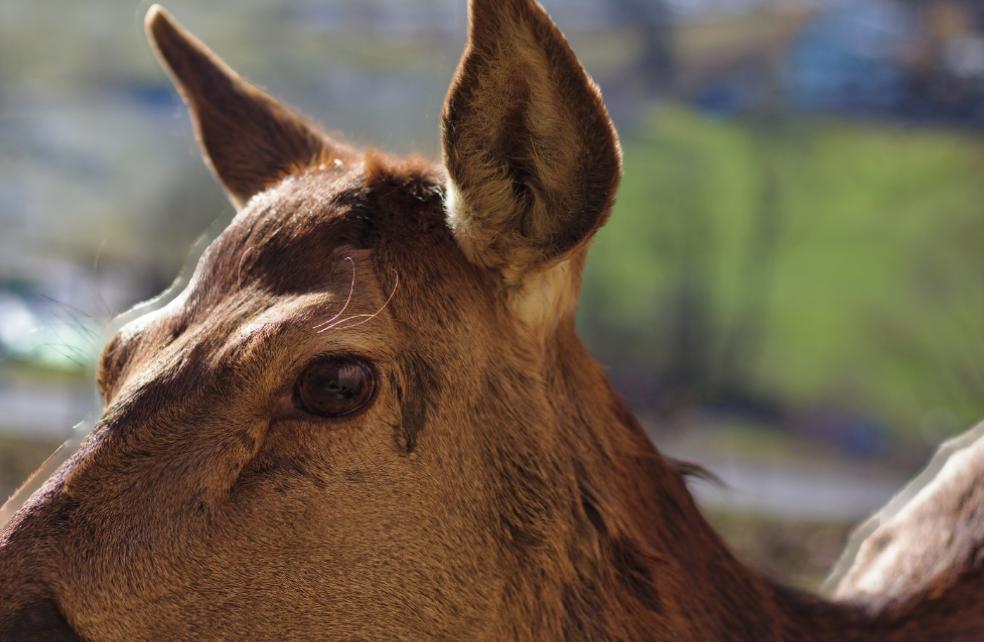}}{
            \begin{tikzpicture}
                \node[anchor=south west,inner sep=0] (image) at (0,0) {\begin{overpic}[height=0.19\textwidth]{fig/tmpi/m_16b22afc/m_16b22afc_marigold_frame0078.jpg}
                % \put (2,2) {\textcolor{white}{Marigold}}
                \end{overpic}};
                \begin{scope}[x={(image.south east)},y={(image.north west)}]
                \draw[white,thick] (0.13,0.49) rectangle (0.28,0.79);
                \end{scope}
            \end{tikzpicture}
        }&
        \stackinset{r}{3pt}{b}{0.5pt}{\adjincludegraphics[height=1.1cm,trim={{.62\width} {.66\height} {.23\width} {.19\height}},clip,cfbox=White 0.25mm -0.25mm]{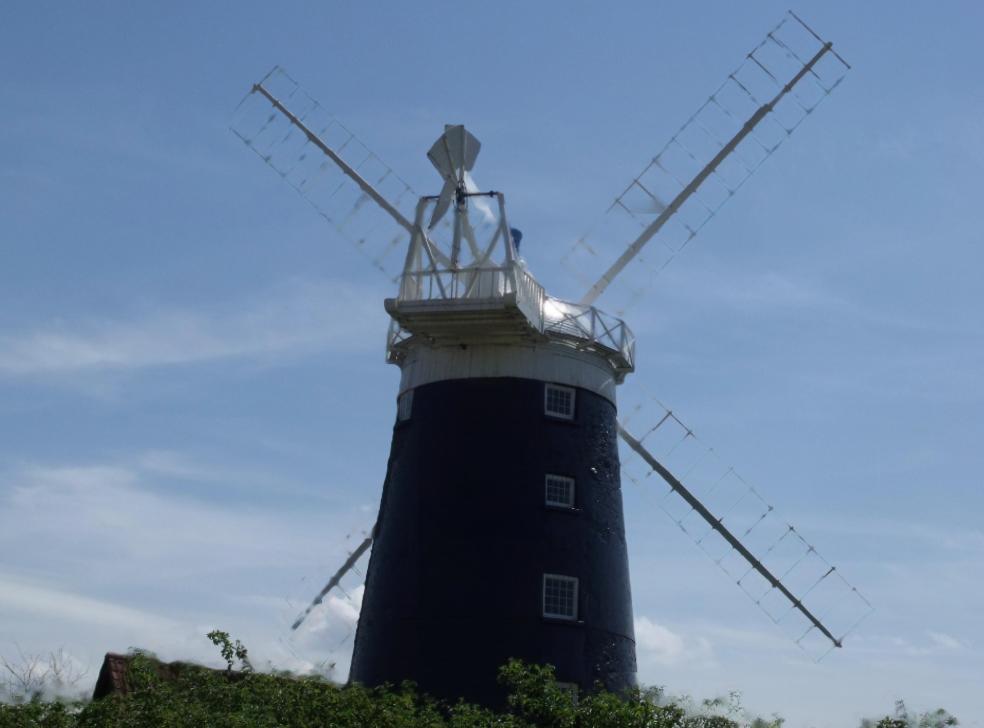}}{
            \begin{tikzpicture}
                \node[anchor=south west,inner sep=0] (image) at (0,0) {\includegraphics[height=0.19\textwidth]{fig/tmpi/024/024_marigold_frame0038.jpg}};
                \begin{scope}[x={(image.south east)},y={(image.north west)}]
                \draw[white,thick] (0.62, 0.66) rectangle (0.77, 0.81);
                \end{scope}
            \end{tikzpicture}
        }&
        \stackinset{r}{3pt}{t}{0.5pt}{\adjincludegraphics[height=1cm,trim={{.4\width} {.45\height} {.5\width} {.45\height}},clip,cfbox=White 0.25mm -0.25mm]{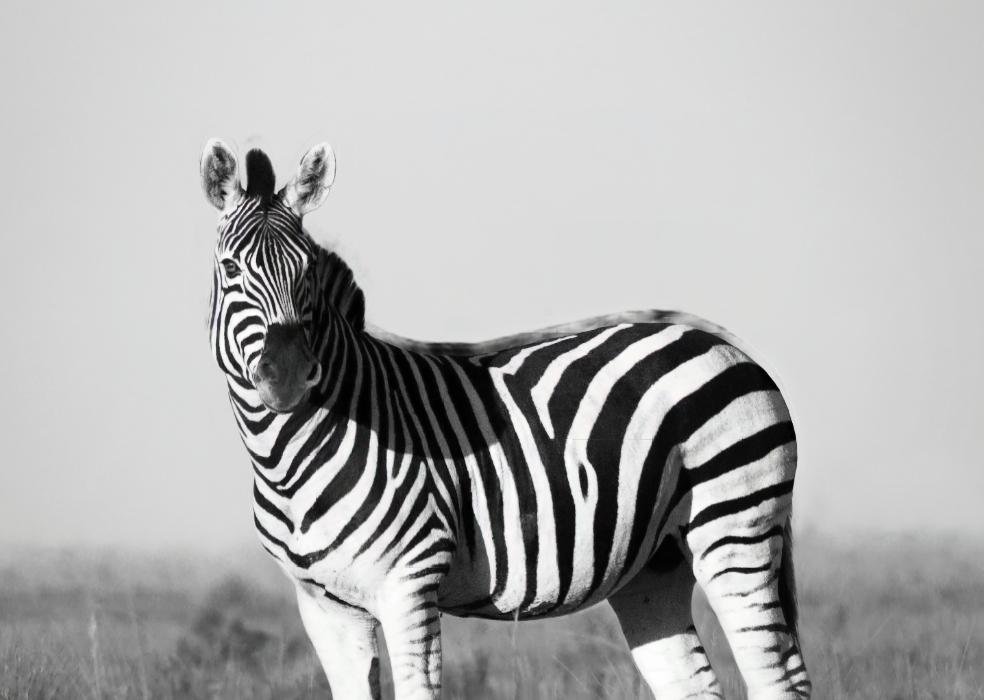}}{
            \begin{tikzpicture}
                \node[anchor=south west,inner sep=0] (image) at (0,0) {\includegraphics[height=0.19\textwidth]{fig/tmpi/m_0c7da85d/m_0c7da85d_marigold_frame0000.jpg}};
                \begin{scope}[x={(image.south east)},y={(image.north west)}]
                \draw[white,thick] (0.4, 0.45) rectangle (0.5, 0.55);
                \end{scope}
            \end{tikzpicture}
        }\\
        %%%%
        \raisebox{1.25cm}[0pt][0pt]{\rotatebox[origin=c]{90}{\footnotesize Metric3D v2}} &
        \stackinset{r}{3pt}{t}{0.5pt}{\adjincludegraphics[height=1.7cm,trim={{.13\width} {.49\height} {.72\width} {.21\height}},clip,cfbox=White 0.25mm -0.25mm]{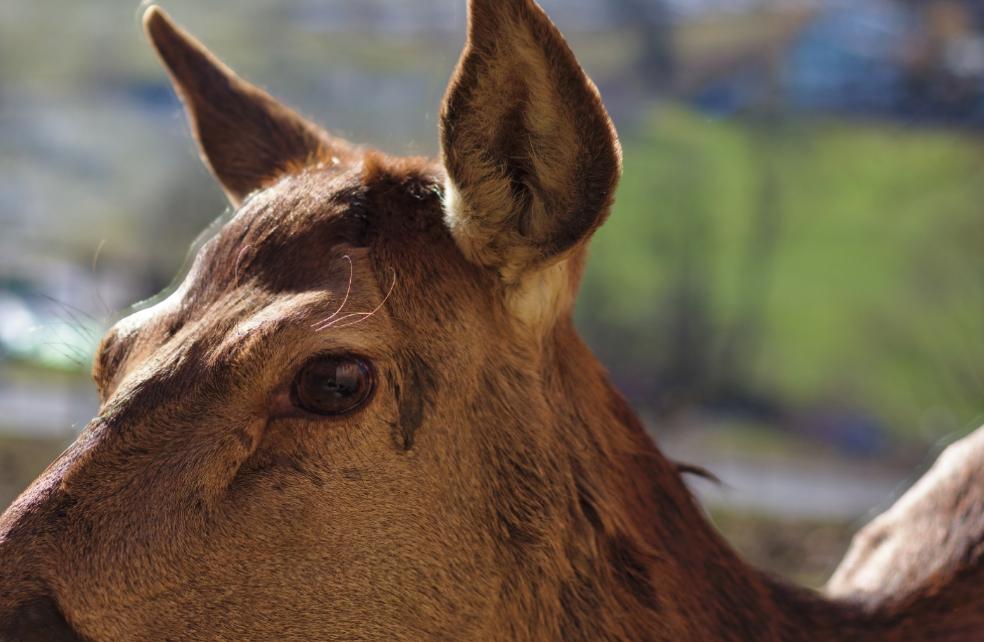}}{
            \begin{tikzpicture}
                \node[anchor=south west,inner sep=0] (image) at (0,0) {\begin{overpic}[height=0.19\textwidth]{fig/tmpi/m_16b22afc/m_16b22afc_metric3d_frame0078.jpg}
                % \put (2,2) {\textcolor{white}{Metric3D v2}}
                \end{overpic}
                };
                \begin{scope}[x={(image.south east)},y={(image.north west)}]
                 \draw[white,thick] (0.13,0.49) rectangle (0.28,0.79);
                \end{scope}
            \end{tikzpicture}
        }&
        \stackinset{r}{3pt}{b}{0.5pt}{\adjincludegraphics[height=1.1cm,trim={{.62\width} {.66\height} {.23\width} {.19\height}},clip,cfbox=White 0.25mm -0.25mm]{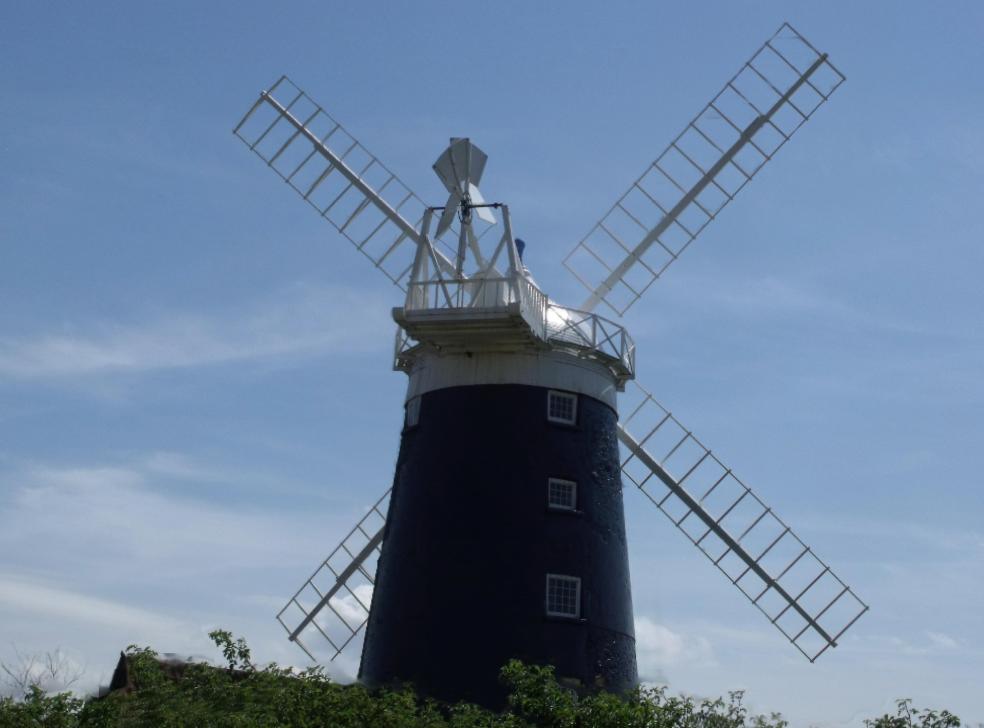}}{
            \begin{tikzpicture}
                \node[anchor=south west,inner sep=0] (image) at (0,0) {\includegraphics[height=0.19\textwidth]{fig/tmpi/024/024_metric3d_frame0038.jpg}};
                \begin{scope}[x={(image.south east)},y={(image.north west)}]
                \draw[white,thick] (0.62, 0.66) rectangle (0.77, 0.81);
                \end{scope}
            \end{tikzpicture}
        }&
        \stackinset{r}{3pt}{t}{0.5pt}{\adjincludegraphics[height=1cm,trim={{.4\width} {.45\height} {.5\width} {.45\height}},clip,cfbox=White 0.25mm -0.25mm]{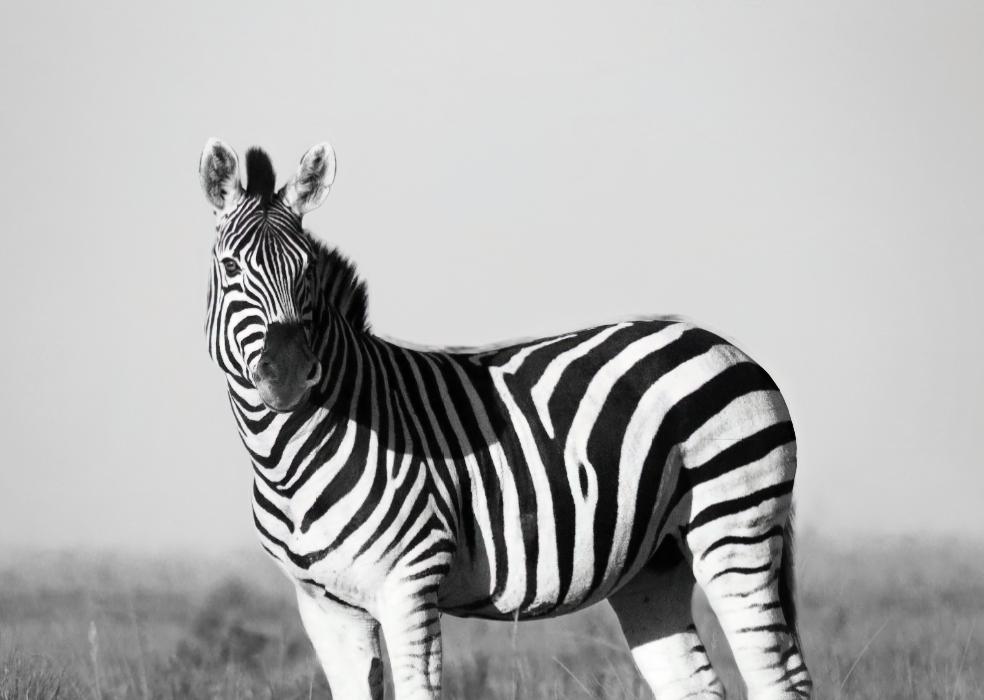}}{
            \begin{tikzpicture}
                \node[anchor=south west,inner sep=0] (image) at (0,0) {\includegraphics[height=0.19\textwidth]{fig/tmpi/m_0c7da85d/m_0c7da85d_metric3d_frame0000.jpg}};
                \begin{scope}[x={(image.south east)},y={(image.north west)}]
                \draw[white,thick] (0.4, 0.45) rectangle (0.5, 0.55);
                \end{scope}
            \end{tikzpicture}
        }\\
    \end{tabular}
    % \vspace{-1mm}
   \caption{\textbf{Impact on novel view synthesis.} We plug depth maps produced by Depth Pro, Marigold~\citep{Ke2024:CVPR}, Depth Anything v2~\citep{Yang2024:arxiv}, and Metric3D v2~\citep{Hu2024:arxiv} into a recent publicly available novel view synthesis system~\citep{Khan2023:ICCV}. We demonstrate results on images from AM-2k~\citep{Li2022:IJCV} (1st \& 3rd column) and DIS-5k~\citep{Qin2022:ECCV} (2nd column). The insets highlight typical artifacts from inaccurate boundaries. All methods except Depth Pro add ghosting edges to the horse (1st column) and zebra (3rd column). Marigold and Depth Anything v2 miss thin structures of the windmill (2nd column). Depth Pro produces sharper and more accurate depth maps, yielding cleaner synthesized views.}
  \label{fig:applications_nvs}
  % \vspace{-3mm}
\end{figure}

% Results interpretation
We find that Depth Pro produces more accurate boundaries than all baselines on all datasets, by a significant margin. As can be observed in \Fig~\ref{fig:teaser}, in the images in \Tab~\ref{tab:sota_boundaries}, and the additional results in \Sec~\ref{sec:supp_additional_results}, the competitive metric accuracy of Metric3D v2 and Depth Anything v2 does not imply sharp boundaries. Depth Pro has a consistently higher recall for thin structures like hair and fur and yields sharper boundaries. This is also true in comparison to the diffusion-based Marigold, %which leverages a prior trained on billions of real-word images, 
as well as PatchFusion, which operates at variable resolution. Note that Depth Pro is orders of magnitude faster than Marigold and PatchFusion (see \Fig~\ref{fig:am_vs_runtime} \& \Tab~\ref{tab:runtime}).
\Fig~\ref{fig:applications_nvs} demonstrates the benefits of sharp boundary prediction for novel view synthesis from a single image.

\mypara{Focal length estimation.}
Prior work~\citep{Piccinelli2024:CVPR,Kocabas2021:ICCV,Baradad2020:CVPR} does not provide comprehensive systematic evaluations of focal length estimators on in-the-wild images. To address this, we curated a \emph{zero-shot} test dataset. To this end, we selected diverse datasets with intact EXIF data, enabling reliable assessment of focal length estimation accuracy. FiveK~\citep{Bychkovsky2011:CVPR}, DDDP~\citep{Abuolaim2020:ECCV}, and RAISE~\citep{Dang2015:MMSys} contribute professional-grade photographs taken with SLR cameras. SPAQ~\citep{Fang2020:CVPR} provides casual photographs from mobile phones. PPR10K~\citep{jie2021:cvpr} provides high-quality portrait images. Finally, ZOOM~\citep{Zhang2019:CVPR} includes sets of scenes captured at various optical zoom levels.
\begin{table}[thb]
\caption{\textbf{Comparison on focal length estimation.}  We report $\delta_{25\%}$ and $\delta_{50\%}$ for each dataset, i.e., the percentage of images with relative error (focal length in mm) less than 25\% and 50\%, respectively.}
\label{tab:fov}
% \vspace{-1mm}
\scriptsize
\begin{tabularx}{\linewidth}{@{}X*{6}{@{\hspace{3mm}}c@{\hspace{2mm}}c}}
\toprule
    & \multicolumn{2}{c}{DDDP}
    & \multicolumn{2}{c}{FiveK}
    & \multicolumn{2}{c}{PPR10K}
    & \multicolumn{2}{c}{RAISE}
    & \multicolumn{2}{c}{SPAQ}
    & \multicolumn{2}{c}{ZOOM} \\
    & $\delta_{25\%}$ & $\delta_{50\%}$
    & $\delta_{25\%}$ & $\delta_{50\%}$
    & $\delta_{25\%}$ & $\delta_{50\%}$
    & $\delta_{25\%}$ & $\delta_{50\%}$
    & $\delta_{25\%}$ & $\delta_{50\%}$
    & $\delta_{25\%}$ & $\delta_{50\%}$ \\
    \midrule
    UniDepth~\citep{Piccinelli2024:CVPR}
    & 6.8  & \cellthird{40.3}
    & 24.8 & 56.2
    & 13.8 & 44.2
    & 35.4 & 74.8
    & \cellthird{44.2} & \cellthird{77.4}
    & 20.4 & \cellsecond{45.4} \\
    SPEC~\citep{Kocabas2021:ICCV}
    & \cellsecond{14.6} & \cellsecond{46.3}
    & \cellsecond{30.2} & \cellthird{56.6}
    & \cellsecond{34.6} & \cellsecond{67.0}
    & \cellthird{49.2} & \cellsecond{78.6}
    & \cellsecond{50.0} & \cellsecond{82.2}
    & \cellsecond{23.2} & \cellthird{43.6} \\
    im2pcl~\citep{Baradad2020:CVPR}
    & \cellthird{7.3} & 29.6
    & \cellthird{28.0} & \cellsecond{60.0}
    & \cellthird{24.2} & \cellthird{61.4}
    & \cellsecond{51.8} & \cellthird{75.2}
    & 26.6 & 55.0
    & \cellthird{22.4} & 42.8 \\
\midrule
    Depth Pro (Ours) & \cellfirst{66.9} & \cellfirst{85.8} & \cellfirst{74.2} & \cellfirst{92.4} & \cellfirst{64.6} & \cellfirst{88.8} & \cellfirst{84.2} & \cellfirst{96.4} & \cellfirst{68.4} & \cellfirst{85.2} & \cellfirst{69.8} & \cellfirst{91.6}\\
\bottomrule
\end{tabularx}
\end{table}

\Tab~\ref{tab:fov} compares Depth Pro against state-of-the-art focal length estimators and shows the percentage of images with relative estimation error under 25\% and 50\%, respectively. Depth Pro is the most accurate across all datasets. For example, on PPR10K, a dataset of human portraits, our method leads with 64.6\% of the images having a focal length error below 25\%, while the second-best method, SPEC, only achieves 34.6\% on this metric. We attribute this superior performance to our network design and training protocol, which decouple training of the focal length estimator from the depth network, enabling us to use different training sets for these two tasks. Further controlled experiments are reported in \Sec~\ref{sec:supp_focal_length} of the appendices.

\section{Conclusion \& limitations}
\label{sec:limitations}
Depth Pro produces high-resolution metric depth maps with high-frequency detail at sub-second runtimes.
Our model achieves state-of-the-art zero-shot metric depth estimation accuracy without requiring metadata such as camera intrinsics, and traces out occlusion boundaries in unprecedented detail, facilitating applications such as novel view synthesis from single images `in the wild'. While Depth Pro outperforms prior work along multiple dimensions, it is not without limitations. For example, the model is limited in dealing with translucent surfaces and volumetric scattering, where the definition of pixelwise depth is ill-posed and ambiguous.

\bibliography{paper}
\bibliographystyle{iclr2025_conference}

\appendix
\section*{Supplemental material}

In Section~\ref{sec:supp_additional_results}, we provide additional results and experiments. \Sec~\ref{sec:supp_qualitative} presents further qualitative comparisons to baselines, \Sec{\ref{sec:supp_zero_shot}} presents a more detailed zero-shot evaluation, \Sec{\ref{sec:supp_runtime}} lists runtimes for all evaluated methods, and \Sec{\ref{sec:supp_additional_experiments}} presents additional experiments on boundary accuracy.
Section~\ref{sec:controlled_experiments} showcases a selection of controlled experiments on Depth Pro that helped guide architectural choices (\Sec{\ref{sec:supp_network_architecture}}, \Sec{\ref{sec:high_resolution_alternatives}}, and \Sec{\ref{sec:supp_focal_length}}), training objective design (\Sec{\ref{sec:supp_training_objectives}}), and curriculum training (\Sec{\ref{sec:supp_full_curricula}}).
In Section~\ref{sec:details}, we provide additional implementation, training and evaluation details, including a complete summary of the datasets that were involved in this paper.
Finally, Section~\ref{sec:supp_applications} provides additional material on downstream applications.

\section{Additional Results}
\label{sec:supp_additional_results}

\subsection{Qualitative results}
\label{sec:supp_qualitative}
%%%%%% 
% Generic function to generate a figure overlaid crop
\newcommand{\generateFigure}[5]{
    {
        \begin{tikzpicture}
            \node[anchor=south west,inner sep=0] (image) at (0,0) {
                \begin{overpic}[width=0.195\textwidth]{fig_supp/qualitative/#1.jpg}
                \end{overpic}
            };
            \begin{scope}[x={(image.south east)},y={(image.north west)}]
            \draw[red,thick] (#2,#3) rectangle (#4, #5);
            \end{scope}
        \end{tikzpicture}
    }
}
% Generic function to generate an inset below the figure
\newcommand{\generateInsetBelow}[5]{
    \adjincludegraphics[width=0.195\textwidth,trim={{#2\width} {#3\height} {#4\width} {#5\height}},clip,cfbox=Red 0.25mm -0.25mm]{fig_supp/qualitative/#1.jpg}
}
% Recursive methods to populate a tabular row with figures and insets
\newcommand{\generateFigureFromList}[6]{%
    \recfna{#1}{#2,\relax}{#3}{#4}{#5}{#6}%
}
\def\recfna#1#2#3#4#5#6#7{%
    \expandafter\recfnb\expandafter#2#1\relax{#3}{#4}{#5}{#6}
}
\def\recfnb#1,#2\relax#3\relax#4#5#6#7{%
    \generateFigure{#3#1}{#4}{#5}{#6}{#7} &%
    \ifx\relax#2\relax%
    \else%
    \expandafter\recfna\expandafter{#3}{#2\relax}{#4}{#5}{#6}{#7}%
    \fi%
}
\newcommand{\generateInsetFromList}[6]{%
    \recfnc{#1}{#2,\relax}{#3}{#4}{#5}{#6}%
}
\def\recfnc#1#2#3#4#5#6#7{%
    \expandafter\recfnd\expandafter#2#1\relax{#3}{#4}{#5}{#6}
}
\def\recfnd#1,#2\relax#3\relax#4#5#6#7{%
    \generateInsetBelow{#3#1}{#4}{#5}{#6}{#7} &%
    \ifx\relax#2\relax%
    \else%
    \expandafter\recfnc\expandafter{#3}{#2\relax}{#4}{#5}{#6}{#7}%
    \fi%
}
%%%%%% 

\begin{figure}[htbp]
    \centering
    \scriptsize
    \begin{tabular}{@{}c@{\hspace{0.05mm}}c@{\hspace{0.05mm}}c@{\hspace{0.05mm}}c@{\hspace{0.05mm}}cc@{}}
        Input Image & Depth Pro (Ours) & Depth Anything v2 & Marigold & Metric3D v2 &\\
 %       
        % 28058219806_28e05ff24a_o
        \generateFigureFromList{28058219806\_28e05ff24a\_o}{{},\_depthpro\_pv5i9yzjef,\_depth\_anything_v2_relative,\_marigold,\_metric3d\_v2g}{0.3}{0.4}{0.65}{0.55}&\\
        \generateInsetFromList{28058219806\_28e05ff24a\_o}{{},\_depthpro\_pv5i9yzjef,\_depth\_anything_v2_relative,\_marigold,\_metric3d\_v2g}{0.3}{0.4}{0.35}{0.45}&\vspace{4mm}\\
%
        % 49867339188_08073f4b76_o
        \generateFigureFromList{49867339188\_08073f4b76\_o}{{},\_depthpro\_pv5i9yzjef,\_depth\_anything_v2_relative,\_marigold,\_metric3d\_v2g}{0.3}{0.22}{0.45}{0.32}&\\
        \generateInsetFromList{49867339188\_08073f4b76\_o}{{},\_depthpro\_pv5i9yzjef,\_depth\_anything_v2_relative,\_marigold,\_metric3d\_v2g}{0.3}{0.22}{0.55}{0.68}&\vspace{4mm}\\
%
        % m_f0cc3919
        \generateFigureFromList{m_f0cc3919}{{},\_depthpro\_pv5i9yzjef,\_depth\_anything_v2_relative,\_marigold,\_metric3d\_v2g}{0.40}{0.85}{0.55}{0.95}&\\
        \generateInsetFromList{m_f0cc3919}{{},\_depthpro\_pv5i9yzjef,\_depth\_anything_v2_relative,\_marigold,\_metric3d\_v2g}{0.40}{0.85}{0.45}{0.05}&\vspace{4mm}\\
        %
        % 16992510572_8a6ff27398_o
        \generateFigureFromList{16992510572\_8a6ff27398\_o}{{},\_depthpro\_pv5i9yzjef,\_depth\_anything_v2_relative,\_marigold,\_metric3d\_v2g}{0.35}{0.5}{0.85}{0.7}&\\
        \generateInsetFromList{16992510572\_8a6ff27398\_o}{{},\_depthpro\_pv5i9yzjef,\_depth\_anything_v2_relative,\_marigold,\_metric3d\_v2g}{0.35}{0.5}{0.15}{0.3}&\\
    \end{tabular}
  \caption{Zero-shot results of Depth Pro, Marigold~\citep{Ke2024:CVPR}, Metric3D v2~\citep{Hu2024:arxiv}, and Depth Anything v2~\citep{Yang2024:arxiv}  on images from Unsplash~\citep{Li2022:IJCV}, AM-2k~\citep{Li2022:IJCV}, and DIS-5k~\citep{Qin2022:ECCV}.}
  \label{fig:qualitative_supp1}
\end{figure}

\begin{figure}[htbp]
    \centering
    \scriptsize
    \begin{tabular}{@{}c@{\hspace{0.05mm}}c@{\hspace{0.05mm}}c@{\hspace{0.05mm}}c@{\hspace{0.05mm}}cc@{}}
        Input Image & Depth Pro (Ours) & Depth Anything v2 & Marigold & Metric3D v2 &\\
%
        % 5821282211_201cefeaf2_o
        \generateFigureFromList{5821282211_201cefeaf2_o}{{},\_depthpro\_pv5i9yzjef,\_depth\_anything_v2_relative,\_marigold,\_metric3d\_v2g}{0.5}{0.3}{0.8}{0.5}&\\
        \generateInsetFromList{5821282211_201cefeaf2_o}{{},\_depthpro\_pv5i9yzjef,\_depth\_anything_v2_relative,\_marigold,\_metric3d\_v2g}{0.5}{0.3}{0.2}{0.5}&\vspace{4mm}\\
        %
        % zg9KLvNHxvI
        \generateFigureFromList{zg9KLvNHxvI}{{},\_depthpro\_pv5i9yzjef,\_depth\_anything_v2_relative,\_marigold,\_metric3d\_v2g}{0.3}{0.35}{0.55}{0.55}&\\
        \generateInsetFromList{zg9KLvNHxvI}{{},\_depthpro\_pv5i9yzjef,\_depth\_anything_v2_relative,\_marigold,\_metric3d\_v2g}{0.3}{0.35}{0.45}{0.45}&\vspace{4mm}\\
        %
        % IMG_20210520_205442
        \generateFigureFromList{IMG\_20210520\_205442}{{},\_depthpro\_pv5i9yzjef,\_depth\_anything_v2_relative,\_marigold,\_metric3d\_v2g}{0.3}{0.45}{0.55}{0.55}&\\
        \generateInsetFromList{IMG\_20210520\_205442}{{},\_depthpro\_pv5i9yzjef,\_depth\_anything_v2_relative,\_marigold,\_metric3d\_v2g}{0.3}{0.45}{0.45}{0.45}&\vspace{4mm}\\
        %
        % 1381120202_9dff6987b2_o
        \generateFigureFromList{1381120202\_9dff6987b2\_o}{{},\_depthpro\_pv5i9yzjef,\_depth\_anything_v2_relative,\_marigold,\_metric3d\_v2g}{0.4}{0.15}{0.7}{0.35}&\\
        \generateInsetFromList{1381120202\_9dff6987b2\_o}{{},\_depthpro\_pv5i9yzjef,\_depth\_anything_v2_relative,\_marigold,\_metric3d\_v2g}{0.4}{0.15}{0.3}{0.65}&\vspace{4mm}\\
    \end{tabular}
  \caption{Zero-shot results of Depth Pro, Marigold~\citep{Ke2024:CVPR}, Metric3D v2~\citep{Hu2024:arxiv}, and Depth Anything v2~\citep{Yang2024:arxiv} on images from Unsplash~\citep{Li2022:IJCV}, AM-2k~\citep{Li2022:IJCV}, and DIS-5k~\citep{Qin2022:ECCV}.}
  \label{fig:qualitative_supp2}
\end{figure}

\begin{figure}[htbp]
    \centering
    \scriptsize
    \begin{tabular}{@{}c@{\hspace{0.05mm}}c@{\hspace{0.05mm}}c@{\hspace{0.05mm}}c@{\hspace{0.05mm}}cc@{}}
        Input Image & Depth Pro (Ours) & Depth Anything v2 & Marigold & Metric3D v2 &\\
%
        % zgFh2mFVsZQ
        \generateFigureFromList{zgFh2mFVsZQ}{{},\_depthpro\_pv5i9yzjef,\_depth\_anything_v2_relative,\_marigold,\_metric3d\_v2g}{0.2}{0.4}{0.6}{0.5}&\\
        \generateInsetFromList{zgFh2mFVsZQ}{{},\_depthpro\_pv5i9yzjef,\_depth\_anything_v2_relative,\_marigold,\_metric3d\_v2g}{0.2}{0.4}{0.4}{0.5}&\vspace{4mm}\\
%
        % zfr_U0ApaOQ
        \generateFigureFromList{zfr\_U0ApaOQ}{{},\_depthpro\_pv5i9yzjef,\_depth\_anything_v2_relative,\_marigold,\_metric3d\_v2g}{0.2}{0.5}{0.6}{0.7}&\\
        \generateInsetFromList{zfr\_U0ApaOQ}{{},\_depthpro\_pv5i9yzjef,\_depth\_anything_v2_relative,\_marigold,\_metric3d\_v2g}{0.2}{0.5}{0.4}{0.3}&\vspace{4mm}\\
%
        % m_f1babe9e
        \generateFigureFromList{m_f1babe9e}{{},\_depthpro\_pv5i9yzjef,\_depth\_anything_v2_relative,\_marigold,\_metric3d\_v2g}{0.1}{0.35}{0.5}{0.45}&\\
        \generateInsetFromList{m_f1babe9e}{{},\_depthpro\_pv5i9yzjef,\_depth\_anything_v2_relative,\_marigold,\_metric3d\_v2g}{0.1}{0.35}{0.5}{0.55}&\vspace{4mm}\\
%
        % m_8996956d
        \generateFigureFromList{m_8996956d}{{},\_depthpro\_pv5i9yzjef,\_depth\_anything_v2_relative,\_marigold,\_metric3d\_v2g}{0.4}{0.3}{0.7}{0.5}&\\
        \generateInsetFromList{m_8996956d}{{},\_depthpro\_pv5i9yzjef,\_depth\_anything_v2_relative,\_marigold,\_metric3d\_v2g}{0.4}{0.3}{0.3}{0.5}&\vspace{4mm}\\
    \end{tabular}
  \caption{Zero-shot results of Depth Pro, Marigold~\citep{Ke2024:CVPR}, Metric3D v2~\citep{Hu2024:arxiv}, and Depth Anything v2~\citep{Yang2024:arxiv}  on images from Unsplash~\citep{Li2022:IJCV}, AM-2k~\citep{Li2022:IJCV}, and DIS-5k~\citep{Qin2022:ECCV}.}
  \label{fig:qualitative_supp3}
\end{figure}

We provide additional qualitative results of Depth Pro, Marigold~\citep{Ke2024:CVPR}, Metric3D v2~\citep{Hu2024:arxiv}, and Depth Anything v2~\citep{Yang2024:arxiv} on in-the-wild images from AM-2k~\citep{Li2022:IJCV}, DIS-5k~\citep{Qin2022:ECCV}, and Unsplash\footnote{\url{https://www.unsplash.com}} in \Fig~\ref{fig:qualitative_supp1}, \Fig~\ref{fig:qualitative_supp2}, and \Fig~\ref{fig:qualitative_supp3}.
Fine details are repeatedly missed by Metric3D~v2 and Depth Anything v2. Marigold reproduces finer details than Metric3D~v2 and Depth Anything v2 but commonly yields noisy predictions.

\subsection{Zero-shot metric depth}
\label{sec:supp_zero_shot}
Expanding on the summary in \Tab{\ref{tab:sota_0shot_metric}}, we provide additional results for zero-shot metric depth estimation in \Tab~\ref{tab:eval_all_metric_0shot}. We report results on Booster~\citep{Ramirez2024:PAMI}, Middlebury~\citep{Scharstein2014:GCPR}, Sun-RGBD~\citep{Song2015:CVPR}, ETH3D~\citep{Schps2017:CVPR}, nuScenes~\citep{Caesar2020:CVPR}, and Sintel~\citep{Butler2012:ECCV}.
Our baselines include Depth Anything~\citep{Yang2024:CVPR} and Depth Anything v2~\citep{Yang2024:arxiv}, Metric3D~\citep{Yin2023:ICCV} and Metric3D v2~\citep{Hu2024:arxiv}, PatchFusion~\citep{Li2024:CVPR}, UniDepth~\citep{Piccinelli2024:CVPR}, ZeroDepth~\citep{Guizilini2023:ICCV}, and ZoeDepth~\citep{Bhat2023:arxiv}.
To preserve the zero-shot setting, we do not report results for models that were trained on the same dataset as the evaluation dataset.
We report commonly used metrics in the depth estimation literature, namely $\mathit{AbsRel}$, $\mathit{Log}_{10}$~\citep{Saxena2009:TPAMI}, $\delta_1$, $\delta_2$ and $\delta_3$ scores~\citep{Ladicky2014:CVPR}, as well as point-cloud metrics~\citep{Spencer2022:TMLR}.
Due to the high resolution of Booster images, we were not able to obtain point-cloud metrics in reasonable time.

\begin{table}[htbp]
    \caption{\textbf{Additional zero-shot metric depth evaluation.} We report additional metrics used in the depth estimation literature, namely $\mathit{AbsRel}$~\citep{Ladicky2014:CVPR}, $\mathit{Log}_{10}$,  $\delta_2$ and $\delta_3$ scores, as well as point-cloud metrics~\citep{Spencer2022:TMLR} on Booster~\citep{Ramirez2024:PAMI}, Middlebury~\citep{Scharstein2014:GCPR}, Sun-RGBD~\citep{Song2015:CVPR}, ETH3D~\citep{Schps2017:CVPR}, nuScenes~\citep{Caesar2020:CVPR}, and Sintel~\citep{Butler2012:ECCV}. For fair comparison, all reported results were reproduced in our environment.}
    \label{tab:eval_all_metric_0shot}
    \centering
    \scriptsize
\begin{tabular}{lllllllll}
\toprule
 \textbf{NuScenes} & AbsRel$\downarrow$ & Log$_{10}\downarrow$ & $\delta_2\uparrow$ & $\delta_3\uparrow$ & SI-Log$\downarrow$ & PC-CD$\downarrow$ & PC-F$\uparrow$ & PC-IoU$\uparrow$ \\

\midrule
 DepthAnything~\citep{Yang2024:CVPR} & 0.453 & 0.151 & 73.876 & \cellthird{90.301} & \cellthird{28.153} & 24.146 & 0.007 & 0.004 \\

DepthAnything v2~\citep{Yang2024:arxiv} & 0.614 & 0.326 & 31.837 & 47.265 & 29.737 & 37.516 & 0.008 & 0.004 \\
 Metric3D ~\citep{Yin2023:ICCV} & 0.422 & 0.132 & 77.220 & 83.605 & 33.827 & 29.284 & 0.007 & 0.004 \\
 Metric3D v2 ~\citep{Hu2024:arxiv} & \cellsecond{0.197} & \cellsecond{0.080} & \cellfirst{93.252} & \cellsecond{95.736} & \cellsecond{27.032} & \cellsecond{14.876} & \cellthird{0.008} & \cellsecond{0.004} \\
 PatchFusion ~\citep{Li2024:CVPR} & 0.392 & 0.226 & 48.742 & 76.035 & 31.171 & \cellthird{20.836} & 0.006 & 0.003 \\
 UniDepth ~\citep{Piccinelli2024:CVPR} & \cellfirst{0.138} & \cellfirst{0.060} & \cellsecond{93.006} & \cellfirst{96.415} & \cellfirst{21.801} & \cellfirst{11.629} & \cellsecond{0.009} & \cellthird{0.004} \\
 ZeroDepth ~\citep{Guizilini2023:ICCV} & \cellthird{0.237} & \cellthird{0.121} & \cellthird{82.596} & 89.908 & 30.703 & 23.348 & 0.007 & 0.004 \\
 ZoeDepth ~\citep{Bhat2023:arxiv} & 0.498 & 0.182 & 64.947 & 82.704 & 31.501 & 39.183 & 0.006 & 0.003 \\

\midrule
 Depth Pro (Ours) & 0.287 & 0.164 & 73.836 & 84.252 & 29.548 & 22.480 & \cellfirst{0.010} & \cellfirst{0.005} \\

\bottomrule
\end{tabular}

\  \\

\begin{tabular}{lllllllll}
\toprule
 \textbf{Sintel} & AbsRel$\downarrow$ & Log$_{10}\downarrow$ & $\delta_2\uparrow$ & $\delta_3\uparrow$ & SI-Log$\downarrow$ & PC-CD$\downarrow$ & PC-F$\uparrow$ & PC-IoU$\uparrow$ \\

\midrule
 DepthAnything~\citep{Yang2024:CVPR} & 3.973 & 0.559 & 15.418 & 27.281 & \cellthird{35.771} & \cellthird{38.592} & 0.057 & 0.030 \\

DepthAnything v2~\citep{Yang2024:arxiv} & 2.226 & 0.494 & 18.696 & 33.820 & 41.923 & 54.931 & 0.057 & 0.031 \\
 Metric3D ~\citep{Yin2023:ICCV} & 1.733 & 0.387 & 32.375 & 44.793 & 48.605 & 45.858 & 0.056 & 0.031 \\
 Metric3D v2 ~\citep{Hu2024:arxiv} & \cellfirst{0.370} & \cellfirst{0.216} & \cellfirst{62.915} & \cellfirst{76.866} & \cellfirst{25.312} & \cellsecond{34.790} & \cellthird{0.091} & \cellthird{0.051} \\
 PatchFusion ~\citep{Li2024:CVPR} & \cellthird{0.617} & 0.391 & 35.515 & 51.443 & 36.806 & 44.615 & 0.077 & 0.045 \\
 UniDepth ~\citep{Piccinelli2024:CVPR} & 0.869 & \cellthird{0.301} & \cellthird{35.722} & \cellthird{57.256} & 42.837 & \cellfirst{32.338} & \cellsecond{0.098} & \cellsecond{0.057} \\
 ZeroDepth ~\citep{Guizilini2023:ICCV} & 0.703 & 0.491 & 25.629 & 37.076 & 50.839 & 76.274 & 0.052 & 0.029 \\
 ZoeDepth ~\citep{Bhat2023:arxiv} & 0.946 & 0.392 & 22.698 & 44.969 & 40.217 & 52.301 & 0.085 & 0.049 \\

\midrule
 Depth Pro (Ours) & \cellsecond{0.508} & \cellsecond{0.230} & \cellsecond{59.247} & \cellsecond{71.138} & \cellsecond{27.494} & 41.968 & \cellfirst{0.121} & \cellfirst{0.073} \\

\bottomrule
\end{tabular}

\  \\

\begin{tabular}{lllllllll}
\toprule
 \textbf{Sun-RGBD} & AbsRel$\downarrow$ & Log$_{10}\downarrow$ & $\delta_2\uparrow$ & $\delta_3\uparrow$ & SI-Log$\downarrow$ & PC-CD$\downarrow$ & PC-F$\uparrow$ & PC-IoU$\uparrow$ \\

\midrule
 DepthAnything~\citep{Yang2024:CVPR} & \cellthird{0.114} & \cellthird{0.053} & \cellsecond{98.811} & \cellsecond{99.770} & 8.038 & \cellthird{0.034} & 0.160 & 0.090 \\

DepthAnything v2~\citep{Yang2024:arxiv} & 0.182 & 0.070 & 97.645 & 99.462 & 8.390 & 0.045 & 0.169 & 0.096 \\
 Metric3D ~\citep{Yin2023:ICCV} & 1.712 & 0.382 & 26.999 & 34.116 & 20.262 & 0.506 & 0.060 & 0.032 \\
 Metric3D v2 ~\citep{Hu2024:arxiv} & 0.156 & 0.076 & 96.348 & \cellthird{99.548} & \cellsecond{7.433} & \cellsecond{0.025} & \cellthird{0.179} & \cellthird{0.102} \\
 PatchFusion ~\citep{Li2024:CVPR} & 0.466 & 0.961 & 60.145 & 60.651 & 69.647 & 331.477 & 0.052 & 0.027 \\
 UniDepth ~\citep{Piccinelli2024:CVPR} & \cellfirst{0.087} & \cellfirst{0.037} & \cellfirst{99.330} & \cellfirst{99.804} & \cellfirst{6.968} & \cellfirst{0.020} & \cellfirst{0.294} & \cellfirst{0.183} \\
 ZoeDepth ~\citep{Bhat2023:arxiv} & 0.123 & 0.053 & 97.954 & 99.505 & 8.964 & 0.048 & 0.135 & 0.075 \\

\midrule
 Depth Pro (Ours) & \cellsecond{0.113} & \cellsecond{0.049} & \cellthird{98.506} & 99.547 & \cellthird{7.841} & 0.039 & \cellsecond{0.179} & \cellsecond{0.103} \\

\bottomrule
\end{tabular}

\  \\

\begin{tabular}{lllllllll}
\toprule
 \textbf{ETH3D} & AbsRel$\downarrow$ & Log$_{10}\downarrow$ & $\delta_2\uparrow$ & $\delta_3\uparrow$ & SI-Log$\downarrow$ & PC-CD$\downarrow$ & PC-F$\uparrow$ & PC-IoU$\uparrow$ \\

\midrule
 DepthAnything~\citep{Yang2024:CVPR} & 1.682 & 0.380 & 19.784 & 31.057 & 10.903 & 0.072 & 0.172 & 0.114 \\

DepthAnything v2~\citep{Yang2024:arxiv} & 0.370 & \cellthird{0.173} & \cellthird{64.657} & \cellthird{86.256} & \cellthird{9.683} & \cellthird{0.042}  & 0.330 & 0.233 \\
 Metric3D ~\citep{Yin2023:ICCV} & 0.859 & 0.240 & 49.291 & 57.573 & 14.541 & 0.072 & 0.303 & 0.219 \\
 Metric3D v2 ~\citep{Hu2024:arxiv} & \cellfirst{0.124} & \cellfirst{0.053} & \cellfirst{99.553} & \cellfirst{99.900} & \cellfirst{6.197} & 0.083 & \cellsecond{0.466} & \cellsecond{0.358} \\
 PatchFusion ~\citep{Li2024:CVPR} & \cellsecond{0.256} & \cellsecond{0.106} & \cellsecond{88.378} & \cellsecond{97.306} & 11.023 & \cellsecond{0.042} & 0.209 & 0.135 \\
 UniDepth ~\citep{Piccinelli2024:CVPR} & 0.457 & 0.186 & 57.670 & 81.483 & \cellsecond{7.729} & \cellfirst{0.031} & \cellthird{0.409} & \cellthird{0.305} \\
 ZoeDepth ~\citep{Bhat2023:arxiv} & 0.500 & 0.176 & 64.452 & 81.434 & 13.250 &0.078 & 0.127 & 0.082 \\

\midrule
 Depth Pro (Ours) & \cellthird{0.327} & 0.193 & 61.309 & 71.228 & 10.170 & 0.094 & \cellfirst{0.487} & \cellfirst{0.398} \\

\bottomrule
\end{tabular}

\  \\

\begin{tabular}{lllllllll}
\toprule
 \textbf{Middlebury} & AbsRel$\downarrow$ & Log$_{10}\downarrow$ & $\delta_2\uparrow$ & $\delta_3\uparrow$ & SI-Log$\downarrow$ & PC-CD$\downarrow$ & PC-F$\uparrow$ & PC-IoU$\uparrow$ \\

\midrule
 DepthAnything~\citep{Yang2024:CVPR} & 0.273 & 0.149 & 69.619 & 86.060 & 12.420 & 0.102 & 0.103 & 0.055 \\

DepthAnything v2~\citep{Yang2024:arxiv} & 0.262 & 0.141 & 72.074 & 90.549 & 9.639 & \cellsecond{0.063} & 0.127 & 0.069 \\
 Metric3D ~\citep{Yin2023:ICCV} & 1.251 & 0.305 & 37.528 & 58.733 & 12.091 & \cellthird{0.069} & 0.069 & 0.036 \\
 Metric3D v2 ~\citep{Hu2024:arxiv} & 0.450 & 0.152 & 73.321 & 88.610 & \cellfirst{5.519} & \cellfirst{0.022} & \cellsecond{0.215} & \cellsecond{0.122} \\
 PatchFusion ~\citep{Li2024:CVPR} & \cellsecond{0.250} & \cellsecond{0.108} & \cellsecond{87.166} & \cellsecond{98.154} & 14.641 & 0.319 & 0.084 & 0.044 \\
 UniDepth ~\citep{Piccinelli2024:CVPR} & 0.324 & 0.127 & \cellthird{80.047} & \cellfirst{99.621} & \cellsecond{7.379} & 0.113 & \cellfirst{0.221} & \cellfirst{0.129} \\
 ZeroDepth ~\citep{Guizilini2023:ICCV} & 0.377 & 0.179 & 67.060 & 78.952 & 14.482 & 0.232 & 0.052 & 0.027 \\
 ZoeDepth ~\citep{Bhat2023:arxiv} & \cellfirst{0.214} & \cellthird{0.115} & 77.683 & 90.860 & 10.448 & 0.069 & 0.114 & 0.062 \\

\midrule
 Depth Pro (Ours) & \cellthird{0.251} & \cellfirst{0.089} & \cellfirst{93.169} & \cellthird{96.401} & \cellthird{8.610} & 0.107 & \cellthird{0.161} & \cellthird{0.091} \\

\bottomrule
\end{tabular}

\  \\

\begin{tabular}{llllll}
\toprule
 \textbf{Booster} & AbsRel$\downarrow$ & Log$_{10}\downarrow$ & $\delta_2\uparrow$ & $\delta_3\uparrow$ & SI-Log$\downarrow$ \\

\midrule
 DepthAnything~\citep{Yang2024:CVPR} & \cellsecond{0.317} & \cellsecond{0.114} & \cellfirst{79.615} & \cellsecond{95.228} & 10.507 \\

DepthAnything v2~\citep{Yang2024:arxiv} & \cellfirst{0.315} & \cellfirst{0.110} & \cellthird{76.239} & \cellthird{94.276} & \cellthird{7.056} \\
 Metric3D ~\citep{Yin2023:ICCV} & 1.332 & 0.346 & 13.073 & 33.975 & 10.631 \\
 Metric3D v2 ~\citep{Hu2024:arxiv} & 0.417 & 0.140 & 75.783 & 92.833 & \cellfirst{3.932} \\
 PatchFusion ~\citep{Li2024:CVPR} & 0.719 & 0.213 & 49.387 & 72.892 & 14.128 \\
 UniDepth ~\citep{Piccinelli2024:CVPR} & 0.500 & 0.166 & 60.904 & 89.213 & 7.436 \\
 ZoeDepth ~\citep{Bhat2023:arxiv} & 0.610 & 0.195 & 52.655 & 75.508 & 10.551 \\

\midrule
 Depth Pro (Ours) & \cellthird{0.336} & \cellthird{0.118} & \cellsecond{79.429} & \cellfirst{96.524} & \cellsecond{4.616} \\

\bottomrule
\end{tabular}
\end{table}

\subsection{Runtime}
\label{sec:supp_runtime}
To assess the latency of our approach in comparison to baselines, we test all approaches on images of varying sizes and report results in \Tab~\ref{tab:runtime}. We pick common image resolutions (VGA: 640$\times$480, HD: 1920$\times$1080, 4K: 4032$\times$3024) and measure each method's average runtime for processing an image of the given size.
All reported runtimes are reproduced in our environment and include preprocessing, eventual resizing (for methods operating at a fixed internal resolution), and inference of each model.
We further report the parameter counts and flops (at HD resolution) for each method as measured with the fvcore package.

Among all approaches with a fixed output resolution, Depth Pro has the highest native output resolution, processing more than 3 times as many pixels as the next highest, Metric3D v2~\citep{Hu2024:arxiv}. Yet Depth Pro has less than half the parameter count and requires only a third of the runtime compared to Metric3D v2.

The variable-resolution approaches (PatchFusion~\citep{Li2024:CVPR} and ZeroDepth~\citep{Guizilini2023:ICCV}) have considerably larger runtime, with the faster model, ZeroDepth, taking almost 4 times as long as Depth Pro, even for small VGA images.

\begin{table}[htbp]
  \scriptsize
  \centering
  \caption{\textbf{Model performance, measured on a V100-32G GPU.}
  We report runtimes in milliseconds (ms) on images of multiple sizes, as well as model parameter counts and flops. For fairness, the reported runtimes are reproduced in our environment. Entries are sorted by the native output resolution.}
    \label{tab:runtime}
  \begin{tabular}{@{}lccccrrrr@{}}
    Method &
    \multirow{2}{*}{\shortstack{Parameter\\count}} &
    Flops$_{\mathit{HD}}$$\downarrow$ &
    \multicolumn{2}{c}{\multirow{2}{*}{\shortstack{Native output\\resolution}} $\!\uparrow$} &
    $t_{\mathit{VGA}}$ (ms) $\downarrow$ &
    $t_{\mathit{HD}}$ (ms) $\downarrow$ &
    $t_{\mathit{4K}}$ (ms) $\downarrow$ \\
    \\
    \midrule
    DPT & 123M & - & 384 $\times$ 384 & = 0.15 MP & 33.2 & 30.6 & 27.8 \\
    ZoeDepth & 340M & - & 384 $\times$ 512 & = 0.20 MP &235.7 &235.1 & 235.4\\
    DepthAnything v2 &335M & 1827G & 518 $\times$ 518 & = 0.27 MP & 90.9 & 91.1 & 91.2 \\
    UniDepth & 347M & 630G & 462 $\times$ 616 & = 0.28 MP & 178.5 & 183.0 & 198.1 \\
    Metric3D & 203M & 477G & 480 $\times$ 1216 & = 0.58 MP & 217.9 & 263.8 & 398.1 \\
    Marigold & 949M & - & 768 $\times$ 768 & = 0.59 MP & 5174.3 & 4433.6 & 4977.6 \\
    Metric3D v2 & 1.378G & 6830G & 616 $\times$ 1064 & = 0.66 MP & 1299.6 & 1299.7 & 1390.2 \\
    \midrule
    PatchFusion & 203M & - & \multicolumn{2}{c}{Original (tile-based)} & 84012.0 &  84029.9 & 84453.9 \\
    ZeroDepth & 233M & 10862G & \multicolumn{2}{c}{Original} & 1344.3 & 8795.7 & 34992.2\\
    \midrule
    Depth Pro & 504M & 4370G & 1536 $\times$ 1536 & = 2.36 MP & 341.3 & 341.3 & 341.3 \\
  \bottomrule
  \end{tabular}
\end{table}

\subsection{Boundary experiments}
\label{sec:supp_additional_experiments}

\begin{figure}[t!]
\scriptsize
\centering
 \begin{tabular}{@{}c@{\hspace{1mm}}c@{\hspace{1mm}}c@{\hspace{1mm}}c@{\hspace{1mm}}c@{\hspace{1mm}}c@{}}
 & Image & $d$ & $\hat{d}$ & $\bigcup\limits_{(i,j)\in N}c_d(i,j)$& $\bigcup\limits_{(i,j)\in N}c_{\hat{d}}(i,j)$ \\
   \rotatebox[origin=c]{90}{Sintel~\citep{Butler2012:ECCV}} &
   \raisebox{-0.5\height}{\includegraphics[width=0.18\textwidth]{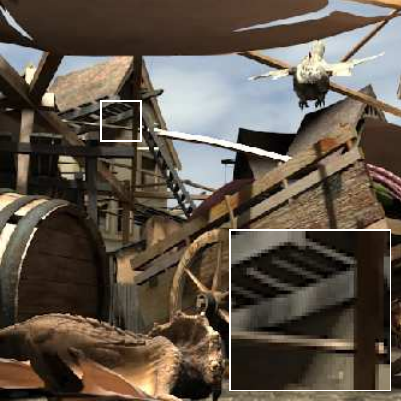}} &
  \raisebox{-0.5\height}{\includegraphics[width=0.18\textwidth]{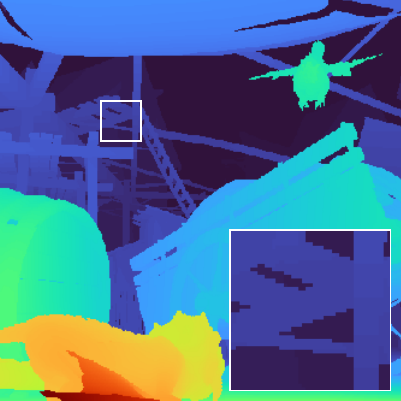}} &
  \raisebox{-0.5\height}{\includegraphics[width=0.18\textwidth]{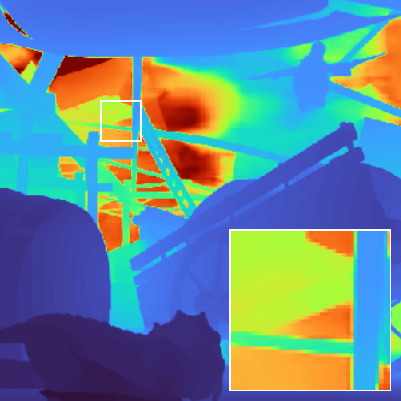}} &
  \raisebox{-0.5\height}{\includegraphics[width=0.18\textwidth]{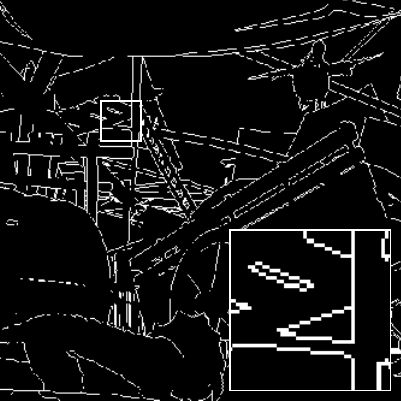}} &
  \raisebox{-0.5\height}{\includegraphics[width=0.18\textwidth]{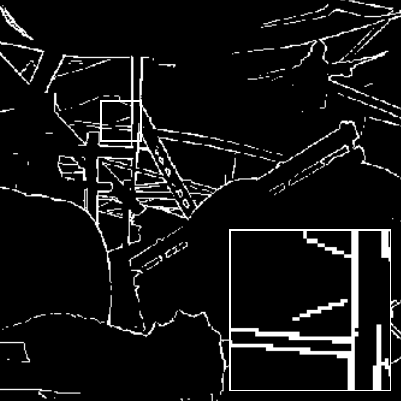}}
  \vspace{1mm}\\
   \rotatebox[origin=c]{90}{Spring~\citep{Mehl2023:CVPR}} &
   \raisebox{-0.5\height}{\includegraphics[width=0.18\textwidth]{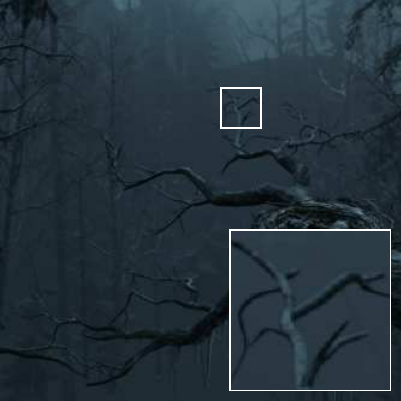}} &
  \raisebox{-0.5\height}{\includegraphics[width=0.18\textwidth]{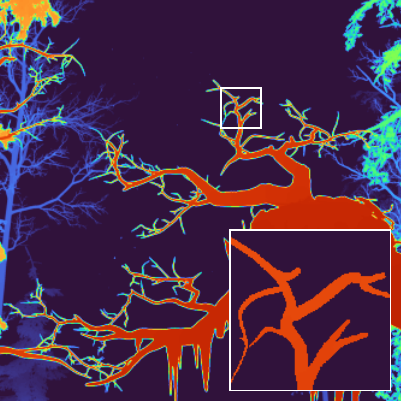}} &
  \raisebox{-0.5\height}{\includegraphics[width=0.18\textwidth]{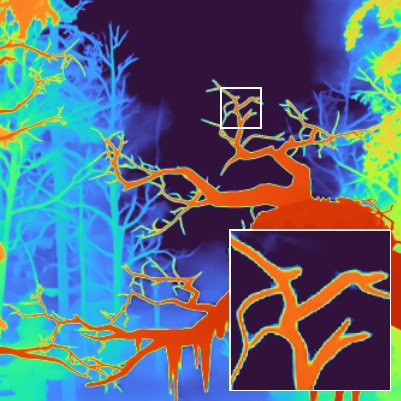}} &
  \raisebox{-0.5\height}{\includegraphics[width=0.18\textwidth]{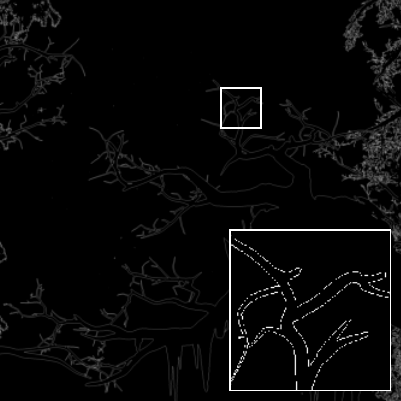}} &
  \raisebox{-0.5\height}{\includegraphics[width=0.18\textwidth]{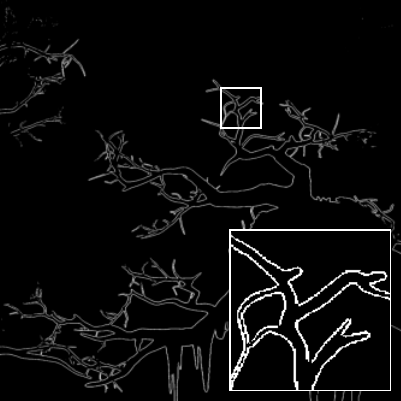}}
  \vspace{1mm}\\
   \rotatebox[origin=c]{90}{AM-2k~\citep{Li2022:IJCV}} &
   \raisebox{-0.5\height}{\includegraphics[width=0.18\textwidth]{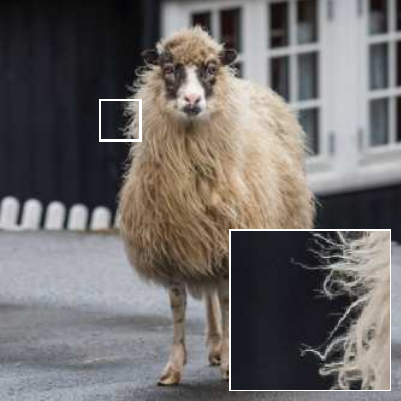}} &
  \raisebox{-0.5\height}{\includegraphics[width=0.18\textwidth]{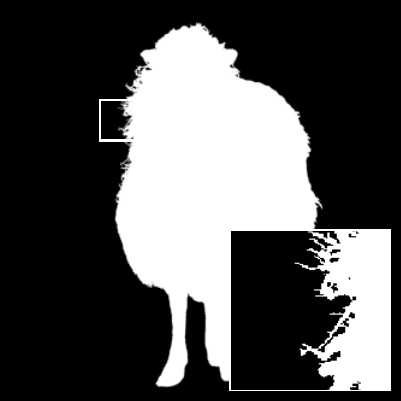}} &
  \raisebox{-0.5\height}{\includegraphics[width=0.18\textwidth]{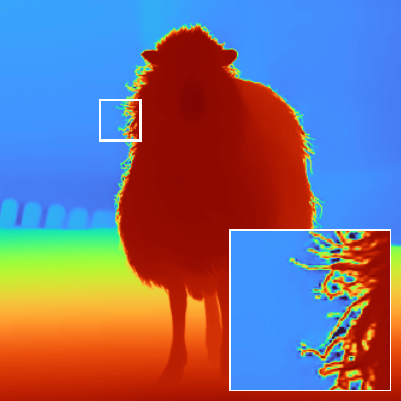}} &
  \raisebox{-0.5\height}{\includegraphics[width=0.18\textwidth]{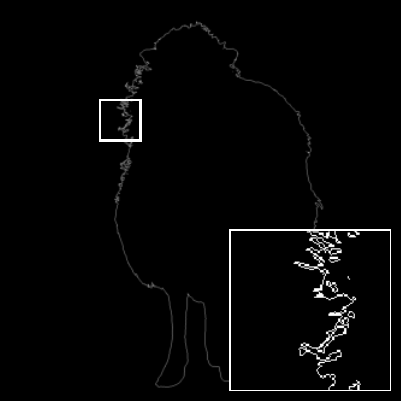}} &
  \raisebox{-0.5\height}{\includegraphics[width=0.18\textwidth]{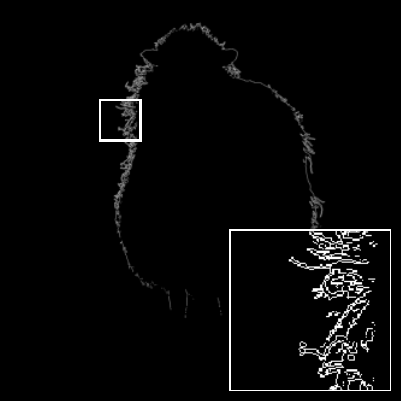}}
  \vspace{1mm}\\
   \rotatebox[origin=c]{90}{P3M-10k~\citep{Li2021:ACMMM}} &
   \raisebox{-0.5\height}{\includegraphics[width=0.18\textwidth]{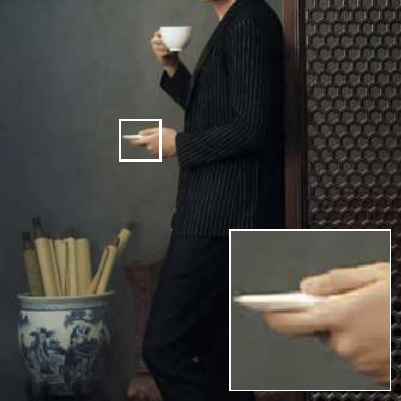}} &
  \raisebox{-0.5\height}{\includegraphics[width=0.18\textwidth]{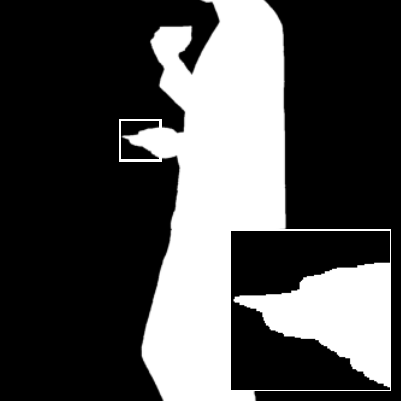}} &
  \raisebox{-0.5\height}{\includegraphics[width=0.18\textwidth]{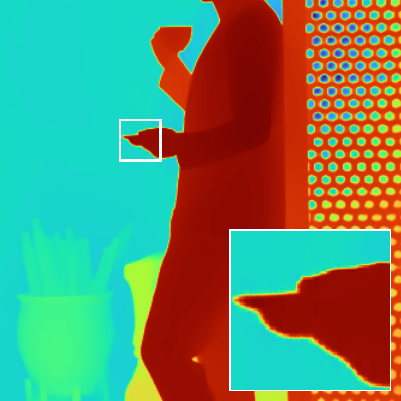}} &
  \raisebox{-0.5\height}{\includegraphics[width=0.18\textwidth]{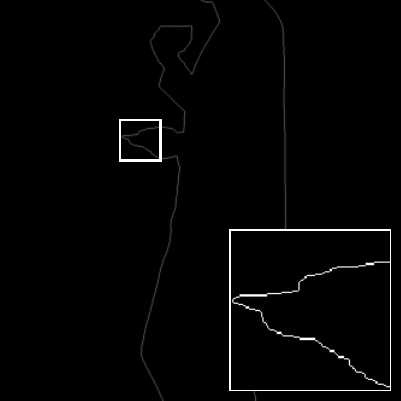}} &
  \raisebox{-0.5\height}{\includegraphics[width=0.18\textwidth]{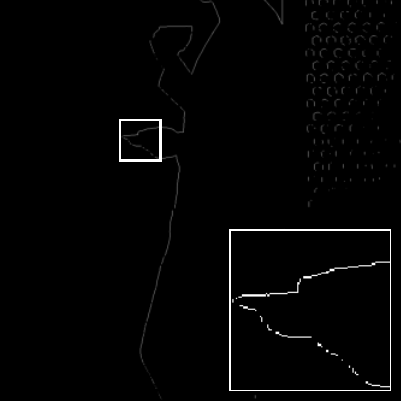}}
  \vspace{1mm}\\
    \rotatebox[origin=c]{90}{DIS-5k~\citep{Qin2022:ECCV}} &
   \raisebox{-0.5\height}{\includegraphics[width=0.18\textwidth]{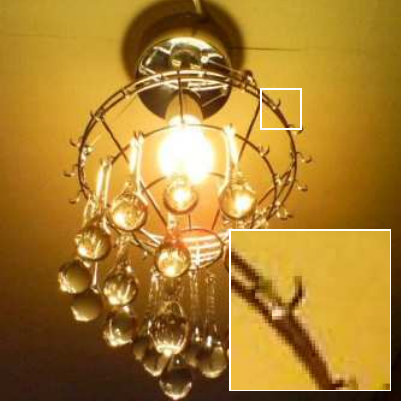}} &
  \raisebox{-0.5\height}{\includegraphics[width=0.18\textwidth]{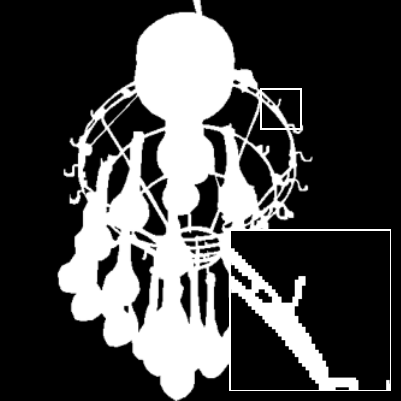}} &
  \raisebox{-0.5\height}{\includegraphics[width=0.18\textwidth]{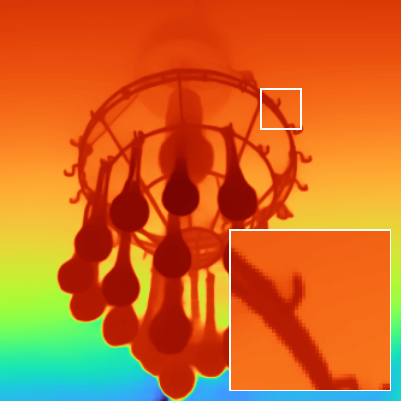}} &
  \raisebox{-0.5\height}{\includegraphics[width=0.18\textwidth]{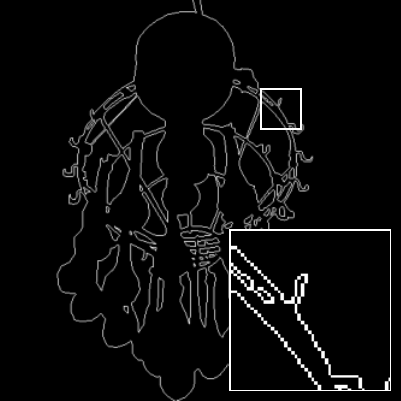}} &
  \raisebox{-0.5\height}{\includegraphics[width=0.18\textwidth]{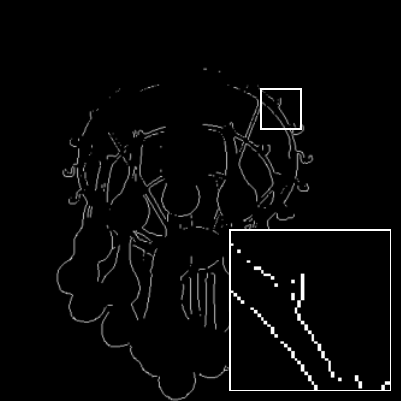}}
 \end{tabular}
\caption{\textbf{Evaluation metrics for sharp boundaries.} We  propose novel metrics to evaluate the sharpness of occlusion boundaries. The metrics can be computed on ground-truth depth maps (first two rows), and binary maps that can be derived from matting or segmentation datasets (subsequent rows). Each row shows a sample image, the ground truth for deriving occlusion boundaries, our prediction, ground-truth occluding contours, and occluding contours from the prediction. For these visualizations we set $t=15$.}
\label{fig:edge_metrics}
\end{figure}

\begin{figure}[thbp]
\scriptsize
\centering

\raisebox{-0.5\height}{\includegraphics[width=0.42\textwidth]{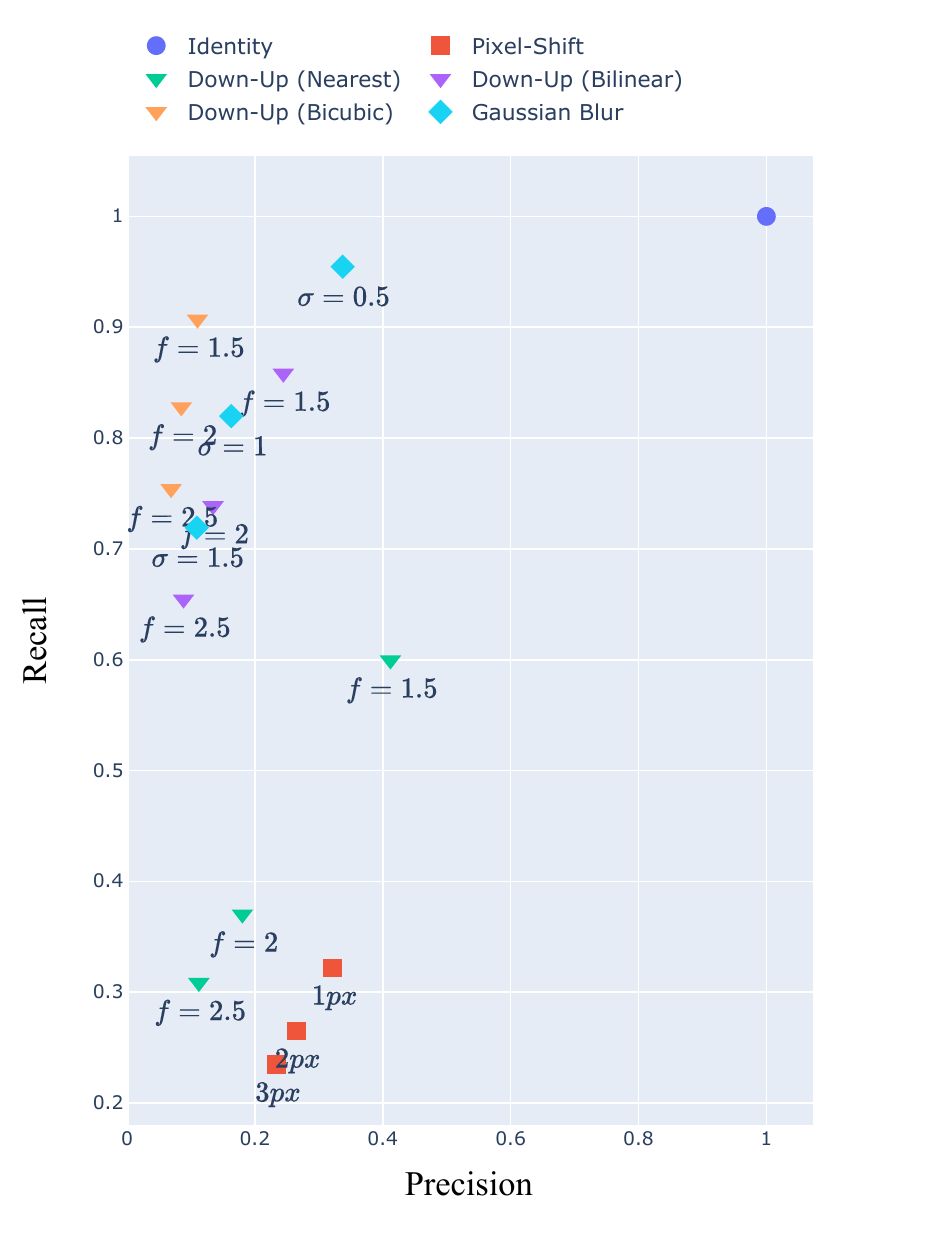}}
 \begin{tabular}{@{}c@{\hspace{1mm}}c@{\hspace{1mm}}c@{\hspace{1mm}}c@{\hspace{1mm}}c}
 & $d$ & $\hat{d}$ & $\bigcup\limits_{(i,j)\in N}c_d(i,j)$ & $\bigcup\limits_{(i,j)\in N}c_{\hat{d}}(i,j)$\\
   \rotatebox[origin=c]{90}{\parbox{1.3cm}{Gaussian\\Blur}} &
   \raisebox{-0.5\height}{\includegraphics[width=0.12\textwidth]{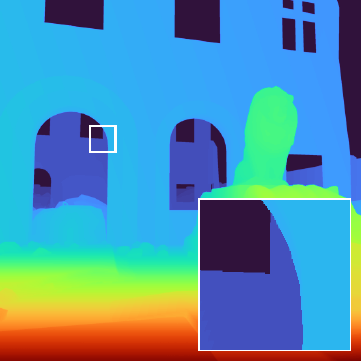}} &
  \raisebox{-0.5\height}{\includegraphics[width=0.12\textwidth]{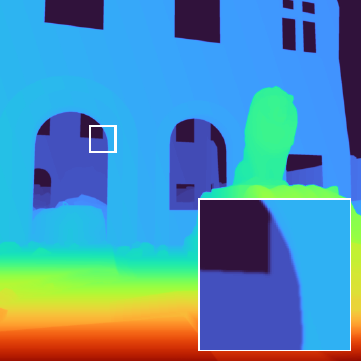}} &
  \raisebox{-0.5\height}{\includegraphics[width=0.12\textwidth]{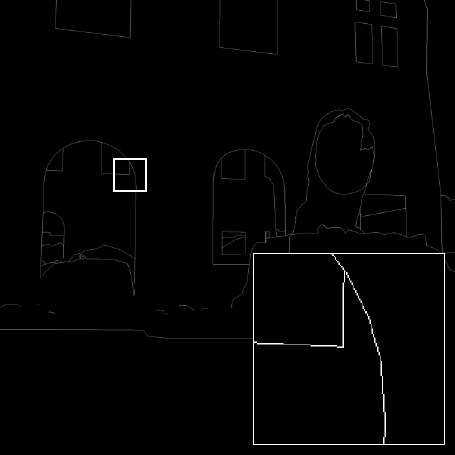}} &
  \raisebox{-0.5\height}{\includegraphics[width=0.12\textwidth]{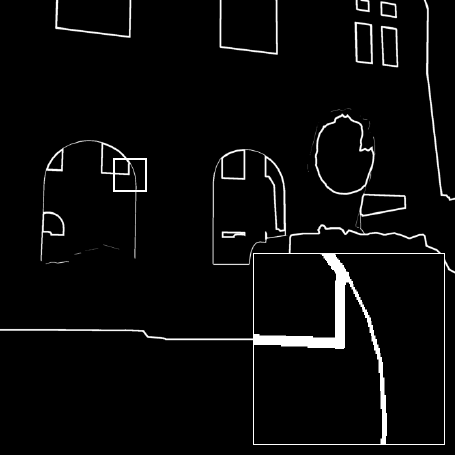}}
  \vspace{0.25em}\\
   \rotatebox[origin=c]{90}{Pixel-Shift} &
   \raisebox{-0.5\height}{\includegraphics[width=0.12\textwidth]{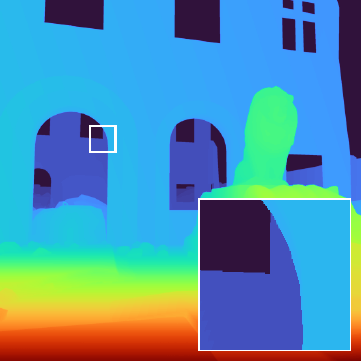}} &
  \raisebox{-0.5\height}{\includegraphics[width=0.12\textwidth]{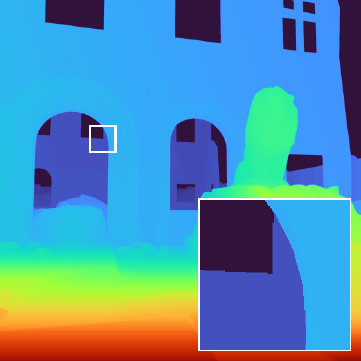}} &
  \raisebox{-0.5\height}{\includegraphics[width=0.12\textwidth]{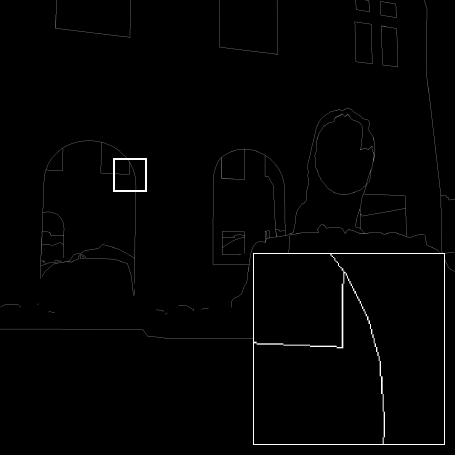}} &
  \raisebox{-0.5\height}{\includegraphics[width=0.12\textwidth]{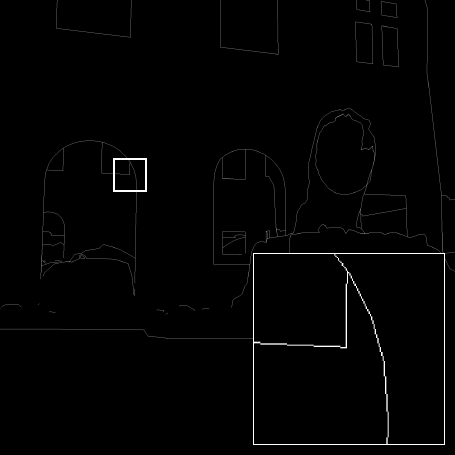}}
  \vspace{0.25em}\\
  \rotatebox[origin=c]{90}{\parbox{1.3cm}{Up-Down\\(nearest)}} &
   \raisebox{-0.5\height}{\includegraphics[width=0.12\textwidth]{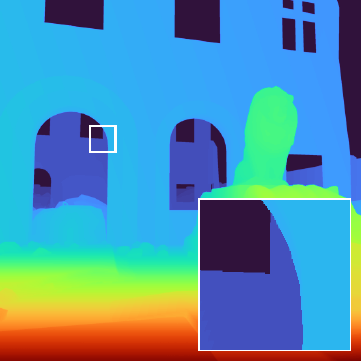}} &
  \raisebox{-0.5\height}{\includegraphics[width=0.12\textwidth]{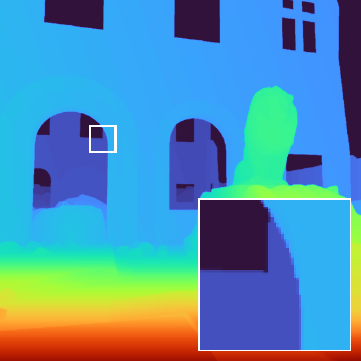}} &
  \raisebox{-0.5\height}{\includegraphics[width=0.12\textwidth]{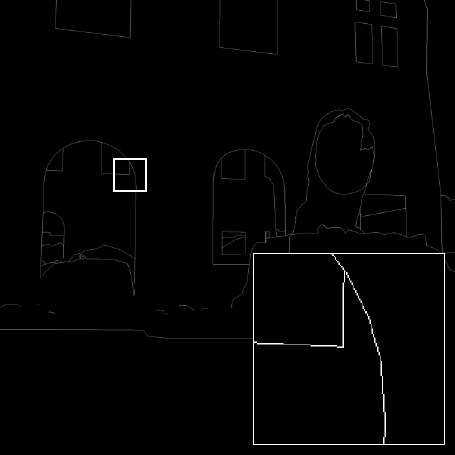}} &
  \raisebox{-0.5\height}{\includegraphics[width=0.12\textwidth]{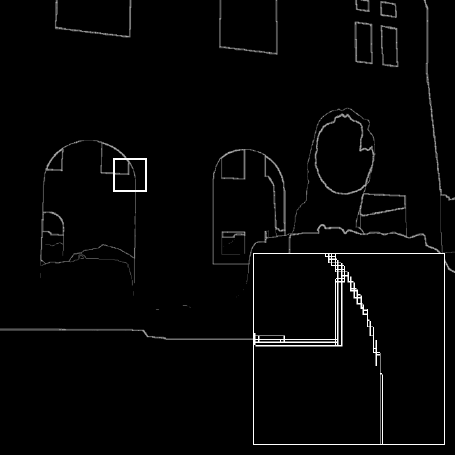}} \vspace{0.25em}\\
  \rotatebox[origin=c]{90}{\parbox{1.3cm}{Up-Down\\(bilinear)}} &
   \raisebox{-0.5\height}{\includegraphics[width=0.12\textwidth]{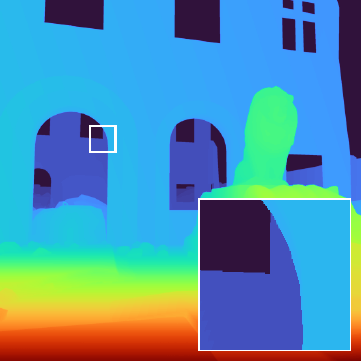}} &
  \raisebox{-0.5\height}{\includegraphics[width=0.12\textwidth]{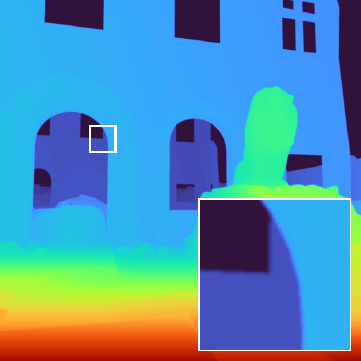}} &
  \raisebox{-0.5\height}{\includegraphics[width=0.12\textwidth]{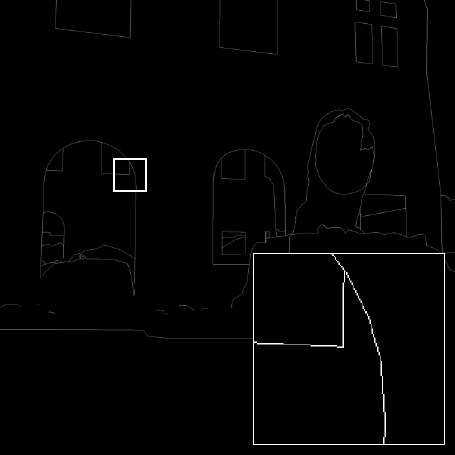}} &
  \raisebox{-0.5\height}{\includegraphics[width=0.12\textwidth]{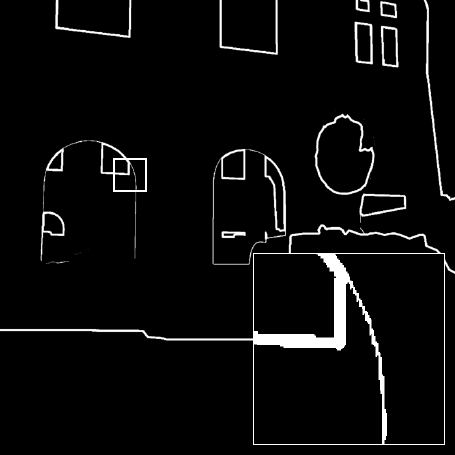}}\vspace{0.25em}\\
  \rotatebox[origin=c]{90}{\parbox{1.3cm}{Up-Down\\(bicubic)}} &
   \raisebox{-0.5\height}{\includegraphics[width=0.12\textwidth]{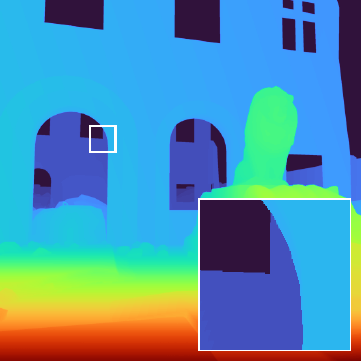}} &
  \raisebox{-0.5\height}{\includegraphics[width=0.12\textwidth]{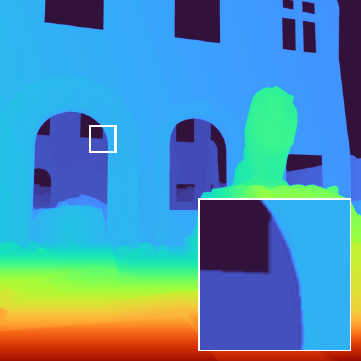}} &
  \raisebox{-0.5\height}{\includegraphics[width=0.12\textwidth]{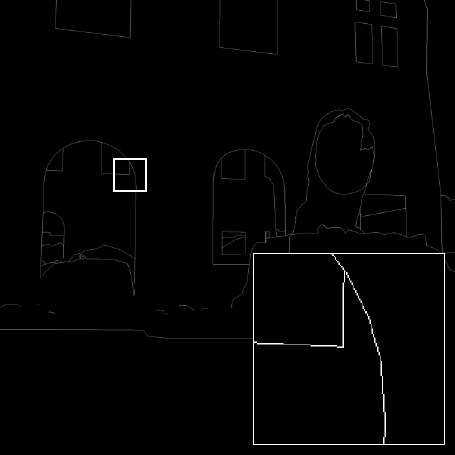}} &
  \raisebox{-0.5\height}{\includegraphics[width=0.12\textwidth]{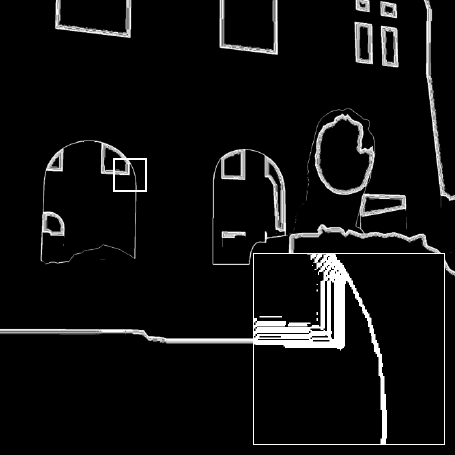}} \vspace{0.25em}\\
 \end{tabular}
\caption{\textbf{Boundary evaluation metrics empirical study.} 
We demonstrate how various types of image perturbations impact our proposed edge metrics. We report quantitative and qualitative results for multiple ground-truth perturbations, such as simple image shifts, downsampling followed by upsamplings, and Gaussian blurring. We report both ground-truth and perturbed occluding contours, used to derive our $F$1 scores. Our results empirically demonstrate the correlation between erroneous depth edge predictions and low precision and recall values.
}
\label{fig:supp_edge_metrics}
\end{figure}

\mypara{Boundary metrics empirical study.} To illustrate how our boundary metrics work, we report additional qualitative edge metric results in \Fig~{\ref{fig:edge_metrics}}. In particular, we show the occluding contours derived from the ground-truth and predicted depth, which illustrate how incorrect depth boundary predictions can impact the metric.
Furthermore, to illustrate the behavior of the boundary precision and recall measurements under various image perturbations we also provide an empirical study in \Fig~{\ref{fig:supp_edge_metrics}}. We report both quantitative and qualitative results on samples from the UnrealStereo4K dataset~\citep{Tosi2021:CVPR}. Our results empirically demonstrate the correlation between erroneous depth
edge predictions and low precision and recall values.

\mypara{Results on the iBims dataset~\citep{Koch2018:ECCVW}.} We supplement our boundary evaluation by results on the iBims dataset, which is commonly used for evaluating depth boundaries.
iBims consists of images of indoor scenes that have been laser-scanned. The images are at $640\times480$ resolution and have been supplemented with manually annotated occlusion boundary maps to facilitate evaluation.
The iBims benchmark uses \emph{Depth Directed Errors} (DDE), which evaluate overall metric depth accuracy, \emph{Depth Boundary Errors} (DBE), which are similar in spirit to our proposed boundary metric but require manual annotation, and \emph{Planar Errors}, which evaluate the accuracy of planes derived from the depth maps.

We find that Depth Pro is on par with the state of the art according to the DDE and PE metrics, and significantly outperforms all prior work according to the boundary metrics.
\begin{table}[thb]
  \caption{\textbf{Zero-shot metric depth evalution on the iBims dataset~\citep{Koch2018:ECCVW}.} We report the iBims-specific \textit{Depth Directed Errors} (DDE), \textit{Depth Boundary Errors} (DBE) and \textit{Planar Errors} (PE). For fairness, all reported results were reproduced in our environment. Please see \Sec~{\ref{sec:supp_additional_experiments}}
  }
  \label{tab:ibims_suppmat}
  \centering
  \scriptsize
  \begin{tabular}{@{}llllllll@{}}
    %\hline
    \multicolumn{1}{p{1.5cm}}{Method} &
    \multicolumn{3}{c}{\centering DDE (in \%)} &
    \multicolumn{2}{c}{\centering DBE (in px)} &
    \multicolumn{2}{c}{\centering PE (in m/$\deg$)} \\
    \cmidrule(lr){2-4} \cmidrule(lr){5-6} \cmidrule(lr){7-8}
     &
     $\varepsilon^{0}_{\text{DDE}}$$\uparrow$ &
     $\varepsilon^{-}_{\text{DDE}}$$\downarrow$ &
     $\varepsilon^{+}_{\text{DDE}}$$\downarrow$ &
     $\varepsilon^{\text{acc}}_{\text{DBE}}$$\downarrow$ &
     $\varepsilon^{\text{comp}}_{\text{DBE}}$$\downarrow$ &
     $\varepsilon^{\text{plan}}_{\text{PE}}$$\downarrow$ &
     $\varepsilon^{\text{orie}}_{\text{PE}}$$\downarrow$  \\
    \midrule
    DPT~\citep{Ranftl2021:ICCV} & 58.744 & 41.255 & \cellfirst{0.000} & 3.580 & 39.372 & 0.138 & 31.837\\
    Metric3D~\citep{Yin2023:ICCV} & \cellsecond{88.608} & \cellthird{1.337} & 10.054 & 2.073 & 19.011 & \cellthird{0.100} & 22.451 \\
    Metric3D v2 ~\citep{Hu2024:arxiv} & 84.721 & \cellsecond{0.546} & 14.732 & \cellthird{1.843} & \cellsecond{10.062} & \cellfirst{0.095} & 19.561 \\
    ZoeDepth~\citep{Bhat2023:arxiv} & 85.600 & 13.874 & \cellthird{0.525} & 1.960 & 18.166 & 0.103 & 20.108 \\
    Depth Anything~\citep{Yang2024:CVPR} & 88.951 & 10.741 & \cellsecond{0.308} & 2.081 & 19.172 & 0.106 & 20.680 \\
    Depth Anything v2~\citep{Yang2024:arxiv} & \cellfirst{91.773} & 1.619 & 6.607 & 1.959 & \cellfirst{8.350} & \cellfirst{0.095} & \cellthird{19.406} \\
    PatchFusion~\citep{Li2024:CVPR} & 85.765 & 12.602 & 1.633 & \cellsecond{1.711} & 20.722 & 0.117 & 23.926\\
    Marigold~\citep{Ke2024:CVPR} & 58.738 & 41.261 & \cellfirst{0.000} & 1.855 & 12.742 & 0.168 & 33.734 \\
    UniDepth ~\citep{Piccinelli2024:CVPR} & 73.020 & \cellfirst{0.041} & 26.939 & 1.999 & 14.234 & \cellsecond{0.098} & \cellsecond{19.114} \\
    \hline
    Depth Pro (Ours) & \cellsecond{89.725} & 1.809 & 8.464 & \cellfirst{1.680} & \cellthird{10.138} & \cellfirst{0.095} & \cellfirst{18.776}\\
  \bottomrule
  \end{tabular}
\end{table}

\clearpage

\section{Controlled Experiments}
\label{sec:controlled_experiments}
We conduct several controlled experiments to investigate the impact of various components and design decisions in Depth Pro. Specifically, we aim to assess the contribution of the native output resolution, key components in the network architecture (\Sec~{\ref{sec:supp_network_architecture}}), the depth representation, training objectives (\Sec~{\ref{sec:supp_training_objectives}}), training curriculum (\Sec~{\ref{sec:supp_full_curricula}}), and the focal length estimation head (\Sec~{\ref{sec:supp_focal_length}}).

\subsection{Native output resolution}
\label{sec:supp_output_resolution}
To assess the importance of the native output resolution, we conduct a simple experiment on UnrealStereo4K. We downsample the ground truth depth maps to several common resolutions found in the literature as well the output resolution of Depth Pro. We then upsample again to the original resolution and evaluate depth and boundary metrics. Results are listed in Tab.~\ref{tab:supp_output_resolution}. We find that the native output resolution has a strong effect on boundary accuracy.
More specifically, doubling the resolution may improve boundary accuracy by a factor of 3. It is important to note however, that these results represent an upper bound. 
The zero-shot comparison to the state of the art (Tab.~\ref{tab:sota_0shot_metric} and \ref{tab:sota_boundaries}) includes approaches that predict depth at the full input resolution, namely PatchFusion and ZeroDepth. Although PatchFusion and ZeroDepth predict at a higher resolution than \eg Metric3D v2 or DepthAnything v2, they trail the lower resolution approaches in performance. We conclude that predicting at a high native output resolution is necessary but not sufficient for predicting accurate metric depth with sharp boundaries.
\begin{table}[thbp]
    \centering
    \small
    \caption{\textbf{Native output resolution.}
    We evaluate the expected impact of the native output resolution on metric depth prediction and boundary accuracy. To that end, we bilinearly downsample the ground truth depth maps to several resolutions found in the literature, upsample again to the input resolution, and evaluate. The strong effect on boundary accuracy (3 fold increase per doubling of resolution) implies that predicting at a high native output resolution is a necessary but not necessarily sufficient condition for predicting accurate boundaries.
    }
    \label{tab:supp_output_resolution}
    \begin{tabular}{@{}clccc@{}}
    \multirow{2}{*}{Output resolution} & \multirow{2}{*}{Example} & \multicolumn{3}{c}{Approximate optimum}\\
    && Log10$\downarrow$ & AbsRel$\downarrow$ & F1$\uparrow$ \\
    \midrule
    1536$\times$1536 & Depth Pro & \cellfirst{0.019} & \cellfirst{0.004} & \cellfirst{0.311} \\
    768$\times$768 & Marigold~\citep{Ke2024:CVPR} & \cellsecond{0.048} & \cellsecond{0.010} & \cellsecond{0.131} \\
    518$\times$518 & Depth Anything v2~\citep{Yang2024:CVPR} & \cellthird{0.084} & \cellthird{0.016} & \cellthird{0.065} \\
    384$\times$384 & DPT~\citep{Ranftl2021:ICCV} & 0.123 & 0.024 & 0.044 \\
    \bottomrule
    \end{tabular}
\end{table}

\subsection{Network backbone}
\label{sec:supp_network_architecture}
To assess the effect of the network architecture, we begin by evaluating various candidate image encoder backbones within our network architecture. To assess their performance, we conduct a comparative analysis utilizing off-the-shelf models available from the TIMM library \citep{Wightman2019:TIMM}.
Using the pretrained weights, we train each backbone at 384 $\times$ 384
resolution across five RGB-D datasets (Keystone, HRWSI, RedWeb, TartanAir, and Hypersim) and  evaluate their performance in terms of metric depth accuracy across multiple datasets, including Booster, Hypersim, Middlebury, and NYUv2, utilizing metrics such as $\mathit{AbsRel}$ for affine-invariant depth and $\mathit{Log}_{10}$ for metric depth in \Tab~\ref{tab:backbones384}.
We find that ViT-L DINOv2 ~\citep{Oquab2024:TMLR} outperforms all other backbones by a significant margin and conclude that the combination of backbone and pretraining strategy considerably affects downstream performance.
Following this analysis, we pick ViT-L DINOv2 for our encoder backbones.
\begin{table}[thbp]
    \caption{\textbf{Comparison of image encoder backbones candidates.} We train each backbone at 384$\times$384 resolution across five RGB-D datasets: Keystone, HRWSI, RedWeb, TartanAir, and Hypersim. To ensure fair comparison, we select backbone candidates with comparable computational complexity, measured in Flops using the fvcore library~\citep{Fvcore}, and an equivalent number of parameters. We identify ViT-L DINOv2~\citep{Oquab2024:TMLR} as the optimal choice for our image encoder backbone, given its superior depth accuracy performance.
      }
      \label{tab:backbones384}
  \small
  \centering
  \begin{tabular}{@{}lllllll@{}}
    \toprule
    Backbone & Flops (G) & Params (M) & AbsRel $\downarrow$ & Log$_{10} \downarrow $\\
    \midrule
    ViT-L DINOv2-reg4~\citep{Oquab2024:TMLR}  & 248 & 345 & \cellfirst{0.039} & \cellthird{0.138} &\\
    ViT-L DINOv2~\citep{Oquab2024:TMLR}  & 247 & 345 & \cellsecond{0.040} & \cellfirst{0.129}\\
    ViT-L MAE~\citep{He2022:CVPR}  & 247 & 343 & \cellthird{0.041} & 0.150\\
    ViT-L BeiTv2~\citep{Peng2022:arxiv}  & 242 & 336 & 0.042 & \cellsecond{0.134}\\
    ViT-L BeiT~\citep{Bhat2023:arxiv} & 259 & 336 & 0.048 & 0.147\\
    ViT-L SO400m-siglip~\citep{Zhai2023:ICCV} & 311 & 471 & 0.051 & 0.174\\
    ViT-L CLIP-quickgelu~\citep{Radford2021:PMLR}  & 247 & 344 & 0.053 & 0.166\\
    ViT-L CLIP~\citep{Radford2021:PMLR}  & 247 & 345 & 0.057 & 0.156\\
    ViT-L~\citep{Dosovitskiy2021:ICLR}  & 247 & 345 & 0.061 & 0.163\\
    ConvNext-XXL~\citep{LiuZhuang2022:CVPR} & 514 & 867 & 0.075 & 0.216\\
    ViT-L DeiT-3~\citep{Touvron2022:ECCV}  & 247 & 345 & 0.078 & 0.176\\
    ConvNext-L-mlp~\citep{LiuZhuang2022:CVPR} & 162 & 214 & 0.081 & 0.222\\
    ConvNextv2-H~\citep{Woo2023:CVPR}  & 405 & 680 & 0.085 & 0.242\\
    SegAnything ViT-L~\citep{Kirillov2023:ICCV} & 245 & 330 & 0.087 & 0.311\\
    SWINv2-L~\citep{Liu2022:CVPR}  & 177 & 212 & 0.091 & 0.240\\
    CAFormer-B36~\citep{Yu2024:TPAMI}  & 124 & 108 & 0.091 & 0.248\\
    EfficientViT-L3~\citep{Liu2023:CVPR}  & -- & -- & 0.109 & 0.303\\
  \bottomrule
  \end{tabular}
\end{table}

\subsection{High-resolution alternatives}
\label{sec:high_resolution_alternatives}
We further evaluate alternative high-resolution 1536$\times$1536 network structures and different pretrained weights (\Tab~\ref{tab:backbone_alternatives}). To do this, we test generalization accuracy by training on a train split of some datasets and testing on a val or test split of other datasets, following the Stage 1 protocol for all models in accordance with \Tab~\ref{tab:spn_hp} and \Tab~\ref{tab:training_loss}. All ViT models use a patch size of 16$\times$16. For weights pretrained with a patch size of 14$\times$14 we apply bicubic interpolation to the weights of the convolutional patch embedding layer and scale these weights inversely to the number of pixels (i.e., the weights are reduced by a factor of 1.3). All ViT models use resolution 1536$\times$1536, for this we apply bicubic interpolation to positional embeddings prior to training. The Depth Pro approach in all cases uses ViT with resolution 384$\times$384 and patch size 16$\times$16 for both the patch encoder and the image encoder. SWINv2 and convolutional models are pretrained on ImageNet~\citep{Deng2009:CVPR}. Other models use different pretraining approaches described in their papers: CLIP~\citep{Radford2021:PMLR}, MAE~\citep{He2022:CVPR}, BeiTv2~\citep{Peng2022:arxiv}, and DINOv2~\citep{Oquab2024:TMLR}. For the Segment Anything model we use publicly available pretrained weights, which were initialized using MAE pretraining~\citep{He2022:CVPR} and subsequently trained for segmentation as described in their paper~\citep{Kirillov2023:ICCV}.
\begin{table}[thb]

    \caption{
      \textbf{High-resolution alternatives.}
      Generalization accuracy of alternative high-resolution 1536$\times$1536 models and different pretrained weights. All models are trained identically using Stage 1 in accordance with \Tab~\ref{tab:spn_hp} and \Tab~\ref{tab:training_loss}. Latency measured on a single GPU V100 with FP16 precision using batch=1. All ViT models use a patch size of 16$\times$16. Depth Pro employs a ViT-L DINOv2~\citep{Oquab2024:TMLR} for the image and patch encoders.
      }
  \label{tab:backbone_alternatives}
  \centering
  \scriptsize
  \begin{tabular}{@{}ll|c|cc|cc}
    & & & \multicolumn{2}{c}{Metric depth accuracy} & \multicolumn{2}{c}{Boundary accuracy} \\
    & Method & Latency, ms $\downarrow$ & NYUv2 $\delta_1\!\uparrow$ & iBims $\delta_1\!\uparrow$ & iBims F1$\uparrow$ & DIS R$\uparrow$ \\
    \midrule
    \multirow{2}{*}{\rotatebox[origin=r]{90}{Conv.}}
    & EfficientNetV2-XL~\citep{Tan2021:ICML} & 118 & 4.4 & 7.0 & 0.005 & 0.000 \\
    & ConvNext-XXL~\citep{LiuZhuang2022:CVPR} & 304 & 68.0 & 38.3 & 0.134 &  0.031 \\
    & ConvNextv2-H~\citep{Woo2023:CVPR} & 287 & 70.0 & 56.6 & 0.131 & 0.044 \\
    \midrule
    \multirow{2}{*}{\rotatebox[origin=r]{90}{Trans.}}
    & S. Anything~\citep{Kirillov2023:ICCV} (ViT-L) & 349 & 53.2 & 38.9 & 0.140 & 0.051  \\
    & S. Anything~\citep{Kirillov2023:ICCV} (ViT-H) & 365 & 51.7 & 41.1 & 0.146 & 0.050  \\
    & SWINv2-L~\citep{Liu2022:CVPR} (window=24) & 272 & 58.4 & 33.1 & 0.117 & 0.028 \\
    \midrule
    \multirow{4}{*}{\rotatebox[origin=r]{90}{ViT}}
    & ViT-L CLIP~\citep{Radford2021:PMLR} & 384 & 92.2 & 81.9 & 0.157 & \cellthird{0.052} \\
    & ViT-L BeiTv2~\citep{Peng2022:arxiv} & OOM & 90.4 & \cellthird{86.5} & 0.149 & 0.042 \\
    & ViT-L MAE~\citep{He2022:CVPR} & 390 & \cellthird{92.7} & 84.7 & \cellsecond{0.163} & \cellsecond{0.065} \\
    & ViT-L DINOv2~\citep{Oquab2024:TMLR} & 392 & \cellfirst{96.5} & \cellsecond{90.3} & \cellthird{0.161} & \cellsecond{0.065} \\

    \midrule
    & Depth Pro & 341 & \cellsecond{96.1} &  \cellfirst{91.3} &  \cellfirst{0.177} &  \cellfirst{0.080} \\
  \bottomrule
  \end{tabular}\\
\end{table}

We find that the presented Depth Pro approach is faster and more accurate for object boundaries than the plain ViT, with comparable metric depth accuracy.
In comparison to other transformer-based and convolutional models, Depth Pro has comparable latency, several times lower metric depth error, and several times higher recall accuracy for object boundaries.

Importantly, our proposed architecture performs significantly better than straight-forward scaling up the ViT architecture with DINOv2 pretraining. On DIS5K for instance, our architecture improves the boundary recall by relative $23\%$ over DINOv2.

\subsection{Depth representation}
\label{sec:depth_representation}
To assess the effect of the predicted depth representation, we train a ViT encoder and DPT decoder on Hypersim to predict inverse depth, log-depth, or depth. All configurations in this experiment are supervised with a only mean absolute error.
\Tab~\ref{tab:depth_vs_inverse_depth} lists the $delta_1$ error computed over several depth ranges. We find that supervising the network on depth directly leads to worse results than log-depth or inverse depth. Overall, predicting inverse depth works best, with the largest difference in the regions close to the camera. This makes inverse depth the representation of choice for downstream tasks like novel view synthesis, which benefit particularly from higher accuracy close to the camera.
\begin{table}[htbp]
    \caption{\textbf{Depth representation.} Optimizing for inverse-depth yields the most accurate predictions near the camera, which is particularly important for novel-view synthesis applications.}
  \label{tab:depth_vs_inverse_depth}
  \centering
  \begin{tabular}{@{}l|llllll}
    & \multicolumn{6}{c}{Hypersim $\delta_1\!\uparrow$} \\
    Training objective
     & 0-1m
     & 1-2m
     & 2-4m
     & 4-8m
     & 8-16m
     & $>$16m \\
    \midrule
    Inverse-depth & \cellfirst{0.730} & \cellfirst{0.833} & \cellfirst{0.896} & \cellfirst{0.921} & \cellfirst{0.922} & \cellfirst{0.922} \\
    Log-depth & \cellsecond{0.700} & \cellsecond{0.807} & \cellsecond{0.892} &  \cellsecond{0.919} & \cellsecond{0.920} & \cellsecond{0.920} \\
    Depth & \cellthird{0.657} & \cellthird{0.716} & \cellthird{0.819} & \cellthird{0.853} & \cellthird{0.850} & \cellthird{0.850} \\
  \bottomrule
  \end{tabular}
\end{table}

\subsection{Training objectives}
\label{sec:supp_training_objectives}
To assess the efficacy of our training curriculum, we compare it to alternative training schedules. We first examine the different stages individually and then compare full curricula.

\begin{table}[thbp]
    \caption{\textbf{Comparison of stage 1 training objectives.} 1A only applies the $\LMAE$ to metric, and the $\LSSIMAE$ to non-metric datasets. 1D additionally minimizes gradients on all datasets. 1B minimizes gradients only on synthetic datasets. We use 1C, which minimizes gradients with a scale-and-shift-invariant $\LSSIMAGE$ loss on all synthetic datasets irrespective of whether they are metric.}
  \label{tab:stage1_losses}
  \centering
  \small
  \begin{tabular}{@{}l|ll|llll|lll}
    & \multicolumn{2}{c}{HRWSI} & \multicolumn{4}{c}{Hypersim} & \multicolumn{3}{c}{Apolloscape} \\
    Cond. 
     & AbsRel$\downarrow$ 
     & $\delta_1\!\uparrow$
     & Log$_{10}\!\downarrow$ 
     & AbsRel$\downarrow$ 
     & $\delta_1\!\uparrow$
     & F1$\uparrow$
     & Log$_{10}\!\downarrow$ 
     & AbsRel$\downarrow$ 
     & $\delta_1\!\uparrow$
     \\
    \midrule
    1A & 0.166 & 82.1 & 0.083 & 0.259 & 75.4 & 0.221 & 0.156 & \cellthird{0.339} & 45.6 \\
    1D & \cellfirst{0.138} & \cellfirst{85.1} & \cellsecond{0.077} & \cellsecond{0.246} & \cellsecond{78.4} & \cellsecond{0.391} & \cellsecond{0.128} & 0.424 & \cellsecond{60.6} \\
    1B & \cellthird{0.156} & \cellthird{83.3} & \cellthird{0.078} & \cellthird{0.249} & \cellthird{77.3} & \cellthird{0.388} & \cellthird{0.152} & \cellsecond{0.300} & \cellthird{47.3} \\
    \midrule
    1C & \cellsecond{0.150} & \cellsecond{83.7} & \cellfirst{0.074} & \cellfirst{0.235} & \cellfirst{79.9} & \cellfirst{0.442} & \cellfirst{0.084} & \cellfirst{0.235} & \cellfirst{75.6} \\    
  \bottomrule
  \end{tabular}
\end{table}

\mypara{Stage 1 training objectives.}
We first evaluate loss combinations for the first stage and report results in \Tab~\ref{tab:stage1_losses}. 
Condition 1A only applies a mean absolute error loss to all datasets. For non-metric datasets, we use the scale-and-shift-invariant version. Condition 1B adds gradient losses to all synthetic datasets. We again use the scale-and-shift-invariant version for non-metric datasets.
Following our observations from \Sec~\ref{sec:method}, we propose to apply an appropriate mean absolute error loss as in other conditions depending on a dataset being metric, but apply a scale-and-shift-invariant gradient loss irrespective of a dataset being metric or not (C).
We find that loss combinations minimizing gradients (1B, 1C, 1D) consistently outperform just applying an absolute error (1A). Besides improving relative and metric depth estimates, they strongly improve boundary metrics, here up to a factor of 2.
Interestingly, minimizing gradients on all datasets outperforms minimizing gradients on just synthetic data. This suggests that the added diversity from real-world datasets more than balances out their potentially noisy ground truth, even for minimizing gradients, which emphasize the noise.
The best performance however, is achieved by applying a scale-and-shift-invariant gradient loss on the synthetic datasets (1C). We found that this setting improves convergence and overall performs best.

\begin{table}[thbp]
    \caption{\textbf{Comparison of stage 2 training objectives.} We evaluate the efficacy of derivative-based losses for sharpening boundaries. Employing first- and second-order derivative losses (2A) yields the best results on balance as indicated by the average rank over metrics. More details in the text.}
  \label{tab:stage2_losses}
  \scriptsize
  \centering
  \begin{tabular}{@{}l|llll|ll|llll|lll}
    & & & & & \multicolumn{2}{c}{HRWSI} & \multicolumn{4}{c}{Hypersim} & \multicolumn{3}{c}{Apolloscape} \\
    \rot{Condition}
     & \rot{$\LMSE$}
     & \rot{$\LMAGE$}
     & \rot{$\LMSGE$}
     & \rot{$\LMALE$}
     & AbsRel$\downarrow$ 
     & $\delta_1\!\uparrow$
     & Log$_{10}\!\downarrow$ 
     & AbsRel$\downarrow$ 
     & $\delta_1\!\uparrow$
     & F1$\uparrow$
     & Log$_{10}\!\downarrow$ 
     & AbsRel$\downarrow$ 
     & $\delta_1\!\uparrow$
     \\
    \midrule
    2A & \checkmark & \checkmark & \checkmark & \checkmark & \cellsecond{0.149} & \cellsecond{83.6} & \cellfirst{0.072} & \cellsecond{0.235} & \cellfirst{81.3} & \cellsecond{0.465} & \cellsecond{0.092} & \cellthird{0.303} & 72.9  \\
    \midrule
    2B & \checkmark & \checkmark & \checkmark & & \cellfirst{0.148} & \cellfirst{83.7} & \cellfirst{0.072} & \cellfirst{0.230} & \cellsecond{81.0} & \cellthird{0.463} & \cellsecond{0.092} & \cellfirst{0.299} & \cellthird{73.1} \\
    2C & \checkmark & \checkmark & & & \cellthird{0.150} & \cellfirst{83.7} & \cellfirst{0.072} & \cellsecond{0.235} & \cellthird{80.8} & \cellfirst{0.468} &\cellfirst{0.091} & \cellsecond{0.300} & \cellsecond{73.2} \\
    2D & \checkmark & & & & \cellthird{0.150} & \cellthird{83.4} & \cellsecond{0.074} & \cellthird{0.239} & 79.8 &  0.461 & \cellthird{0.096} & 0.349 & 72.8 \\
    2E & & & & & 0.159 & 82.7 & \cellsecond{0.074} & 0.242 & 80.6 & 0.459 & \cellthird{0.096} & 0.346 & \cellfirst{73.3} \\
  \bottomrule
  \end{tabular}
\end{table}

\mypara{Stage 2 training objectives.}
The second stage of our training curriculum focuses on sharpening depth boundaries while retaining high metric depth accuracy.
To that end, we only employ synthetic datasets due to their high quality ground truth.
The obvious strategy for sharpening predictions is the application of gradient losses. We evaluate our combination of multi-scale derivative-based losses in an ablation study. Condition 2A uses all of the losses, namely $\LMAE$, $\LMSE$ ,$\LMAGE$, $\LMALE$, and $\LMSGE$. See \Tab~\ref{tab:stage2_losses}.
2B removes the second-order loss $\LMALE$. 2C further removes the squared first order losses $\LMSGE$. 2D removes all derivative-based losses.
2E applies the $\LMAE$ to all datasets.
Removing $\LMALE$ improves results on Apolloscape.
Our combination of 0th- to 2nd-order derivative losses (2A) performs best across metrics and datasets in aggregate (e.g., in terms of the average rank across metrics).

\subsection{Full curricula}
\label{sec:supp_full_curricula}
\begin{table}[thbp]
  \caption{\textbf{Comparison of full curricula.} We evaluate our curriculum (3A) against single stage training (3B), and pretraining on synthetic and fine-tuning on real data (3C).}
  \label{tab:curricula}
  \small
  \centering
  \begin{tabular}{@{}l|ll|llll|lll}
    & \multicolumn{2}{c}{HRWSI} & \multicolumn{4}{c}{Hypersim} & \multicolumn{3}{c}{Apolloscape} \\
    Cond. 
     & AbsRel$\downarrow$ 
     & $\delta_1\!\uparrow$
     & Log$_{10}\!\downarrow$ 
     & AbsRel$\downarrow$ 
     & $\delta_1\!\uparrow$
     & F1$\uparrow$
     & Log$_{10}\!\downarrow$ 
     & AbsRel$\downarrow$ 
     & $\delta_1\!\uparrow$
     \\
    \midrule
    3A (Ours) & \cellsecond{0.149} & \cellsecond{83.6} & \cellfirst{0.072} & \cellfirst{0.235} & \cellfirst{81.3} & \cellsecond{0.465} & \cellfirst{0.092} & \cellsecond{0.303} & \cellfirst{72.9} \\
    \midrule
    3B & \cellfirst{0.148} & \cellfirst{83.9} & \cellsecond{0.073} & \cellsecond{0.245} &  \cellfirst{81.3} & \cellfirst{0.478} & \cellsecond{0.095} & \cellfirst{0.292} & \cellsecond{72.1}  \\ 
    3C & \cellthird{0.153} & \cellsecond{83.6} & \cellthird{0.166} & \cellthird{0.386} & \cellsecond{37.1} & \cellthird{0.095} & \cellthird{0.586} & \cellthird{0.712} & \cellthird{0.5} \\ 
  \bottomrule
  \end{tabular}
\end{table}

We assess the efficacy of our complete training curriculum in comparison to alternatives. Condition 3A represents our two-stage curriculum. Condition 3B trains in a single stage and applies all the second-stage gradient losses throughout the whole training. Condition 3C reverses our two stages and represents the established strategy of pretraining on synthetic data first and fine-tuning with real-world data.
We find that training with a single stage is a reasonable default strategy and works much better than first training on synthetic data and then fine-tuning on real data. Our proposed strategy however, yields further improvements on metric depth with slightly worse boundary accuracy.

\subsection{Focal length estimation}
\label{sec:supp_focal_length}

\begin{figure}[htbp]
    \centering
    \includegraphics[width=\textwidth]{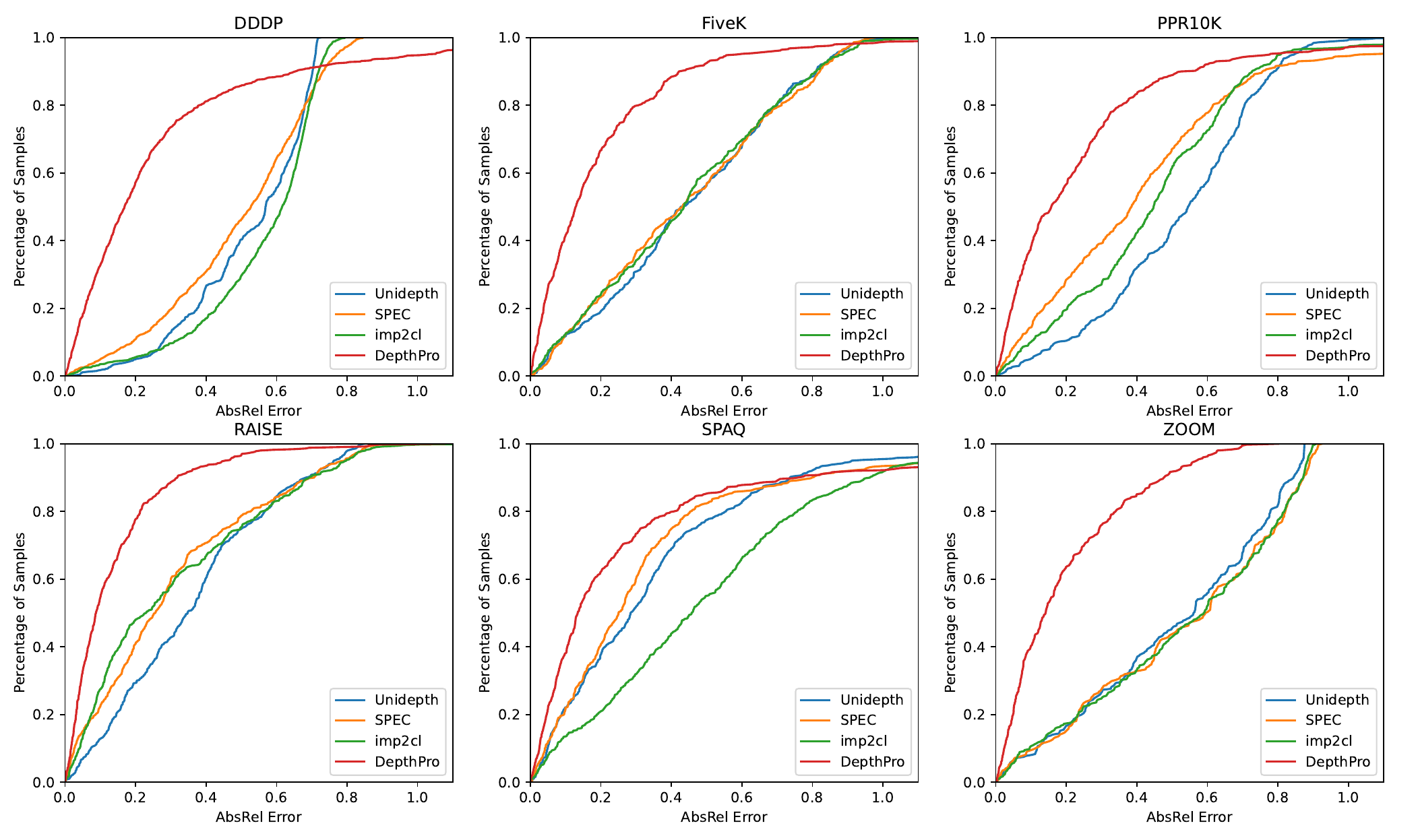}
    \caption{\textbf{Evaluation of focal length estimation.} Each plot compares a number of methods on a given dataset. The $x$ axis represents the AbsRel error and the $y$ axis represents the percentage of samples whose error is below that magnitude.}
    \label{fig:supp:focal}
\end{figure}

\mypara{Additional analysis of zero-shot focal length estimation accuracy.}
In \Fig~\ref{fig:supp:focal}, we present a more comprehensive analysis of our focal length predictor's performance compared to baseline models.
To that end, we plot the percentage of samples below a certain absolute relative error for each method and dataset in our zero-shot evaluation set up.
Depth Pro outperforms all approaches on all datasets.

\begin{table}[thbp]
\centering
\caption{Controlled experiment on focal length estimation. We evaluate several variants of a focal length branch and find that the combination of a separate image encoder trained for focal length estimation and a frozen image encoder for depth estimation perform best.}
\label{tab:abl:fov}
\begin{tabular}{lc}
\toprule
 Architecture     & $\delta_{25\%}\uparrow$ \\
 \midrule
Encoder for depth only    & {60.0} \\       
Encoder for focal length only    & \cellsecond{74.4}           \\
Encoder for depth and refinement network & \cellthird{63.6}   \\
Parallel encoders for depth and focal length & \cellfirst{78.2}   \\
\bottomrule
\end{tabular}
\end{table}

\mypara{Controlled evaluation of network structures.}
We evaluate a number of choices for the focal length estimation head and report results in \Tab~\ref{tab:abl:fov}. The models are evaluated on 500 images randomly sampled from Flickr~\citep{Thomee2016:CACM}.
As the first condition, we extract features from the frozen image encoder trained for depth estimation and add a small convolutional head. As the second condition, we train a separate ViT-based image encoder~\citep{Dosovitskiy2021:ICLR}. 
As the third condition, we train a ViT-based encoder on extracted features from the frozen image encoder for depth estimation. 
The final condition represents our chosen architecture depicted in \Fig~\ref{fig:architecture}, which utilizes frozen features from the depth network and task-specific features from a separate ViT image encoder in parallel.

We observe that refining depth features performs on par with just using the frozen depth features, suggesting that adding more computation on top of the frozen DPT features in addition to our small convolutional head does not provide extra benefits despite the increased computation. Training a separate image encoder from scratch improves performance by 14.6 percentage points, which indicates that accurate focal length prediction requires extra task-specific knowledge in addition to depth information.  Furthermore, the two encoders in parallel outperform using just a single image encoder for focal length prediction, which highlights the importance of features from the pretrained depth network for obtaining a high-performing focal length estimator.

\section{Implementation, Training and Evaluation Details}
\label{sec:details}
In this section we provide additional details on the datasets used for training and evaluation, hyperparameter settings for our method, and details on the evaluation setup.

\subsection{Merge operation}
\label{sec:merge}
\begin{figure}[thb]
    \centering
    \includegraphics[trim=2cm 28cm 47.4cm 0cm, clip, width=0.5\linewidth]{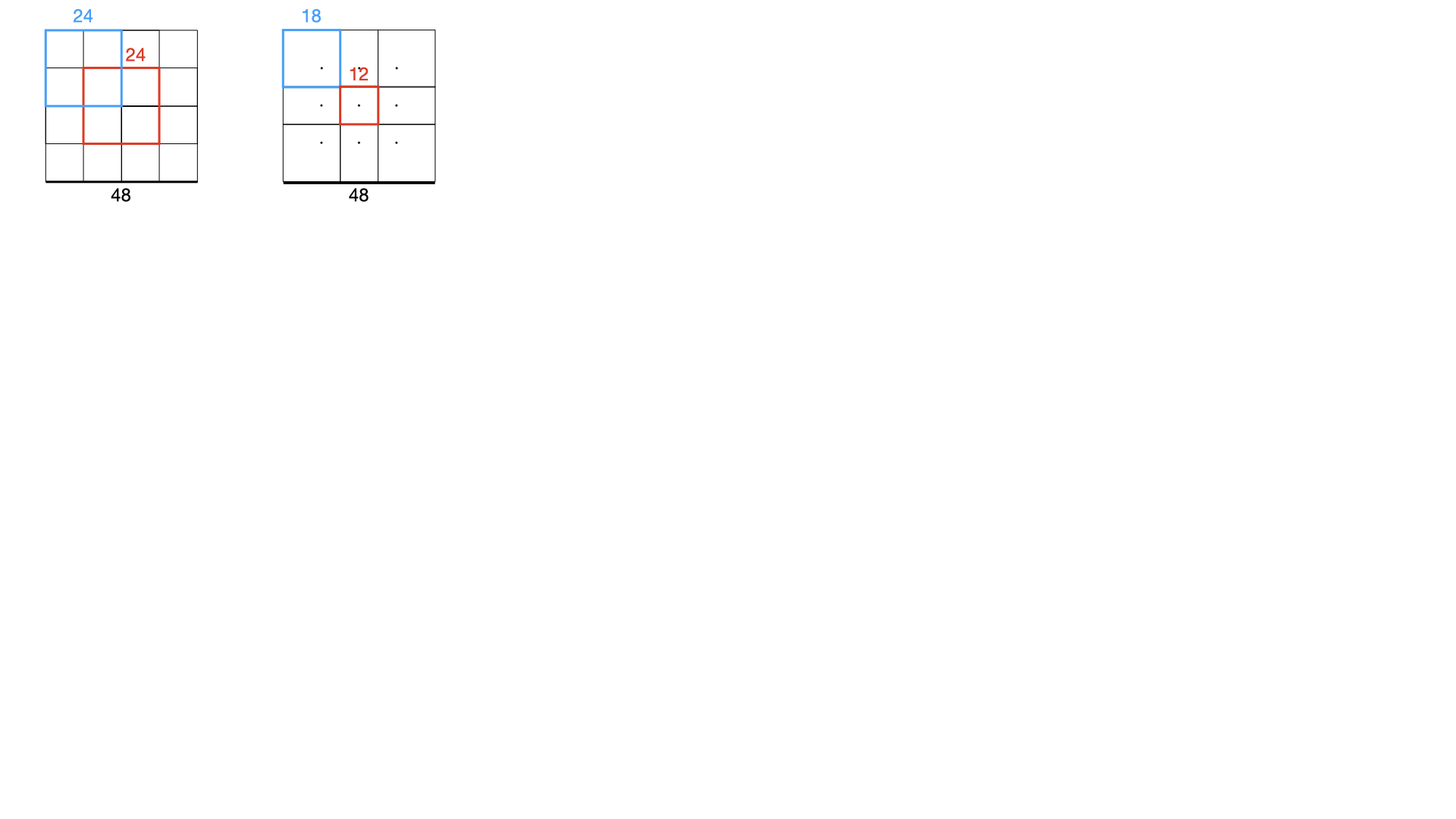}
    \caption{\textbf{Merge operation.} To merge overlapping feature patches (left), we generate a Voronoi partition of the feature map using the patch centers as generators (right). Red and blue rectangles exemplify the original feature patch on the left and the fraction to be retained on the right. The patch centers are indicated by dots on the right, numbers denote side lengths of the feature patches and the feature map.}
    \label{fig:merge}
\end{figure}

We merge overlapping feature patches to feature maps by generating a Voronoi partition of the desired feature map. To generate the partition, we use the patch centers as seeds and obtain a Voronoi cell per feature patch. The area of the patch covered by the Voronoi cell is copied to the feature map, the remaining area discarded. By overlapping patches we ensure that the receptive field of the patch encoder partially covers neighboring patches. \Fig~\ref{fig:merge} illustrates the approach for merging $3\times3$ patches into the \emph{Feature 4} feature map.

\subsection{Focal length head}
\label{sec:focal_head}
The focal length head consists of a three-layer convolutional network with kernel sizes 3,3,6, and strides 2,2,1. Each layer halves the channel dimension and is followed by a rectified linear unit. The first layer starts from 128 channels and the last reduces channels to a single focal length value per image.

\subsection{Datasets}
\Tab~\ref{tab:datasets} provides a comprehensive summary of the datasets utilized in our study, detailing their respective licenses and specifying their roles (e.g., training or testing).
\begin{table}[htbp]
  \caption{Datasets used in this work.}
  \label{tab:datasets}  
  \centering
  \scriptsize
  \begin{tabular}{@{}llll@{}}
    \toprule
    Dataset & URL & License & Usage \\
    \midrule
    3D Ken Burns~\citep{Niklaus2019:SIGGRAPH}
    &\tiny{\url{https://github.com/sniklaus/3d-ken-burns}}
    & CC-BY-NC-SA 4.0 & Train \\
    AM-2K~\citep{Li2022:IJCV}
    & \tiny{\url{https://github.com/JizhiziLi/GFM}}
    & \href{https://jizhizili.github.io/files/gfm_datasets_agreements/AM-2k_Dataset_Release_Agreement.pdf}{Custom} & Testing \\
    Apolloscape~\citep{Huang2020:TPAMI} & \tiny{\url{https://apolloscape.auto/}} & \href{https://apolloscape.auto/license.html}{Custom} & Val \\
    ARKitScenes~\citep{Dehghan2021}
    & \tiny{\url{https://github.com/apple/ARKitScenes}}
    & \href{https://github.com/apple/ARKitScenes?tab=License-1-ov-file#readme}{Custom} & Train \\
    Bedlam~\citep{Black2023:CVPR} 
    & \tiny{\url{https://bedlam.is.tue.mpg.de/#data}} 
    & \href{https://bedlam.is.tuebingen.mpg.de/license.html}{Custom} 
    & Train \\
    BlendedMVG~\citep{Yao2020:CVPR}
    & \tiny{\url{https://github.com/YoYo000/BlendedMVS}} 
    & CC BY 4.0 & Train \\
    Booster~\citep{Ramirez2024:PAMI} 
    & \tiny{\url{https://cvlab-unibo.github.io/booster-web/}} 
    & CC BY NC 4.0 & Test \\
    DDAD~\citep{Guizilini2020:CVPR} 
    & \tiny{\url{https://github.com/TRI-ML/DDAD}} 
    & CC-BY-NC-SA 4.0 & Testing \\   
    DIML (indoor)~\citep{Kim2016:TIP} 
    & \tiny{\url{https://dimlrgbd.github.io/}} 
    & \href{https://dimlrgbd.github.io/}{Custom} & Train \\
    DIS5K~\citep{Qin2022:ECCV} 
    & \tiny{\url{https://xuebinqin.github.io/dis/index.html}} 
    & \href{https://github.com/xuebinqin/DIS/blob/main/DIS5K-Dataset-Terms-of-Use.pdf}{Custom} & Test \\
    DPDD~\citep{Abuolaim2020:ECCV} 
    & \tiny{\href{https://github.com/Abdullah-Abuolaim/defocus-deblurring-dual-pixel}{\nolinkurl{https://github.com/Abdullah-Ab...pixel}}} 
    &  MIT & Testing \\
    Dynamic Replica~\citep{Karaev2023:ICCV} & \tiny{\href{https://github.com/facebookresearch/dynamic_stereo}{\nolinkurl{https://github.com/facebookres...stereo}}} & CC BY-NC 4.0 & Train \\
    EDEN~\citep{Le2021:WACV} & \tiny{\url{https://lhoangan.github.io/eden/}} & \href{https://lhoangan.github.io/eden/}{Custom} & Train \\
    ETH3D~\citep{Schps2017:CVPR} & \tiny{\url{https://www.eth3d.net/}} & CC-BY-NC-SA 4.0 & Testing \\
    FiveK~\citep{Bychkovsky2011:CVPR} & \tiny{\url{https://data.csail.mit.edu/graphics/fivek/}} &     \href{https://data.csail.mit.edu/graphics/fivek/legal/LicenseAdobe.txt}{Custom}
     & Testing \\
    HRWSI~\citep{Xian2020:CVPR} & \tiny{\href{https://kexianhust.github.io/Structure-Guided-Ranking-Loss/}{\nolinkurl{https://kexianhust.github....Ranking-Loss/}}} & \href{https://github.com/KexianHust/Structure-Guided-Ranking-Loss}{Custom} & Train, Val \\
    Hypersim~\citep{Roberts2021:ICCV} & \tiny{\url{https://github.com/apple/ml-hypersim}} & \href{https://github.com/apple/ml-hypersim?tab=License-1-ov-file#readme}{Custom} & Train, Val \\
    iBims~\citep{Koch2018:ECCVW} & \tiny{\url{https://www.asg.ed.tum.de/lmf/ibims1/}} & \href{https://www.asg.ed.tum.de/lmf/ibims1/}{Custom} & Test \\
    IRS~\citep{Wang2019:arxiv} & \tiny{\url{https://github.com/HKBU-HPML/IRS}} & \href{https://github.com/HKBU-HPML/IRS}{Custom} & Train \\
    KITTI~\citep{Geiger2013:IJRR} & \tiny{\url{https://www.cvlibs.net/datasets/kitti/}} & CC-BY-NC-SA 3.0 & Testing \\
    Middlebury~\citep{Scharstein2014:GCPR} & \tiny{\url{https://vision.middlebury.edu/stereo/data/}} & \href{https://vision.middlebury.edu/stereo/data/}{Custom} & Testing \\
    MVS-Synth~\citep{Huang2018:CVPR} & \tiny{\href{https://phuang17.github.io/DeepMVS/mvs-synth.html}{\nolinkurl{https://phuang17....mvs-synth.html}}} & \href{https://phuang17.github.io/DeepMVS/mvs-synth.html}{Custom} & Train \\
    NYUv2~\citep{Silberman:ECCV12} & \tiny{\href{https://cs.nyu.edu/~fergus/datasets/nyu_depth_v2.html}{\nolinkurl{https://cs.nyu.edu/...v2.html}}} & \href{https://cs.nyu.edu/~fergus/datasets/nyu_depth_v2.html}{Custom} & Testing \\
    nuScenes~\citep{Caesar2020:CVPR} & \tiny{\url{https://www.nuscenes.org/}} & \href{https://www.nuscenes.org/nuscenes}{Custom} & Testing \\
    P3M-10k~\citep{Li2021:ACMMM} & \tiny{\url{https://github.com/JizhiziLi/P3M}} & \href{https://jizhizili.github.io/files/p3m_dataset_agreement/P3M-10k_Dataset_Release_Agreement.pdf}{Custom} & Testing \\
    PPR10K~\citep{jie2021:cvpr} & \tiny{\url{https://github.com/csjliang/PPR10K}} &  Apache 2.0 & Testing \\
    RAISE~\citep{Dang2015:MMSys}
    & \tiny{\href{http://loki.disi.unitn.it/RAISE/download.html}{\nolinkurl{http://loki...download.html}}}
    & \href{http://loki.disi.unitn.it/RAISE/download.html}{Custom} 
    & Testing \\
    ReDWeb~\citep{Xian2018:CVPR} & \tiny{\url{https://sites.google.com/site/redwebcvpr18/}} & \href{https://sites.google.com/site/redwebcvpr18/}{Custom} & Train \\
    SAILVOS3D~\citep{Hu2021:CVPR} & \tiny{\href{https://sailvos.web.illinois.edu/_site/_site/index.html}{\nolinkurl{https://sailvos.web.illin...index.html}}} & \href{https://sailvos.web.illinois.edu/_site/_site/index.html}{Custom} & Train \\
    ScanNet~\citep{Dai2017:CVPR} & \tiny{\url{http://www.scan-net.org/}} & \href{https://kaldir.vc.in.tum.de/scannet/ScanNet_TOS.pdf}{Custom} & Train \\
    Sintel~\citep{Butler2012:ECCV} & \tiny{\url{http://sintel.is.tue.mpg.de/}} & \href{http://sintel.is.tue.mpg.de/}{Custom} & Testing \\
    SmartPortraits~\citep{Kornilova2022:CVPR}
    & \tiny{\href{https://mobileroboticsskoltech.github.io/SmartPortraits/}{\nolinkurl{https://mobile...SmartPortraits/}}} 
    & \href{https://mobileroboticsskoltech.github.io/SmartPortraits/}{Custom} & Train \\
    SPAQ~\citep{Fang2020:CVPR} & \tiny{\url{https://github.com/h4nwei/SPAQ}} &  \href{https://drive.google.com/drive/folders/1wZ6HOHi5h43oxTe2yLYkFxwHPgJ9MwvT}{Custom} & Testing \\
    Spring~\citep{Mehl2023:CVPR} & \tiny{\url{https://spring-benchmark.org/}} & CC BY 4.0 & Testing \\
    Sun-RGBD~\citep{Song2015:CVPR} & \tiny{\url{https://rgbd.cs.princeton.edu/}} & \href{https://rgbd.cs.princeton.edu/}{Custom} & Testing \\
    Synscapes~\citep{Wrennige2018:arxiv} & \tiny{\url{https://synscapes.on.liu.se/}} & \href{https://synscapes.on.liu.se/}{Custom} & Train \\
    TartanAir~\citep{Wang2020:IROS} & \tiny{\url{https://theairlab.org/tartanair-dataset/}} & CC BY 4.0 & Train \\
    UASOL~\citep{Bauer2019SD} & \tiny{\url{https://osf.io/64532/}} & CC BY 4.0 & Train \\
    UnrealStereo4K~\citep{Tosi2021:CVPR} & \tiny{\url{https://github.com/fabiotosi92/SMD-Nets}} & \href{https://github.com/fabiotosi92/SMD-Nets}{Custom} & Train \\
    Unsplash & \tiny{\url{https://unsplash.com/data}} & \href{https://unsplash.com/data}{Custom} & Testing \\
    UrbanSyn~\citep{Gomez2023:arxiv} & \tiny{\url{https://www.urbansyn.org/}} & CC BY-SA 4.0 & Train \\
    VirtualKITTI2~\citep{Gaidon2016:CVPR} & \tiny{\href{https://europe.naverlabs.com/research-old2/computer-vision/proxy-virtual-worlds/}{\nolinkurl{https://europe.naverlabs.com...-worlds/}}} & CC BY-NC-SA 3.0 & Train \\
    ZOOM~\citep{Zhang2019:CVPR} & \tiny{\href{https://github.com/ceciliavision/zoom-learn-zoom?tab=readme-ov-file#quick-inference}{\nolinkurl{https://github.com/ceciliav...inference}}} & - & Testing \\
    \bottomrule
  \end{tabular}
\end{table}

\subsection{Training hyperparameters}
We specify the training hyperparameters in \Tab~\ref{tab:spn_hp} and \Tab~\ref{tab:training_loss}.
\begin{table}[tb]
  \caption{Training hyperparameters.}
  \label{tab:spn_hp}
  \small
  \centering
  \begin{tabular}{@{}l|ll@{}}
     & \begin{tabular}{@{}p{30mm}|p{30mm}@{}}Stage 1  & Stage 2\end{tabular}
     \\
    \midrule
    Epochs & \begin{tabular}{@{}p{30mm}|p{30mm}@{}}250 & 100\end{tabular} \\
    \midrule
    Epoch length & 72000 \\
    Schedule & \SI{1}{\percent} warmup, \SI{80}{\percent} constant LR, \SI{19}{\percent} $\times$0.1 LR \\
    LR for Encoder & 1.28e-5 \\
    LR for Decoder & 1.28e-4 \\
    Batch size & 128 \\
    Optimizer & Adam\\
    Weight decay & 0 \\
    Clip gradient norm & 0.2 \\
    Pretrained LayerNorm & Frozen \\
    \midrule
    Random color change probability & \SI{75}{\percent} \\
    Random blur probability & \SI{30}{\percent} \\
    Center crop probability for FOV-augmentation & \SI{50}{\percent} \\
    Metric depth normalization & CSTM-label~\citep{Yin2023:ICCV} \\
    \midrule
    Number of channels for Decoder & 256 \\
    Resolution & 1536$\times$1536 \\
    \midrule
    DepthPro model structure: \\
    Image-Encoder resolution & 384$\times$384 \\
    Patch-Encoder resolution & 384$\times$384 \\
    Number of 384$\times$384 patches in DepthPro & 35 \\
    Intersection of 384$\times$384 patches in DepthPro & \SI{25}{\percent} \\
  \bottomrule
  \end{tabular}
\end{table}

\begin{table}[tb]
  \caption{Training loss functions for different datasets and stages.}
  \label{tab:training_loss}
  \small
  \centering
  \begin{tabular}{@{}p{60mm}|p{60mm}@{}}
    Loss function & Datasets \\
    \specialrule{.2em}{.1em}{.1em} 
    % \toprule
    \bf{Stage 1} & \\
    \\
    MAE \newline SSI-MAGE & Hypersim, Tartanair, Synscapes, Urbansyn, Dynamic Replica, Bedlam, IRS, Virtual Kitti2, Sailvos3d \\
    \\
    \specialrule{.01em}{.2em}{.2em} 
    MAE (trimmed = \SI{20}{\percent}) & ARKitScenes, Diml Indoor, Scannet, Smart Portraits \\
    \\
    \specialrule{.01em}{.2em}{.2em} 
    SSI-MAE \newline SSI-MAGE & UnrealStereo4k, 3D Ken Burns, Eden, MVS Synth \\
    \\
    \specialrule{.01em}{.2em}{.2em} 
    SSI-MAE (trimmed = \SI{20}{\percent}) & HRWSI, BlendedMVG \\
    \\
    \specialrule{.2em}{.1em}{.1em} 
    %\toprule
    \bf{Stage 2} & \\
    \\
    MAE, MSE, MAGE, MALE, MSGE & Hypersim, Tartanair, Synscapes, Urbansyn, Dynamic Replica, Bedlam, IRS, Virtual Kitti2, Sailvos3d \\
  \specialrule{.2em}{.1em}{.1em}  
  %\bottomrule
  \end{tabular}
\end{table}
%
%% Dataset evaluation code setup
\begin{table}[htb]
\caption{
    \textbf{Dataset evaluation setup.}For each metric depth dataset in our evaluation, we report the range of valid depth values, number of samples, and resolution of ground truth depth maps. Due to the large size of the validation set (approximately 35K samples), we used a randomly sampled subset of NuScenes.
}
\label{tab:sota_dataset_setup}
\centering
\scriptsize
\newcolumntype{Y}{S[table-format=2.3,table-auto-round]}
\newcolumntype{Z}{S[table-format=2.0,table-auto-round]}
\begin{tabularx}{\linewidth}{XYZXX}
\toprule
Dataset & {Minimum distance (m)} & {Maximum distance (m)} & Number of Samples & Depth Resolution (px)\\
\midrule
Booster & 0.001 & 10 & 228 & $3008 \times 4112$ \\ 
ETH3D & 0.1 & 200 & 454 & $4032 \times 6048$\\ 
iBims & 0.1 & 10 & 100 & $480 \times 640$ \\ 
Middlebury & 0.001 & 10 & 15 & $1988 \times 2952$\\ 
NuScenes & 0.001 & 80 & 881 & $900 \times 1600$\\ 
Sintel & 0.01 & 80 & 1064 & $436 \times 1024$ \\ 
Sun-RGBD & 0.001 & 10 & 5050 & $530 \times 730$ \\ 
\bottomrule
\end{tabularx}
\end{table}

\subsection{Baselines}
Below we provide further details on the setup of the baselines.

\mypara{DepthAnything.} Depth Anything v1 and v2 each released a general model for \emph{relative} depth, but their \emph{metric} depth models are tailored to specific domains (indoor vs.\ outdoor). For the metric depth evaluation, we match these models to datasets according to their domain, and for datasets containing both indoor and outdoor images, we select the model with the best performance. For qualitative results and the (scale and shift invariant) zero-shot boundary evaluation, we employ the relative depth models, since they yield better qualitative results and sharper boundaries than the metric models.

\mypara{Metric3D.}
For Metric3D v1 and v2, we found that the crop size parameter strongly affects metric scale accuracy. In fact, using a fixed crop size consistently yielded very poor results on at least one metric dataset. In order to obtain acceptable results, we used different crop sizes for indoor (512, 1088) and outdoor (512, 992) datasets. As in the case of Depth Anything, we mark these results in gray to indicate that they are not strictly zero-shot. For Metric 3D v2, we use the largest (`giant') model.

\mypara{UniDepth.} For UniDepth, we use the \textit{ViT-L} version, which performs best on average among the UniDepth variants.

\mypara{ZoeDepth.} We use the model finetuned on both indoor and outdoor data (denoted \textit{ZoeD\_NK}).

\subsection{Evaluation setup}
In evaluating our approach and baselines, we found the range of valid depth values, the depth map resolution used for computing metrics, the resizing approach used for matching the resolution of the ground truth depth maps, and the choice of intrinsics to affect results, sometimes strongly. This is why we made an effort to set up and evaluate each baseline in the same fair evaluation setup, which we detail below.

\Tab~\ref{tab:sota_dataset_setup} lists our evaluation datasets, the range of depth values used for evaluation, the number of samples, and the resolution of the ground truth depth maps. In case a method predicted depth maps at a different resolution, we resized predictions bilinearly to match the ground truth resolution.

Since several factors outlined above can affect the reported accuracy of a method, few baselines report sufficient detail on their evaluation setup, and the exact evaluation setups may differ across baselines, it is generally impossible to exactly reproduce reported results while guaranteeing fairness. We prioritized fair comparison and tried to evaluate all baselines in the same environment. We were able to match most reported results, with the following three notable differences. ZeroDepth reported better results on nuScenes, which we attribute to the use of a different validation set in their evaluation.
UniDepth reported different results on ETH3D, which we attribute to the handling of raw images; specifically, in our setup, we use the raw images without any post-processing, and take the intrinsics from the accompanying EXIF data; we believe this best adheres to the zero-shot premise for single-image depth estimation. Finally, on SUN-RGBD, Depth Anything fairs better in our evaluation setup than in the evaluation reported in the original paper.

\mypara{Evaluation metric for sharp boundaries.} For both our depth-based and mask-based boundary metrics, we apply the same weighted-averaging strategy to account for multiple relative depth ratios. F1 values (depth-based metrics) and recall values (mask-based metrics) are averaged across thresholds that range linearly from $t_{min}=5$ to $t_{max}=25$. Weights are computed as the normalized range of threshold values between $t_{min}$ and $t_{max}$, such that stronger weights are given towards high threshold values.

\section{Applications}
\label{sec:supp_applications}
Metric, sharp, and fast monocular depth estimation enables a variety of downstream applications. We showcase the utility of Depth Pro  in two additional contexts beyond novel view synthesis: conditional image synthesis with ControlNet \citep{Zhang2023:ICCVb} and synthetic depth of field \citep{Peng2022:CVPR}.

\begin{figure}[ht!]
\centering
    \begin{tabular}{@{}c@{\hspace{0.2mm}}c@{\hspace{2mm}}c@{}}
    \raisebox{2cm}[0pt][0pt]{\rotatebox[origin=c]{90}{\footnotesize Input}}&
    \includegraphics[height=3.8cm]{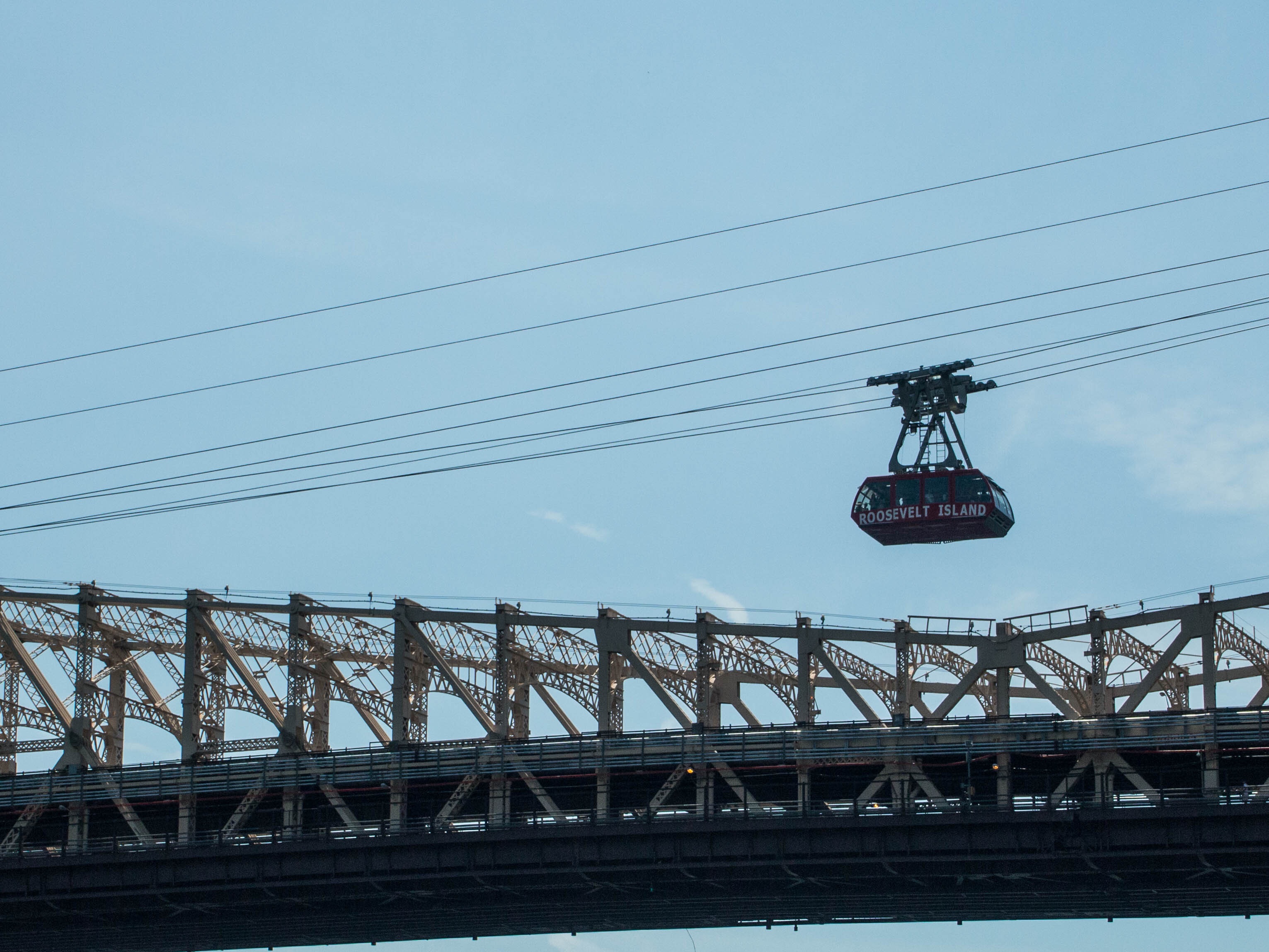} & 
    \raisebox{2.1cm}[0pt][0pt]{\parbox[t]{0.25\textwidth}{\textbf{Prompt:} A watercolor image of a cable-car.}}\\[5pt]
    & \footnotesize{Predicted depth} & \footnotesize{Synthesized image}\\
    \raisebox{2cm}[0pt][0pt]{\rotatebox[origin=c]{90}{\footnotesize Depth Pro}}&
    \includegraphics[height=3.8cm]{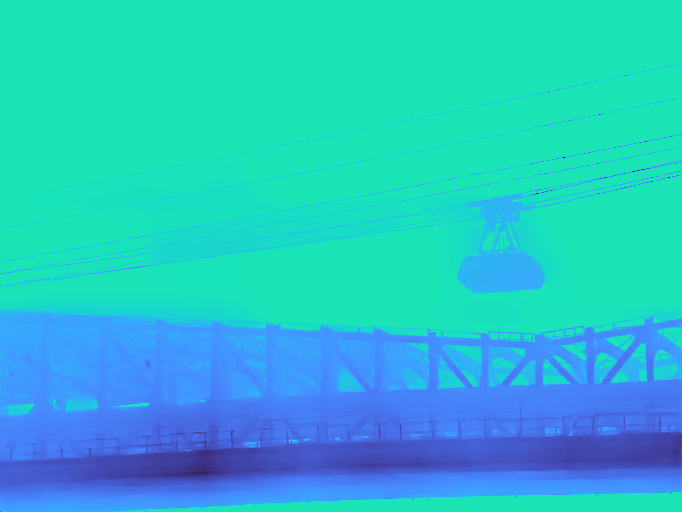} &
    \includegraphics[height=3.8cm]{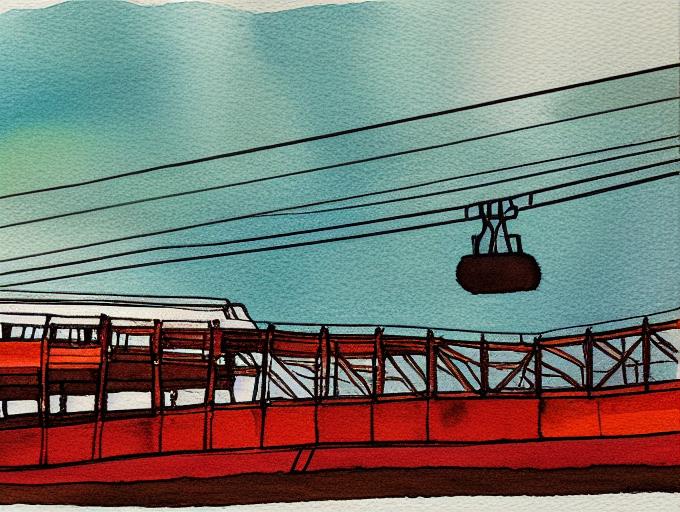}\\
    \raisebox{2cm}[0pt][0pt]{\rotatebox[origin=c]{90}{\footnotesize Depth Anything v2}}&
    \includegraphics[height=3.8cm]{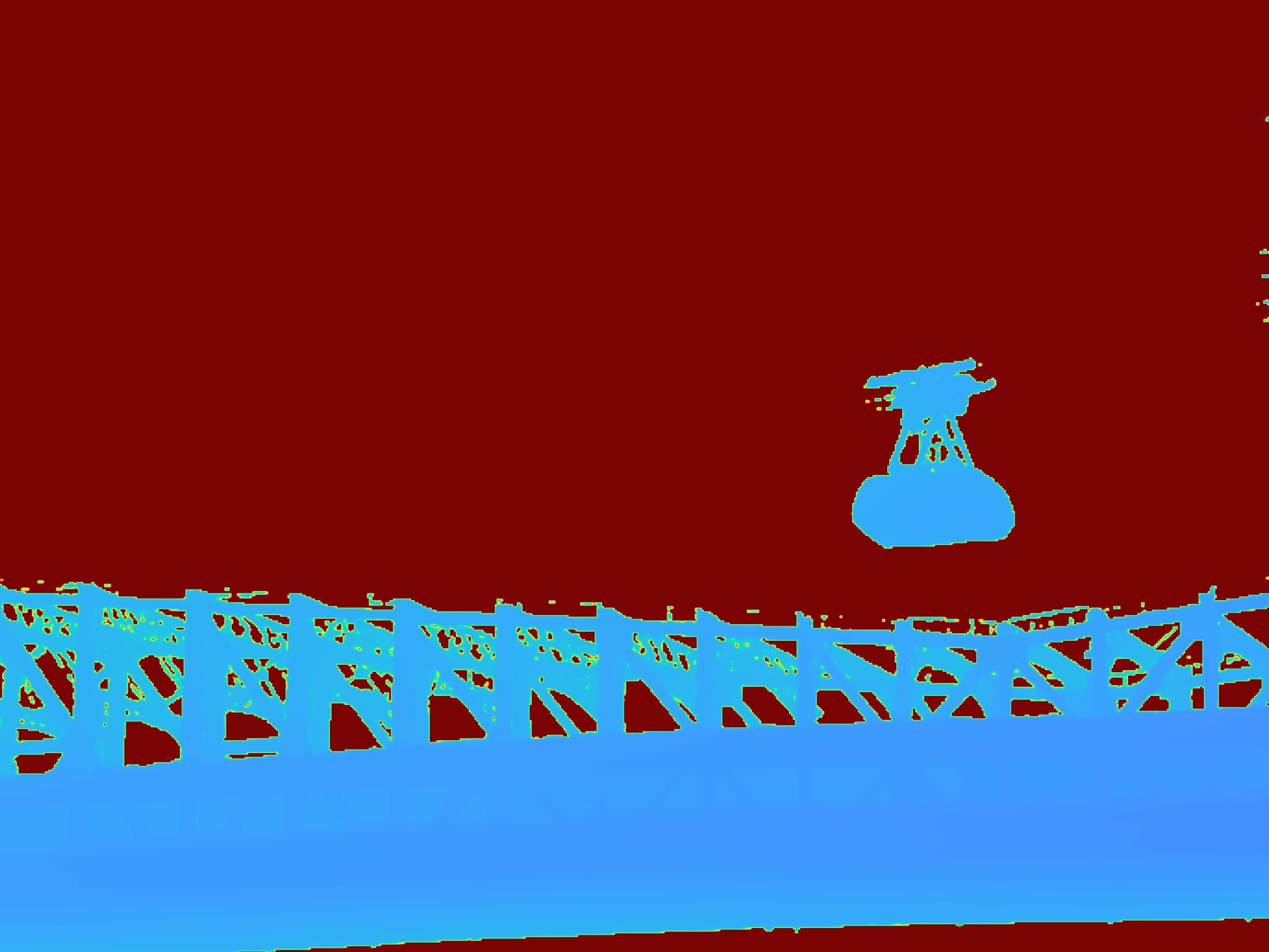} &
    \includegraphics[height=3.8cm]{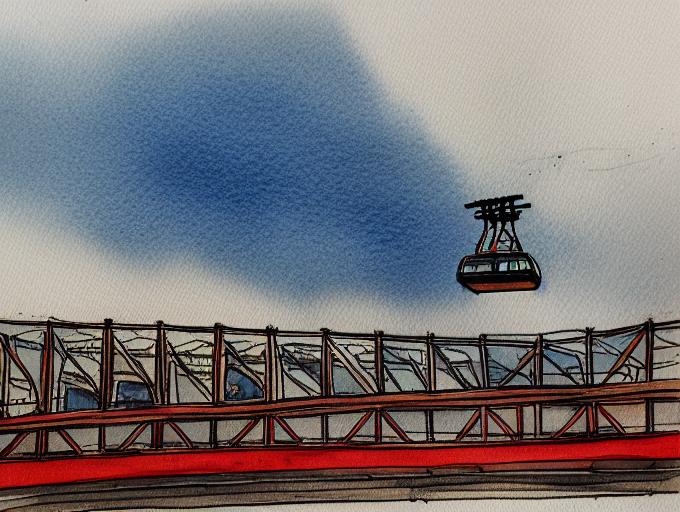}\\
    \raisebox{2cm}[0pt][0pt]{\rotatebox[origin=c]{90}{\footnotesize Marigold}}&
    \includegraphics[height=3.8cm]{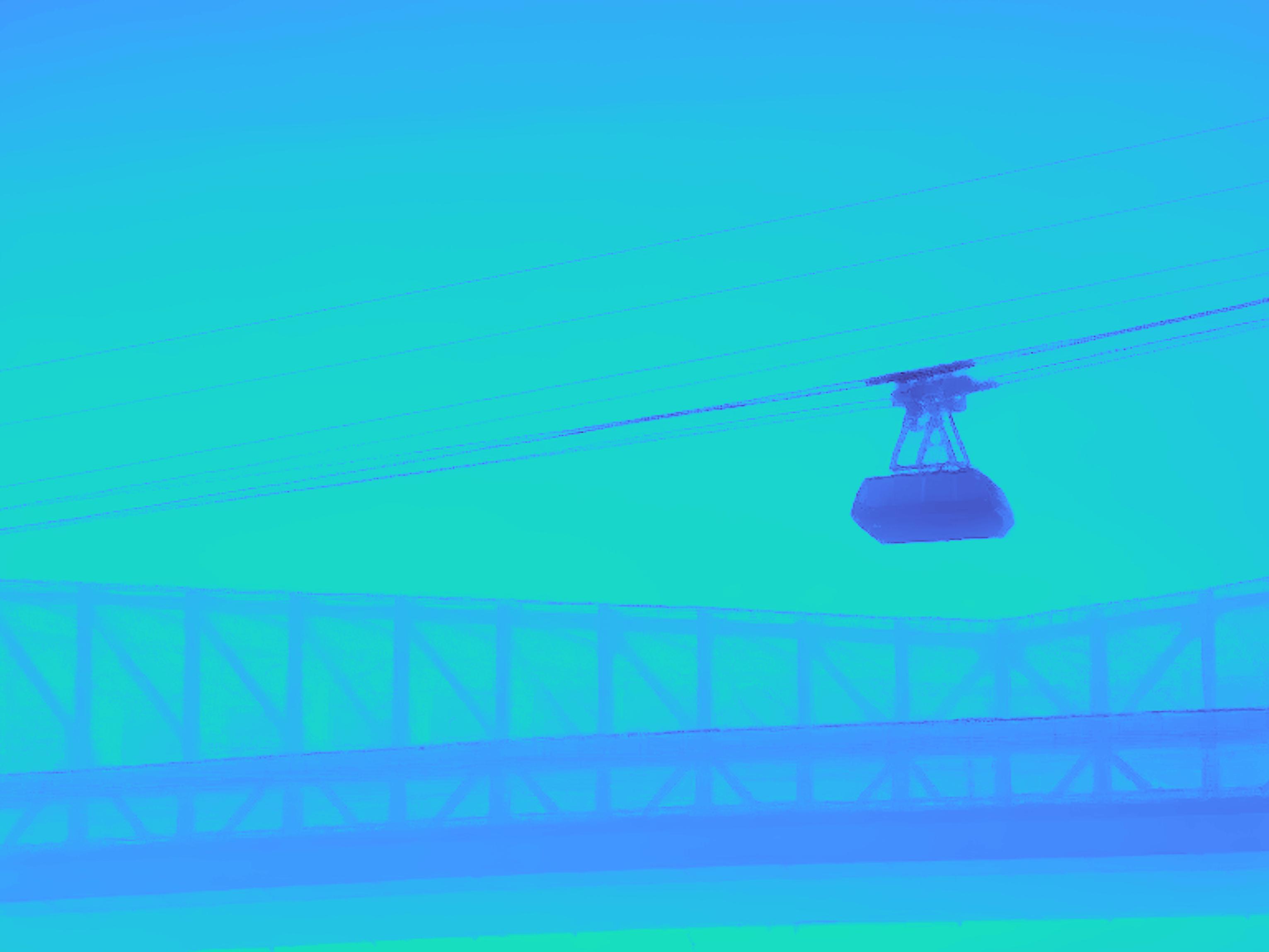} &
    \includegraphics[height=3.8cm]{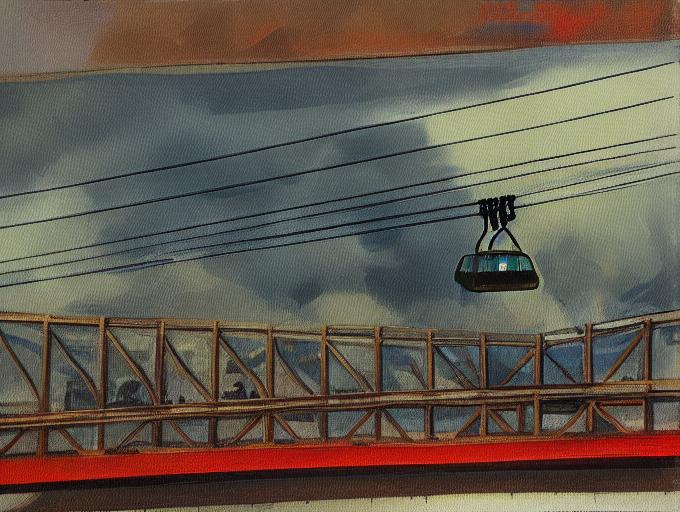}\\
    \raisebox{2cm}[0pt][0pt]{\rotatebox[origin=c]{90}{\footnotesize Metric3D v2}}&
    \includegraphics[height=3.8cm]{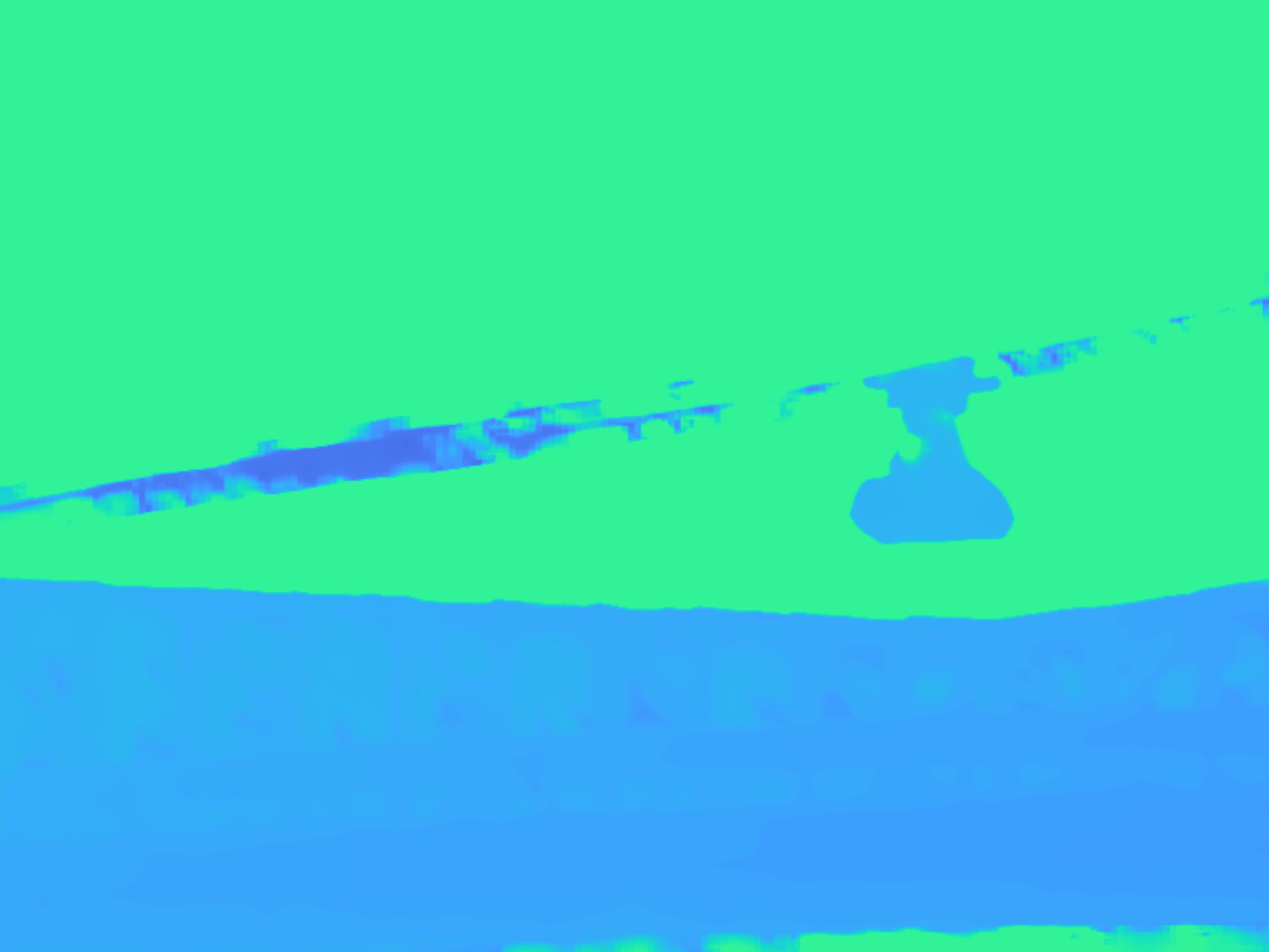} &
    \includegraphics[height=3.8cm]{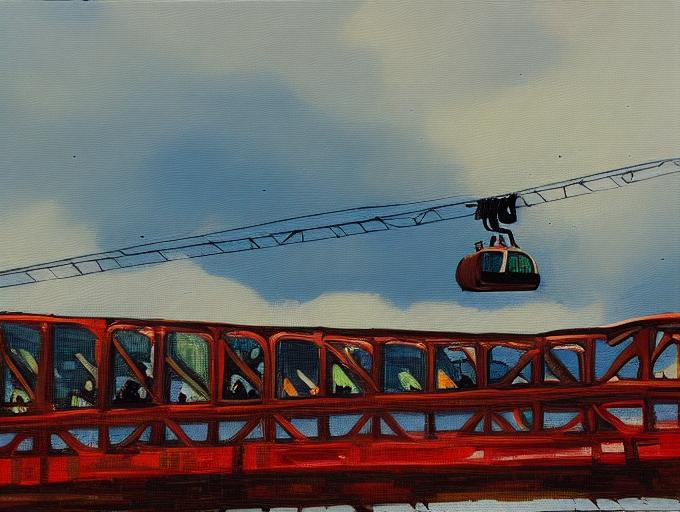}  \\     
    \end{tabular}
    \vspace{-2mm}
    \caption{\textbf{Comparison on conditional image synthesis.} We use ControlNet~\citep{Zhang2023:ICCVa} to synthesize a stylized image given a prompt (top row, right) and a depth map. The depth map is predicted from the input image~\citep{Li2022:IJCV} (top row, left) via Depth Pro, and baselines. The left column shows depth maps, the right column the synthesized image. For the baselines, missing cables (Depth Anything v2 \& Matric3D v2) or a spurious gradient in the sky (Marigold) alter the scene structure of the synthesized image.}
    \label{fig:applications_controlnet}
\end{figure}

\mypara{Depth-conditioned image synthesis.}
In this application we stylize an image through a text prompt via ControlNet~\citep{Zhang2023:ICCVb}. To retain the structure of the input image, we predict a depth map from the input image and use it for conditioning the image synthesis through a pretrained depth-to-image ControlNet SD 1.5 model.
Figure~\ref{fig:applications_controlnet} shows the input image, prompt, and predicted depth maps and synthesis results for Depth Pro, Deoth Anything v2, Marigold, and Metric3D v2. We find that only Depth Pro accurately predicts the cables and sky region, resulting in a stylized image that retains the structure of the input image. Baselines either miss cables, causing the cable car to float in mid-air (Depth Anything v2), or add a gradient to the sky (Marigold).

\begin{figure}[htbp]
    \centering
    \begin{tabular}{cc}
        % First Subfigure
        \begin{subfigure}[b]{0.47\textwidth}
            \centering
            \includegraphics[width=\textwidth]{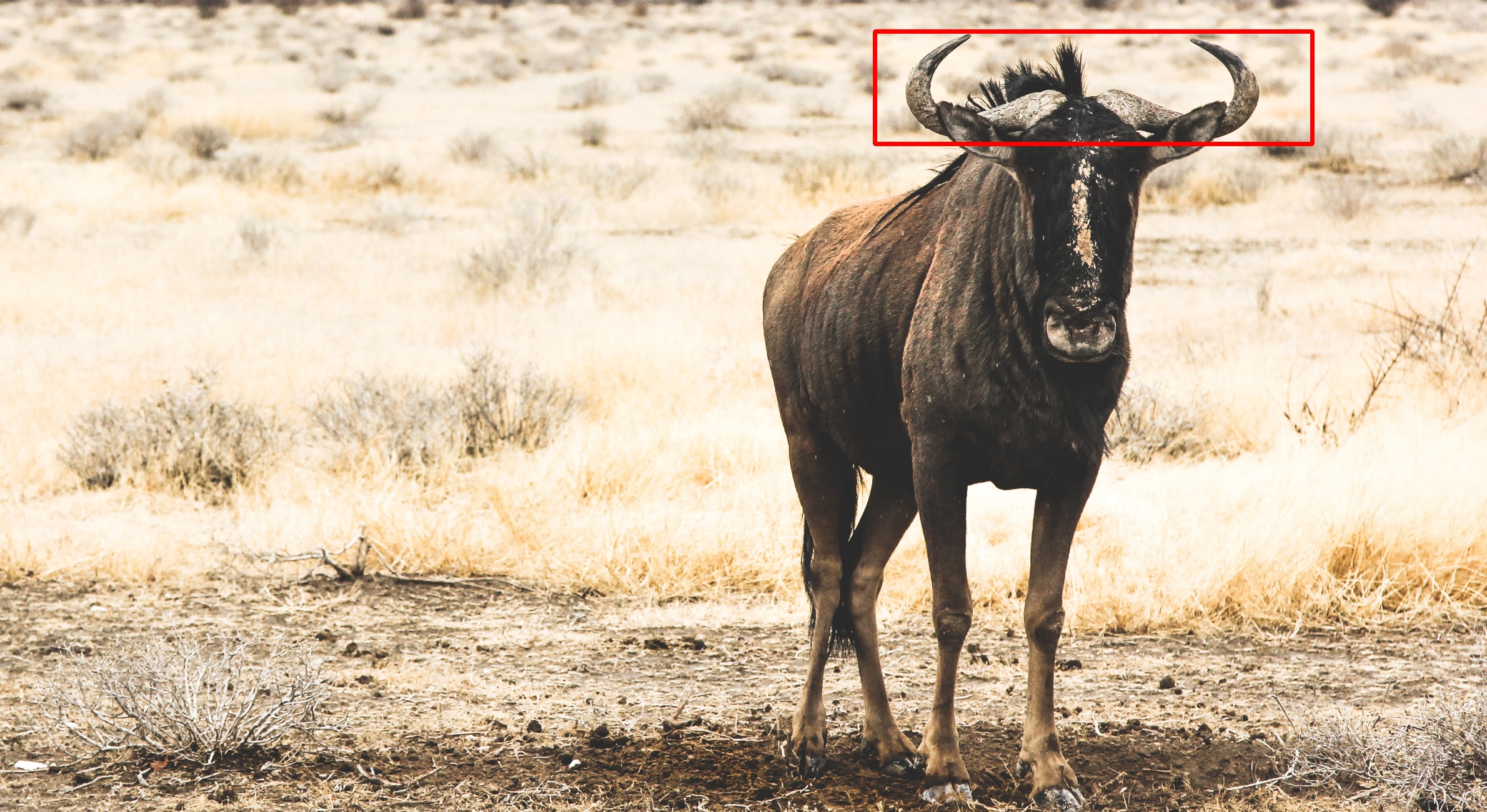}
            \vspace{1pt}
            \begin{minipage}{\textwidth}
                \centering
                \begin{minipage}{0.322\textwidth}
                    \centering
                    \includegraphics[width=\textwidth]{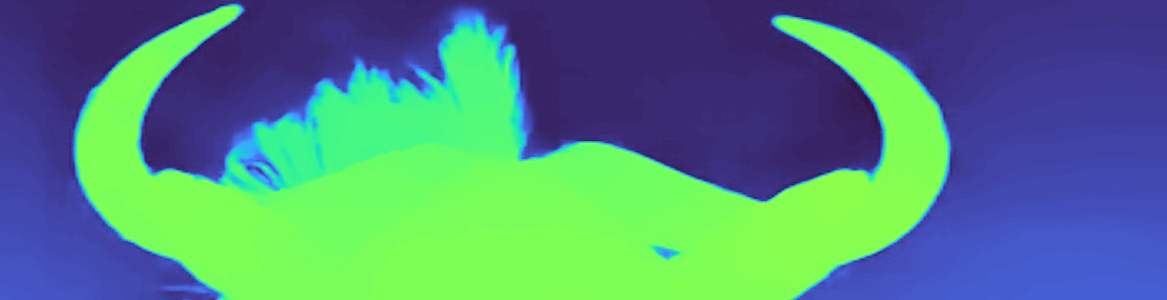}
                \end{minipage}
                \begin{minipage}{0.322\textwidth}
                    \centering
                    \includegraphics[width=\textwidth]{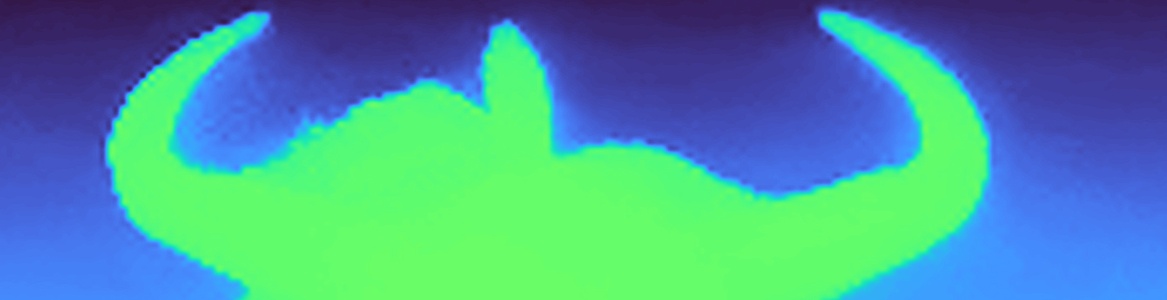}
                \end{minipage}
                \begin{minipage}{0.322\textwidth}
                    \centering
                    \includegraphics[width=\textwidth]{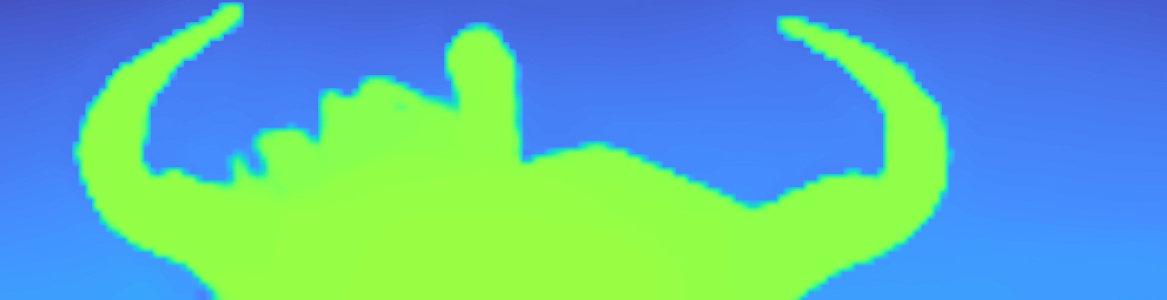}
                \end{minipage}
            \end{minipage}
            \vspace{1pt}
            \begin{minipage}{\textwidth}
                \centering
                \begin{minipage}{0.32\textwidth}
                    \centering
                    \includegraphics[width=\textwidth]{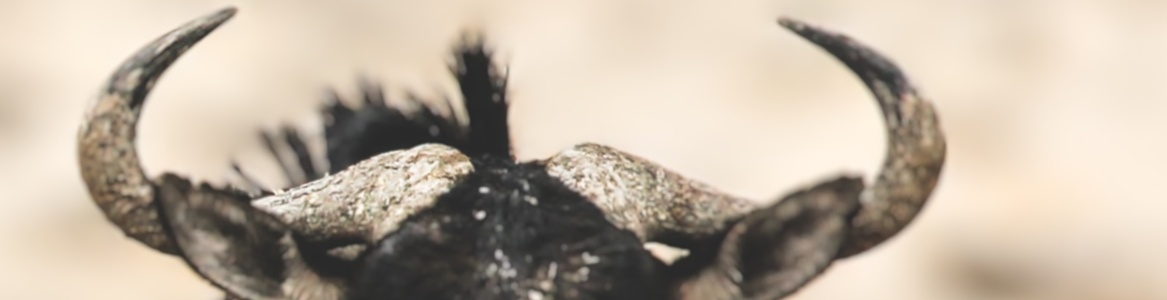}
                    \caption*{\tiny Depth Pro (ours)}
                \end{minipage}
                \begin{minipage}{0.32\textwidth}
                    \centering
                    \includegraphics[width=\textwidth]{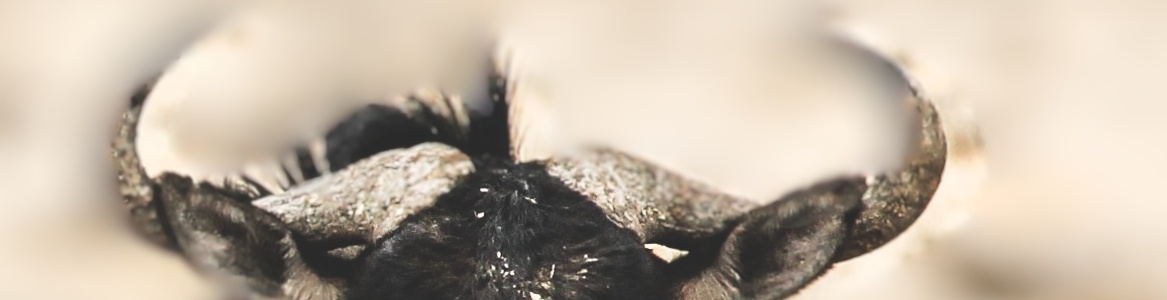}
                    \caption*{\tiny Marigold}% \citep{Ke2024:CVPR}}
                \end{minipage}
                \begin{minipage}{0.32\textwidth}
                    \centering
                    \includegraphics[width=\textwidth]{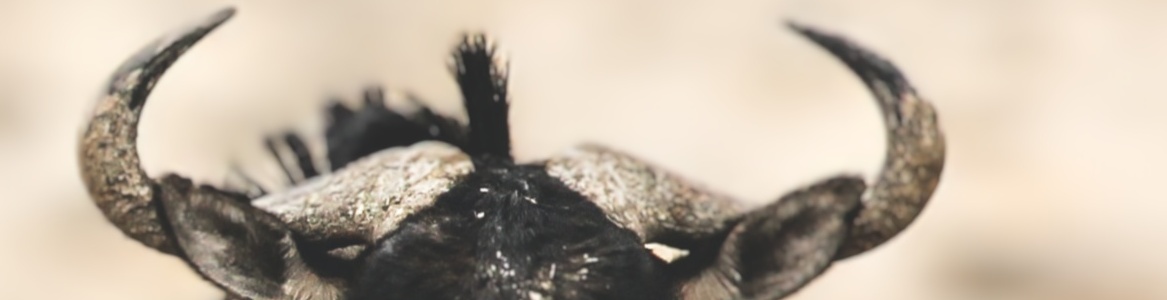}
                    \caption*{\tiny Depth Anything v2}% \citep{Yang2024:CVPR}}
                \end{minipage}
            \end{minipage}
        \end{subfigure}
        &
        % Second Subfigure
        \begin{subfigure}[b]{0.47\textwidth}
            \centering
            \includegraphics[width=\textwidth]{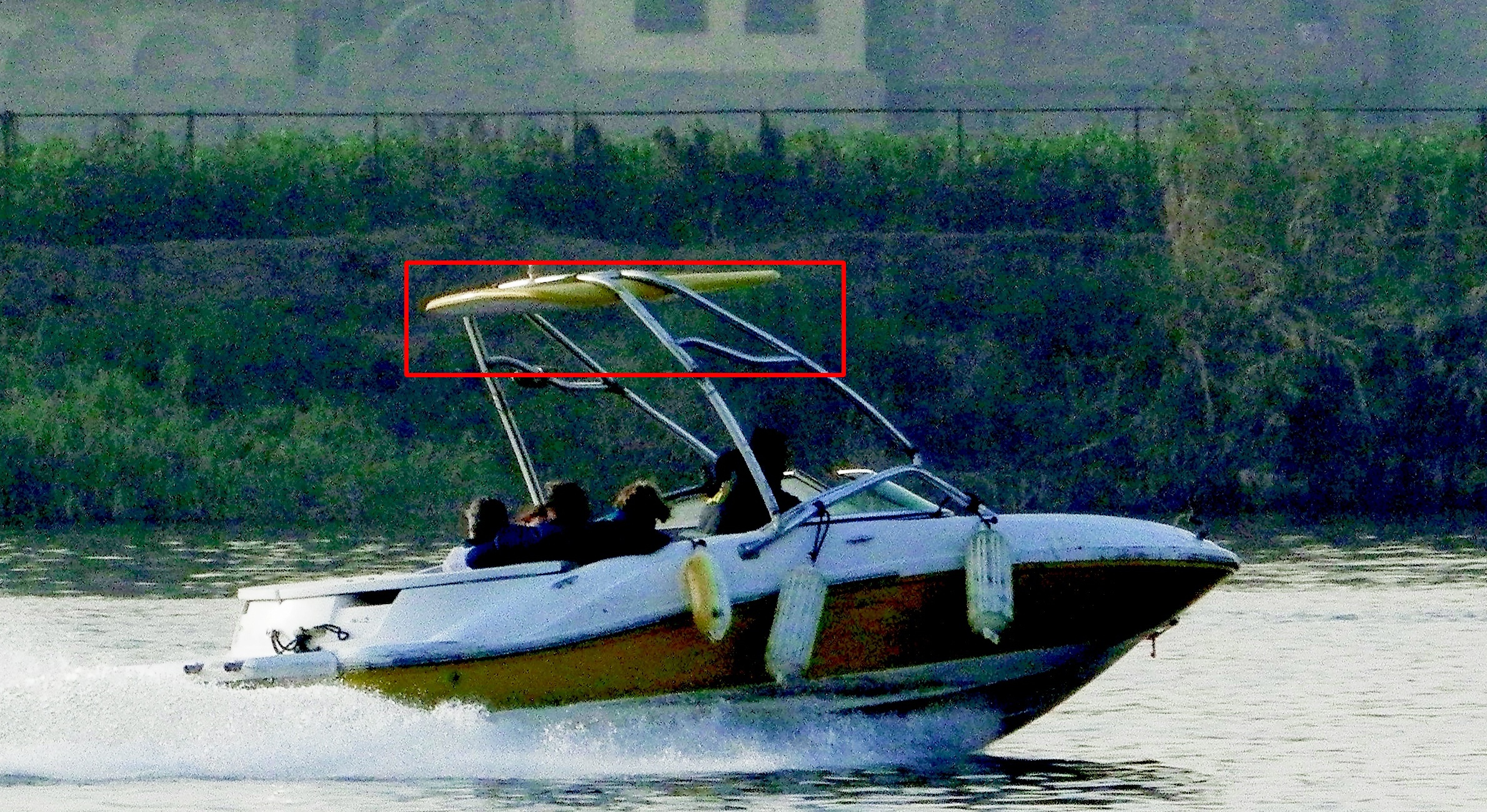}
            \vspace{1pt}
            \begin{minipage}{\textwidth}
                \centering
                \begin{minipage}{0.322\textwidth}
                    \centering
                    \includegraphics[width=\textwidth]{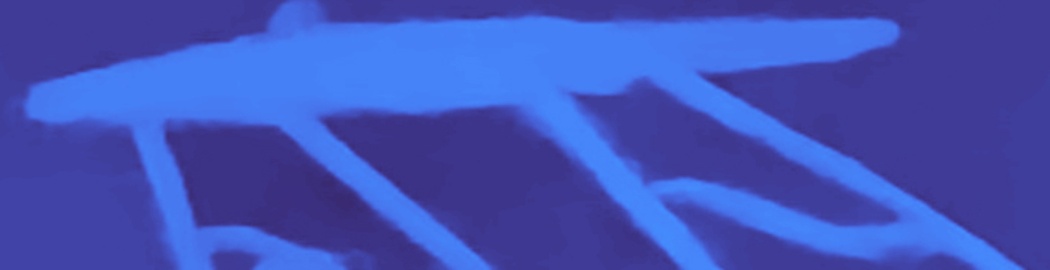}
                \end{minipage}
                \begin{minipage}{0.322\textwidth}
                    \centering
                    \includegraphics[width=\textwidth]{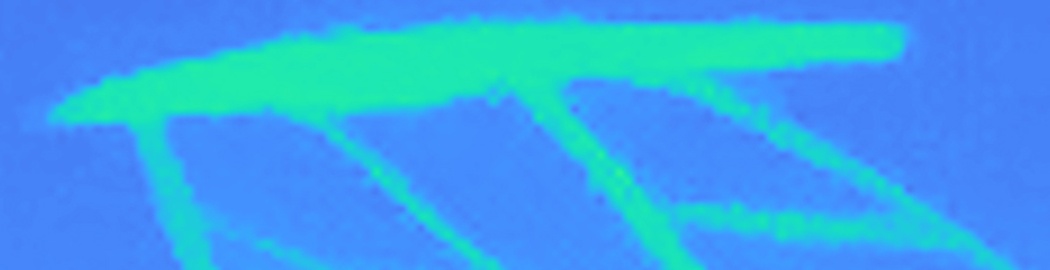}
                \end{minipage}
                \begin{minipage}{0.322\textwidth}
                    \centering
                    \includegraphics[width=\textwidth]{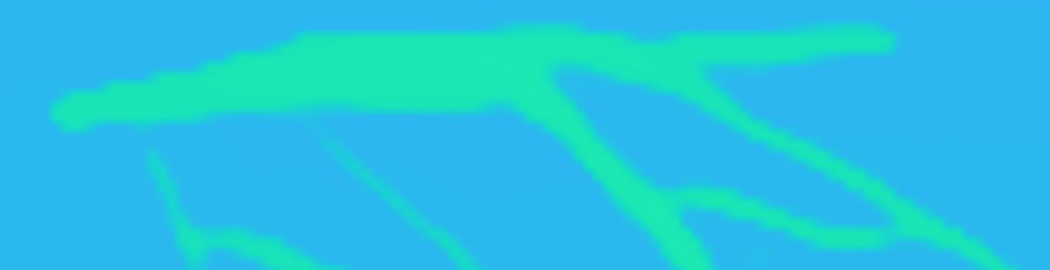}
                \end{minipage}
            \end{minipage}
            \vspace{1pt}
            \begin{minipage}{\textwidth}
                \centering
                \begin{minipage}{0.32\textwidth}
                    \centering
                    \includegraphics[width=\textwidth]{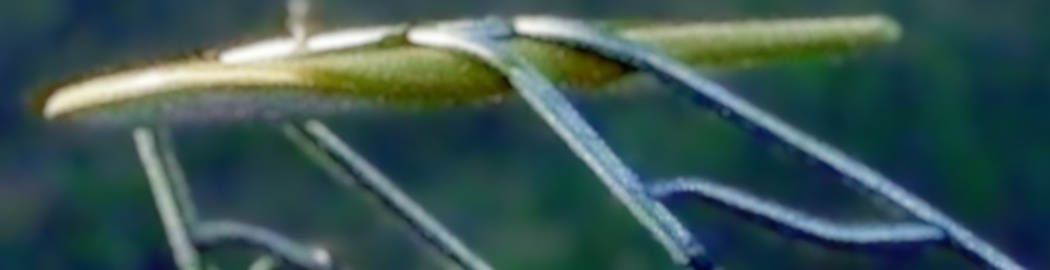}
                    \caption*{\tiny Depth Pro (ours)}
                \end{minipage}
                \begin{minipage}{0.32\textwidth}
                    \centering
                    \includegraphics[width=\textwidth]{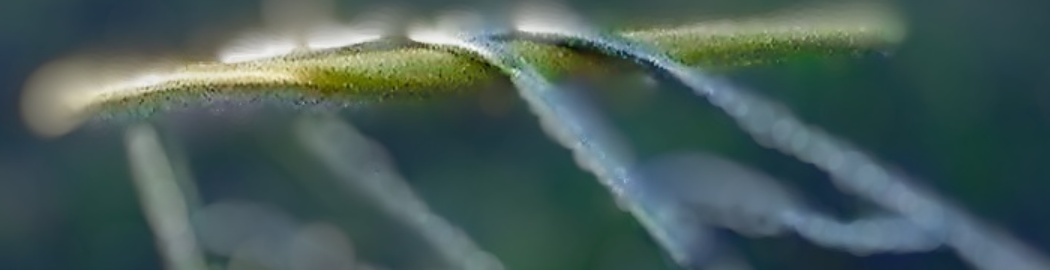}
                    \caption*{\tiny Marigold}% \citep{Ke2024:CVPR}}
                \end{minipage}
                \begin{minipage}{0.32\textwidth}
                    \centering
                    \includegraphics[width=\textwidth]{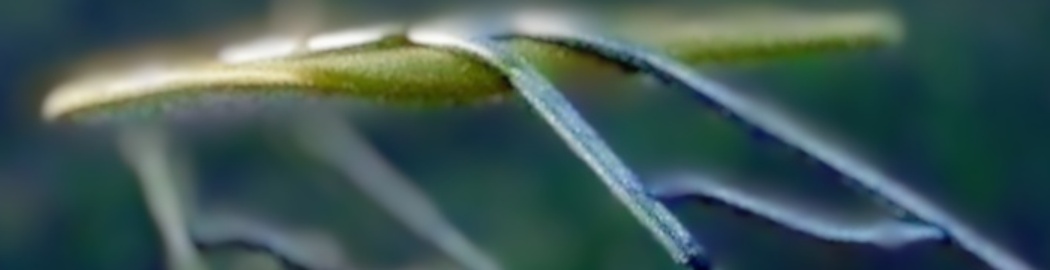}
                    \caption*{\tiny Depth Anything v2}% \citep{Yang2024:CVPR}}
                \end{minipage}
            \end{minipage}
        \end{subfigure}
    \end{tabular}
    \caption{\textbf{Comparison on synthetic depth of field.} We compare the synthetic depth of field produced by BokehMe~\citep{Peng2022:CVPR} using depth maps from Depth Pro, Marigold~\citep{Ke2024:CVPR}, and Depth Anything v2~\citep{Yang2024:CVPR}. Zoom in for detail.}
    \label{fig:bokehme}
\end{figure}

\mypara{Synthetic depth of field.}
Synthetic depth of field can be used to highlight the primary subject in a photo by deliberately blurring the surrounding areas. BokehMe~\citep{Peng2022:CVPR} introduces a hybrid rendering framework that marries a neural renderer with a classical physically motivated renderer. This framework takes a single image along with a depth map as input. In this context, it is essential for the depth map to delineate objects well, such that the photo's subject is kept correctly in focus while other content is correctly blurred out. Furthermore, the depth map should correctly trace out the details of the subject, to keep these (and only these) details correctly in focus. Figure \ref{fig:bokehme} shows the advantage afforded by Depth Pro in this application. (We keep the most salient object in focus by setting the refocused disparity (disp\_focus) hyperparameter of BokehMe as the disparity of the object.)

\end{document}